\documentclass[final]{cvpr}

\usepackage{times}
\usepackage{epsfig}
\usepackage{graphicx}
\usepackage{amsmath}
\usepackage{amssymb}
\usepackage{color}
\usepackage{colortbl}
\usepackage{pifont}
\usepackage{subfigure}
\usepackage{multirow}
\usepackage{caption}
\makeatletter
\@namedef{ver@everyshi.sty}{}
\makeatother
\usepackage{pgfplots}
\usepackage{mathtools}

\usepackage[pagebackref=true,breaklinks=true,colorlinks,bookmarks=false]{hyperref}

\definecolor{color1}{HTML}{DD614A}
\definecolor{color2}{HTML}{FF9B71}
\definecolor{color3}{HTML}{FFC300}
\definecolor{color4}{HTML}{6BA292}
\definecolor{color5}{HTML}{20639B}

\newcommand{\cmark}{\ding{51}}%
\newcommand{\xmark}{\ding{55}}%

\definecolor{red}{rgb}{1,0,0}
\definecolor{blue}{rgb}{0,0,1}
\definecolor{plain}{rgb}{0,0,0}
\definecolor{sunok}{rgb}{0,0,0}
\definecolor{kang}{rgb}{0,0,0}

\newcommand*{\basemodel}{CaffNet\xspace}
\newcommand*{\complexmodel}{CaffNet-C\xspace}
\makeatletter
 \def\hlinewd#1{%
      \noalign{\ifnum0=`}\fi\hrule \@height #1 \futurelet
      \reserved@a\@xhline}
\newcommand{\minus}{\scalebox{0.5}[1.0]{ $-$}}      

\def\cvprPaperID{8136} % *** Enter the CVPR Paper ID here
\def\confYear{CVPR 2021}

\newcommand\blfootnote[1]{%
  \begingroup
  \renewcommand\thefootnote{}\footnote{#1}%
  \addtocounter{footnote}{-1}%
  \endgroup
}

\begin{document}

%%%%%%%%% TITLE
\title{Looking into Your Speech: \\ Learning Cross-modal Affinity for Audio-visual Speech Separation}

\author{Jiyoung Lee$^{1*}$\\
{\tt\small easy00@yonsei.ac.kr}
\and
Soo-Whan Chung$^{1,2*}$\\
{\tt\small soowhan.chung@navercorp.com}
\and
Sunok Kim$^{3}$\\
{\tt\small sunok.kim@kau.ac.kr}
\and
Hong-Goo Kang$^{1\dagger}$\\
{\tt\small hgkang@yonsei.ac.kr}
\and
Kwanghoon Sohn$^{1\dagger}$\\\
{\tt\small khsohn@yonsei.ac.kr}
\and
$^{1}$Department of Electrical \& Electronic Engineering, Yonsei University, Korea\\
$^{2}$Naver Corporation, Korea\;
$^{3}$Korea Aerospace University, Korea\\
{\tt\small \url{https://caffnet.github.io/}}
}

\twocolumn[{%
  \renewcommand\twocolumn[1][]{#1}%
  \maketitle   
  \begin{center}
     \centering    
     \vspace{-23pt} 
     \includegraphics[width=0.94\textwidth]{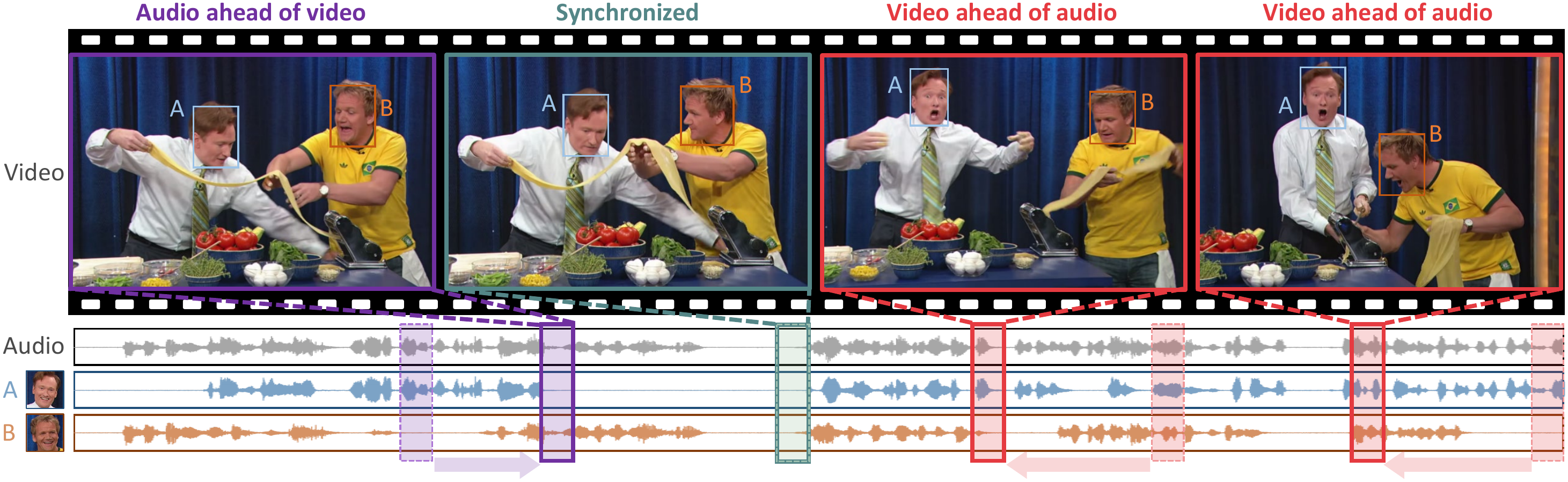}
     \vspace{-8pt}      
     \captionof{figure}{
     We wish to hear individual speech of a desired speaker only even if there is frame discontinuity in the audio-visual data.
     When audio and video segments are taken from different points in time~(\textbf{solid box}), it is intuitively difficult to separate speech of each speaker compared to the aligned cases~(\textbf{dashed box}). Best viewed in color.
     }\label{fig:1}
  \end{center}
}]

\blfootnote{
\hspace{-14pt}$^{*}$ Both authors contributed equally to this work\\
$^{\dagger}$ Corresponding authors\\
 This work was supported by the National Research Foundation of Korea~(NRF) grant funded by the Korea government~(MSIT). (NRF-2021R1A2C2006703).}

\begin{abstract}
\vspace{-5pt}
In this paper, we address the problem of separating individual speech signals from videos using audio-visual neural processing.
Most conventional approaches utilize frame-wise matching criteria to extract shared information between co-occurring audio and video.
Thus, their performance heavily depends on the accuracy of audio-visual synchronization and the effectiveness of their representations.
To overcome the frame discontinuity problem between two modalities due to transmission delay mismatch or jitter,
we propose a cross-modal affinity network (\basemodel) that learns global correspondence as well as locally-varying affinities between audio and visual streams.
Given that the global term provides stability over a temporal sequence at the utterance-level, this resolves the label permutation problem characterized by inconsistent assignments. 
By extending the proposed cross-modal affinity on the complex network, we further improve the separation performance in the complex spectral domain.
Experimental results verify that the proposed methods outperform conventional ones on various datasets, demonstrating their advantages in real-world scenarios.
\vspace{-15pt}

\end{abstract}
\section{Introduction}
Humans have a remarkable auditory system that can perceive sound sources separately in their conversations even in the presence of many surrounding sounds, including background noise, crowded babbling, thumping music, and sometimes other loud voices~\cite{haykin2005cocktail,bregman1994auditory}.
However, reliably separating a target speech signal for human-computer interaction (HCI) systems such as speech recognition~\cite{hinton2012deep,bahdanau2016end,chan2016listen}, speaker recognition~\cite{variani2014deep,snyder2018x,kwon2020intra}, and emotion recognition~\cite{jaitly2011learning,trigeorgis2016adieu} is still a challenging task because it is an ill-posed problem.

With the impressive advent of deep learning technologies that utilize high-dimensional embeddings~\cite{hershey2016deep,chen2017deep,luo2018tasnet}, it is possible nowadays to simultaneously analyze the unique acoustic characteristics of different speakers even from mixed signals.
Although these deep learning-based methods are effective compared to conventional statistical signal processing-based ones, they are prone to a label permutation (or ambiguity) error due to their frame-by-frame or short segment-based processing paradigm~\cite{hershey2016deep,weng2015deep}.
In order to address this problem, permutation invariant training~\cite{yu2017permutation,kolbaek2017multitalker} that utilizes a permutation loss criterion was presented, but the label ambiguity problem still occurs at the inference stage, especially for unseen speakers.

Leveraging the visual streams of target speech signals can be one of the best alternatives. In psychology, several experiments have proved that looking at speakers' faces is helpful for auditory perception under background noise environments~\cite{sumby1954visual,partan1999communication}.
For example, lip reading, which matches lip movements onto utterances, is widely used to recognize others' words better~\cite{chung2017lip}.
In audio-visual speech separation (AVSS) systems, audio and visual features are used together or complement each other to derive unique characteristics~\cite{afouras2018conversation,afouras2019my,afouras2020self,li2020deep,ephrat2018looking,lu2018listen,chung2020seeing}.
Mostly, AVSS first extracts the common correspondence features between speaker/linguistic information of speech signals and face/articulatory lip movements of video signals, after which the extracted features are exploited for the following source separation task.
Consequently, the AVSS problem can be viewed as a local matching (\ie frame-wise matching) task, where segmented visual features are matched with frames of specific sounds.
Thus, the separation performance highly depends on the alignment accuracy between audio and video streams.

In real-world scenarios, however, audio and video are recorded from different devices with their own specifications, and they are transmitted through independent communication channels and saved with different codec protocols.
These practical issues frequently cause mutually unaligned states in talking videos. 
\figref{fig:1} shows an example of a video with a speech that has physical errors in its video contents, where sometimes audio plays ahead of video and vice versa.
When there are even subtle data transformations caused by jitters, omissions, and out-of-synchronization in video streams, conventional local matching strategies~\cite{afouras2018conversation,li2020deep,lu2018listen} are vulnerable.
This issue can be detrimental to the performance of AVSS systems in video-telephony, broadcasting, video conferencing, or filming.

In this paper, we highlight those limitations and tackle the alignment problems in AVSS processing.
We propose a novel cross-modal affinity network for robust speech separation, referred to as \textit{\basemodel}, by utilizing visual cues in consideration of relative timing information.
Affinity, \ie mutual correlation, learned in \basemodel compensates for abrupt discontinuities in audio-visual data without external information or additional supervision.
Furthermore, we propose an affinity regularization module that tiles the diagonal term of the affinity matrix to match audio-visual sequences at the utterance level.
Since the affinity regularization provides a global positional constraint, it avoids the label permutation problem that occurred by inconsistent assignment over time of the speech signals to the visual target.
In addition, considering the estimation of the magnitude mask in tandem with the phase mask is one of the keys to reasonable speech reconstruction because such factors are correlated with each other~\cite{krawczyk2014stft,lee2019ajoint}.
To accomplish this, we extend \basemodel to have a complex-valued convolution network architecture~\cite{lee2017fully,trabelsi2017deep,choi2019phase} such that speech quality is indeed increased by restoring the mask of the magnitude and phase spectrum together.
We demonstrate the effectiveness of the proposed networks with extensive experiments, achieving large improvements in unconditioned scenarios on several benchmark datasets~\cite{afouras2018deep, lrs3, voxceleb2}.

\section{Related Work}
\paragraph{Audio-visual Speech Separation.}
In terms of multi-sensory integration, it has been proved that looking at talking faces during conversation is helpful for speech perception~\cite{mesgarani2012selective}.
Inspired by this psychological mechanism, numerous works have tried to effectively utilize visual context on speech separation tasks~\cite{fisher2000learning,barzelay2007harmony,lu2018listen,afouras2018conversation,owens2018audio,ephrat2018looking,afouras2019my,afouras2020self,chung2020facefilter,li2020deep}.
With the emergence of deep neural networks and the availability of new large-scale datasets~\cite{chung2017lip,lrs3,voxceleb2,ephrat2018looking}, a series of works~\cite{owens2018audio, hou2018audio} have been published in the past few years on audio-visual speech separation (AVSS). 
Although they have shown promising results in various speech-oriented applications, these methods have concentrated on isolating the magnitude of speech only, which restricts their applicability.
To alleviate this limitation, several works have been proposed to generate both magnitude and phase masks \cite{afouras2018conversation,ephrat2018looking,li2020deep,afouras2020self}.
However, these methods generate phase masks with estimated magnitude masks and noisy phase without learning complex-valued representations, which requires them to consider the correlation between each complex component.
This limitation has manifested itself in significantly degraded performance under extremely noisy conditions~\cite{choi2019phase}.  
Furthermore, the aforementioned methods have all assumed one-to-one correspondence between the audio-visual segments.

\begin{figure*}[ht]
\begin{center}
    \includegraphics[width=0.95\textwidth]{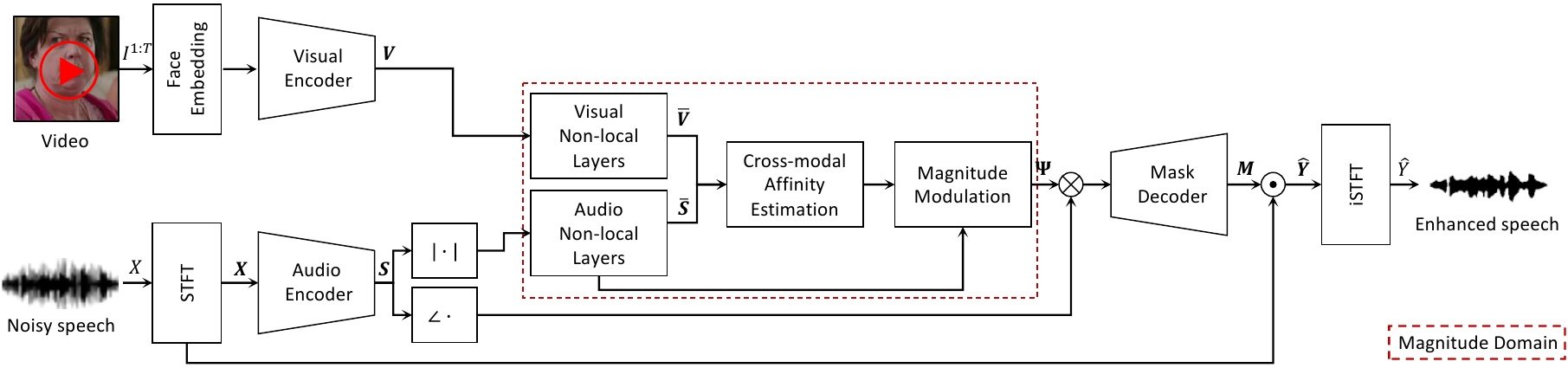}\vspace{-5pt}
    \caption{Overall network configuration: (1) encoding audio and visual features; (2) learning cross-modal affinity; (3) predicting spectral soft mask $\mathbf{M}$ to reconstruct target speech $\mathbf{\hat{Y}}$. A red-dotted box means the magnitude operation processing.}
    \label{fig:2}\vspace{-20pt}
\end{center}
\end{figure*}

Most recently, this problem has been tackled in~\cite{afouras2019my,chung2020facefilter} for situations in which visual face sequences were not fully reserved.
In~\cite{afouras2019my}, although they considered the case that visual cues abruptly disappear due to occlusion, it still required video sequences aligned to audio.
Even though the speaker identity extracted from the still face image seemed to provide a promising visual cue for the separation~\cite{chung2020facefilter}, it showed far less separation performance than ones using visual sequences since they only regulated global information rather than local contexts. 
In this paper, we leverage sequential audio-visual frames as the inputs to our networks under the assumption that locally misaligned visual frames with audio frames can still provide local context and speaker identity for robust speech separation. \vspace{-13pt}

\paragraph{Cross-modal Alignment.}
As audio and video sequences are recorded using different devices, synchronization problem often appears in recordings.
Most recent audio-visual synchronization methods rely on cross-modal representation techniques that measure the linguistic similarity between audio-visual embedding pairs~\cite{chung2019perfect,chung2016out,kim2021end}.
However, there has been little work on exploring the problem of synchronization along with mismatched audio and video pairs because prior works generally assumed that a paired audio and video set has only one speech.
More related to our work are affinity-based multi-modal approaches in various other challenging tasks, such as music sound separation~\cite{gan2020music}, emotion recognition~\cite{lee2020multi}, language understanding~\cite{tsai2019multimodal}, and self-supervised learning~\cite{cheng2020look}.
We further extend the cross-modal affinity learning to generate time-independent audio-visual representations using an affinity regularization with an utterance-wise matching criterion.

\section{Approach}\label{sec:method}
\subsection{Motivation and Overview}\vspace{-1pt}
In general, humans experience severe confusion when there is a linguistic discrepancy between what we see and what we hear, \ie a difference between the perceived words from the mouth and actual speech.
This is called cognitive dissonance, which is known as the McGurk effect~\cite{mcgurk1976hearing}.
This effect could be observed in previous frame-wise matching based methods~\cite{afouras2018deep,afouras2018conversation,li2020deep,owens2018audio,lu2018listen} inducing poor performance when the cross-modal data is conflicted.
To deal with such inconsistency problem, we introduce a \basemodel to estimate time-frequency soft masks to isolate a single speech signal from a mixture of sounds (such as other speakers and background noise), taking into account time-agnostic mutual correlation.

Concretely, our model is split into three parts, including an audio-visual encoder, learning cross-modal affinity, and soft-mask estimation, as shown in~\figref{fig:2}.
The key idea of \basemodel is to learn cross-modal affinity between the audio and video streams even if they have different sampling rates in the wild environments.
By this, we mean that information from the video stream stretches or compresses to match the audio signal for the reconstruction of the target speaker's speech regardless of the frame discontinuity problem.
Due to matching ambiguity generated in parts that are muted or that have similar pronunciations from simultaneous speakers, the initially computed affinity includes erroneous values and causes the label permutation problem while degrading the separation performance. 
We resolve this problem by suggesting an affinity regularization to induce global consistency of cross-modal affinity.
Furthermore, we extend this approach to complex-valued neural networks, estimating the magnitude and phase components jointly.

Given a noisy time-domain speech $X$, \basemodel is trained to isolate a clean speech $Y$ from $X$ with corresponding a user-chosen speaker's face video $I^{1:T}$, where $T$ is a length of the video stream.
The noisy sound $X=Y+H$ is assumed to be a sum of clean speech $Y$ and natural environmental factors $H$ such as background noise, distortions in speech, and sound from other speakers.
As it has been a common practice to transform a time-domain speech to a time-frequency representation (\ie spectrogram) via short-time Fourier transform (STFT), each of the corresponding time-frequency representations for $X$, $H$, and  $Y$ is computed by 512-point STFT and denoted by $\mathbf{X} \in \mathbb{C}$, $\mathbf{H} \in \mathbb{C}$ and $\mathbf{Y} \in \mathbb{C}$, respectively.

\subsection{Cross-modal Affinity Network}\vspace{-1pt}
\paragraph{Audio-visual Encoder.}
As in~\cite{afouras2018conversation,afouras2019my,gan2020music}, the audio-visual encoder has a two-stream architecture consisting of an audio encoder stream and a video encoder stream, which take noisy audio and video frames containing the target face, respectively.
At first, the audio and video encoders generate their own embedding features independently.
In specific, the audio encoder $\mathcal{F}_s$ takes the magnitude spectrum of consecutive input frames, $\vert \mathbf{X} \vert$. The audio embedding features are extracted by stacked 1D convolutional layers $\mathbf{S} = \mathcal{F}_s{(\vert \mathbf{X} \vert)}$,
where $\mathbf{S} \in \mathbb{R}^{N \times C}$ is a speech embedding feature, $C$ and $N$ indicate the dimension of a channel and the temporal length of the spectrogram, respectively.
Besides, visual features are extracted from the temporal stack of five consecutive video frames via the state-of-the-art audio-visual synchronization model $\mathcal{E}(\cdot)$~\cite{chung2019perfect} by a feed-forward process.
Finally, a visual embedding feature $\mathbf{V} \in \mathbb{R}^{M \times C}$ is obtained through the visual encoder $\mathcal{F}_v$:
\begin{equation}
\mathbf{V} = \mathcal{F}_v{(\Pi(\mathcal{E}_{f}{(I^{1:5})}, \mathcal{E}_{f}{(I^{2:6})}, \cdots, \mathcal{E}_{f}{(I^{T\text{-}4:T})}))},
\end{equation}
where $I$ denotes video frames, $T$ is the number of video frames, $\Pi(\cdot)$ indicates a concatenation operator, and $M=T-4$ is an entire length of clips tied 5 frames.
As the audio and video have different sampling rates, it requires either up-sampling or down-sampling process to equalize the temporal resolution of audio and video embedding matrices~\cite{afouras2018conversation, li2020deep}. However, our network is fully-convolutional network that effectively learns the audio-visual affinity regardless of the size of each embedding matrix.\vspace{-13pt}

\begin{figure}[t]
\centering
    {\includegraphics[width=0.9\linewidth]{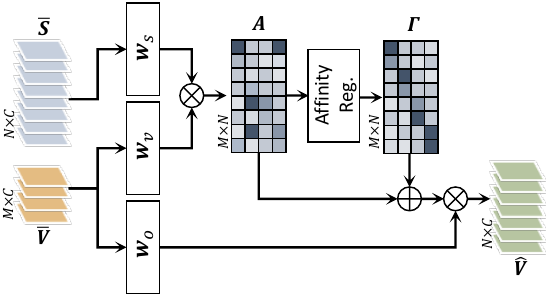}}\hfill\\
    \vspace{-4pt}
    \caption{Illustration of cross-modal affinity estimation module.
    It takes speech feature $\mathbf{\bar{S}}$ and video feature $\mathbf{\bar{V}}$ to calculate the affinity matrix $\mathbf{A}$.
    The cross-modal identity matrix $\mathbf{\Gamma}$ regularizes the affinity matrix $\mathbf{A}$ to maintain global correspondence.
    }\vspace{-8pt}
    \label{fig:3}
\end{figure}

\paragraph{Learning Cross-modal Affinity.}
We assume that audio and video embeddings are naturally out of joint in the unconstrained environments due to temporal mismatch between two media.
Considering the fact that learning affinity can draw linguistic dependencies between audio and visual features, it is possible to model relative timing dependencies without considering their distances.

\begin{figure}[t]
\begin{center}
    \subfigure[]{\includegraphics[width=0.32\linewidth]{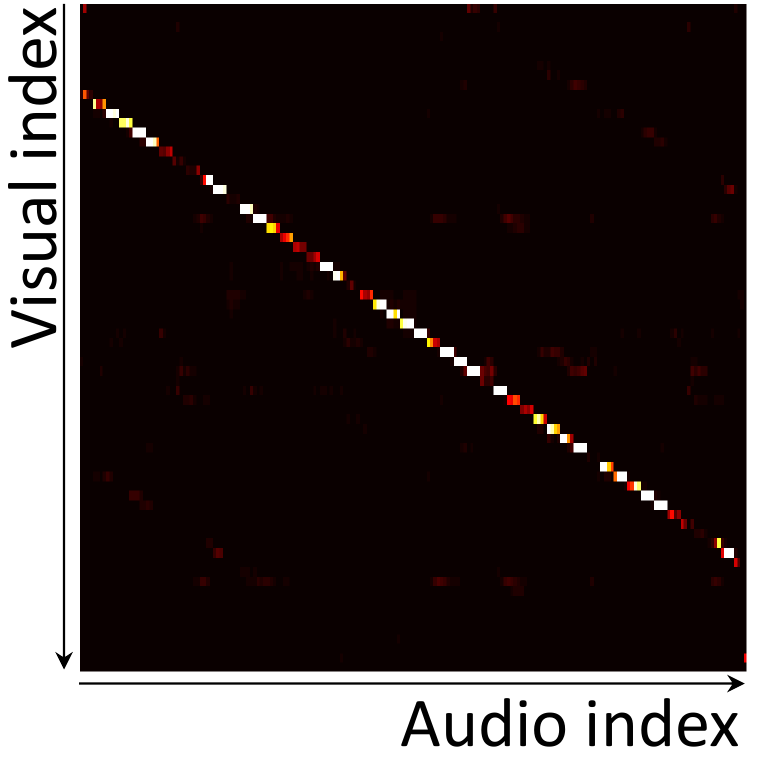}\label{fig:4a}}\hfill
    \subfigure[]{\includegraphics[width=0.32\linewidth]{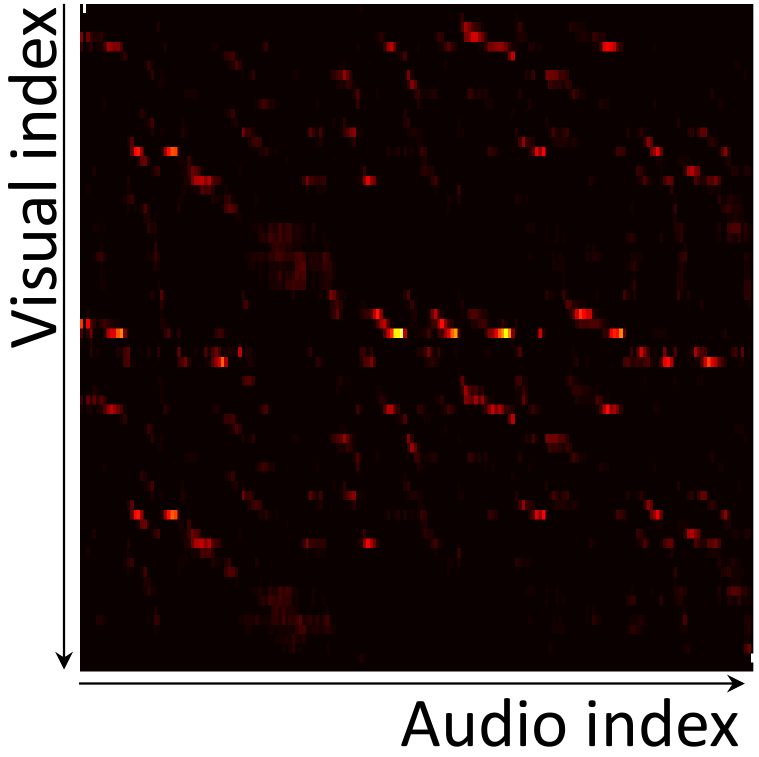}\label{fig:4b}}\hfill
    \subfigure[]{\includegraphics[width=0.32\linewidth]{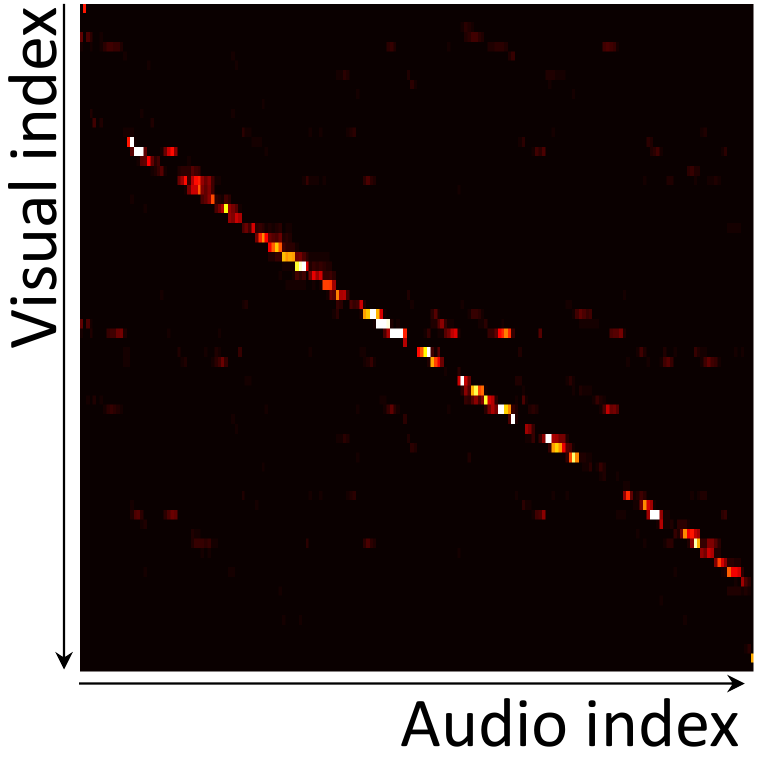}\label{fig:4c}}\hfill\\
    \vspace{-9pt}
    \captionof{figure}{Visualization of affinity matrices. 
    (a) When audio contains only one voice, it shows monotonous pattern with audio-visual correspondence.
    (b) When there is mixture input, the pattern is tangled in absence of sequential consistency.
    (c) Final affinity is arranged with the help of affinity regularization.}\vspace{-19pt}
    \label{fig:4}
\end{center}
\end{figure}

More specifically, we first extract audio feature $\mathbf{\bar{S}}$ and visual feature $\mathbf{\bar{V}}$ by feeding outputs of the audio-visual encoder to modality-separated two non-local layers~\cite{wang2018non} to measure the affinity on the nearest embedding space as possible.
Then, an affinity matrix $\mathbf{A}_{i,j}$ between $i$-th audio feature and $j$-th visual feature is computed using cosine similarity with $L_2$ normalization on an embedding space:
\begin{equation}
\mathbf{A}_{i,j} = \text{softmax}(\frac{< \mathbf{\bar{S}}_i \mathbf{w}_s, \mathbf{\bar{V}}_j \mathbf{w}_v >}{\Vert \mathbf{\bar{S}}_i \mathbf{w}_s \Vert_2 \Vert \mathbf{\bar{V}}_j \mathbf{w}_v \Vert_2} ),
\label{eq:aff}
\end{equation}
where $\mathbf{w}_s$ and $\mathbf{w}_v$ are embedding weights of audio and visual features, respectively, as illustrated in~\figref{fig:3}. 
In \eqref{eq:aff}, softmax activation function is applied row-wise for normalization to obtain an affinity matrix $\mathbf{A} \in \mathbb{R}^{N \times M}$.
Ideally, the linear pattern can be found along the diagonal of the affinity matrix as shown in~\figref{fig:4a}.
However, we observe that the affinity weights in different frames are often similar in the silent speech or regions having similar pronunciation between different speakers as depicted in~\figref{fig:4b}, which can cause the label permutation problem.
To resolve this problem, we propose an affinity regularization to penalize the probabilistic affinity matrix corresponding to global alignment context to reliably infer the spectrogram mask of an interest such that
\begin{equation}
    \mathbf{\Gamma}_{i,j} = \frac{\text{exp}(\textstyle\sum_{i,j} \mathcal{D}^{k}_{i,j}\mathbf{A}_{i,j} / \tau)}{\textstyle\sum_{k} \text{exp} ( \textstyle\sum_{i,j} \mathcal{D}^{k}_{i,j}\mathbf{A}_{i,j} / \tau )},
\end{equation}
where $\mathbf{\Gamma}$ is a cross-modal identity matrix, $k \in \mathcal{N}_k$ is the search window of offset range across the diagonal term, $\tau$ is a softening parameter~\cite{hinton2015distilling} set as 0.1, and $\mathcal{D}$ is a diagonal mask which satisfies:
\begin{equation}
\mathcal{D}^{k}_{i,j}=
    \begin{cases}
      1, & \text{if}\ \frac{f_a}{f_v}(j-k)+1 \leq i \leq \frac{f_a}{f_v}(j-k+1) \\
      0, & \text{otherwise},
    \end{cases}
\end{equation}
where $f_*$ is a sampling rate of each segment (\eg, ${f_a}/{f_v}=4$ in our experimental setting).
In our experiments, we search for the offsets over $[-9, +9]$ frame range, where negative offset means that audio is ahead of video and vice versa.
We set $\mathcal{N}_k \in [0, \cdots, 19]$ and if the input pair is matched in a timely manner, diagonal term appears from the 9-th video frame index as shown in~\figref{fig:4c}.
\figref{fig:4} clearly shows that the regularization encourages the model to maintain temporal consistency in the matching process.
Then the final visual features $\mathbf{\hat{V}} \in \mathbb{R}^{N \times C'}$ are then represented as follows:
\begin{equation}
    \mathbf{\hat{V}}_i = \textstyle\sum_j (\mathbf{A}_{i,j}+ \gamma \mathbf{\Gamma}_{i,j}) \cdot (\mathbf{w}_o \mathbf{\bar{V}}_j)^\top,
\end{equation}
where the identity matrix $\mathbf{\Gamma}$ is added to the initial affinity matrix $\mathbf{A}$ without breaking its global behavior~\cite{wang2018non,park2020sumgraph}. $\mathbf{w}_o$ is a projection parameter which has $C'$ output channel. Balance parameter $\gamma$ is set to $1.0$ in our experiments.\vspace{-13pt}

\paragraph{Soft Mask Estimation.}
The mask decoder $\mathcal{F}_m$ takes both the transformed visual features $\mathbf{\hat{V}}$ and corresponding audio features $\mathbf{\bar{S}}$ to generate a soft mask~\cite{wang2018voicefilter}, which filters the mixture spectrogram to produce the enhanced spectrogram.
Audio-visual features are concatenated over the channel dimension to compute an integrated feature $\mathbf{\Psi}= \Pi (\mathbf{\bar{S}}, \mathbf{\hat{V}})$.
In this way, each audio feature is associated with corresponding visual features, which will be used to recover clean speech.
We employ the similar mask decoder architecture used in~\cite{afouras2018conversation} as the residual building block of our decoder.
The sequentially up-scaled output to the original size of the input spectrogram is then passed through sigmoid activation to regularize output values between 0 and 1.
Finally, the estimated speech spectrogram $\mathbf{\hat{Y}} = \mathbf{M} \odot \vert \mathbf{X} \vert$ is computed by element-wise multiplying the estimated mask $\mathbf{M} = \mathcal{F}_m(\mathbf{\Psi})$ on the input spectrogram $\vert \mathbf{X} \vert$.
Then, the estimated speech $\hat{Y}$ is computed by inverse STFT.
We note that the architectural detail of \basemodel is explained in supplementary materials.\vspace{-13pt}

\paragraph{Training.}
The terminal objective of our model is to estimate a target speech $Y$ of an interest person associated visual inputs.
During training, while previous local matching methods assume that audio is correctly aligned to video~\cite{afouras2018conversation,li2020deep,lu2018listen}, we consider that sometimes the audio stream leads the video or sometimes the video stream is going ahead of the audio.
To accomplish these cases in the training scheme, \basemodel leverages a video clip that is recorded a little longer than the randomly sampled audio in the datasets. 
However, there is no additional label on what time the audio will be matched to in the video.

To train \basemodel, we minimize the magnitude loss $\mathcal{L}_\text{MAG}$ that makes the magnitude of enhanced spectrogram be similar to that of clean spectrogram on a logarithmic scale~\cite{chung2020facefilter}:
\begin{equation}
    \mathcal{L}_\text{MAG}(\mathbf{Y}, \mathbf{\hat{Y}}) = \Vert \log(\vert \mathbf{Y} \vert / \vert \mathbf{\hat{Y}} \vert ) \Vert_2.
    \label{eq:mag}
\end{equation}
\vspace{-15pt}

\subsection{Complex Cross-modal Affinity Network}\vspace{-1pt}
In this section, we explain how to further improve the separation ability of \basemodel generating complex ratio mask that considers magnitude and phase simultaneously with simple modifications.
The complex model, \complexmodel, has a similar architecture configuration as that of the \basemodel.
Although using only \basemodel provides satisfactory performance, we upbuild inflated complex \basemodel (\complexmodel) based on complex-valued building blocks~\cite{trabelsi2017deep} to handle complex matrices represented in the spectrograms.
In tasks related to audio signal reconstruction, such as speech enhancement~\cite{choi2019phase} and separation~\cite{lee2017fully}, it is ideal to perform correct estimation of both components.
Details on batch normalization and weight initialization for complex networks can be found in~\cite{trabelsi2017deep,choi2019phase}.

Different from \basemodel, which solely takes the magnitude of spectrogram as input, the audio encoder stream of \complexmodel leverages the whole amount of complex-valued spectrogram to extract the audio embedding feature with stacked complex-valued convolutional layers such that $\mathbf{S} = \mathcal{F}_s^{c}{(\mathbf{X})}$,
where $\mathbf{S} \in \mathbb{R}^{N \times D \times 2}$ contains the real and imaginary parts of a complex number, and $\mathcal{F}_s^{c}$ denotes complex audio encoder.

Inspired by~\cite{choi2019phase}, we decode the corresponding phase along with both the noisy phase and magnitude from the feature representation step.
This solution makes the complex-valued mask $\mathbf{M}$ estimated using the magnitude feature and noise phase feature at the same time.
The noise phase is refined to clean phase with the complex mask decoder $\mathcal{F}_m^c$:
\begin{equation}
    \mathbf{M} = \mathcal{F}_m^c ( \Pi (\mathbf{ \vert \bar{S}} \vert, \mathbf{\hat{V}}) \cdot e^{ i\theta_{\mathbf{\bar{S}}}}).
\end{equation}
The estimated speech spectrogram $\mathbf{\hat{Y}}$ is computed by multiplying the estimated mask $\mathbf{M}$ on the input spectrogram $\mathbf{X}$:
\begin{equation}
    \mathbf{\hat{Y}} = \mathbf{M} \odot \mathbf{X} = \vert \mathbf{M} \vert \cdot \vert \mathbf{X} \vert \cdot e^{i(\theta_{\mathbf{M}}+\theta_\mathbf{X})}.
\end{equation}
Finally, we compute the estimated speech $\hat{Y}$ with inverse STFT.
By inducing complex convolutions in \complexmodel, we use the scale-invariant source-to-distortion ratio (SI-SDR) to the objective function such as
\begin{equation}
\mathcal{L}_\text{SI-SDR}(Y, \hat{Y}) = - <Y, \hat{Y}>/\Vert Y \Vert \Vert \hat{Y} \Vert,
\label{eq:sdr}
\end{equation}
where it makes more phase sensitive, as inverted phase gets penalized as well.
Combining \eqref{eq:mag} and \eqref{eq:sdr}, the final overall objective function in the \complexmodel is given by $\mathcal{L}_\text{ALL} = \mathcal{L}_\text{MAG} + \alpha \mathcal{L}_\text{SI-SDR}$,
where $\alpha$ is a hyper parameter to balance two objective functions and we set it to $1.0$.
\section{Experiments}

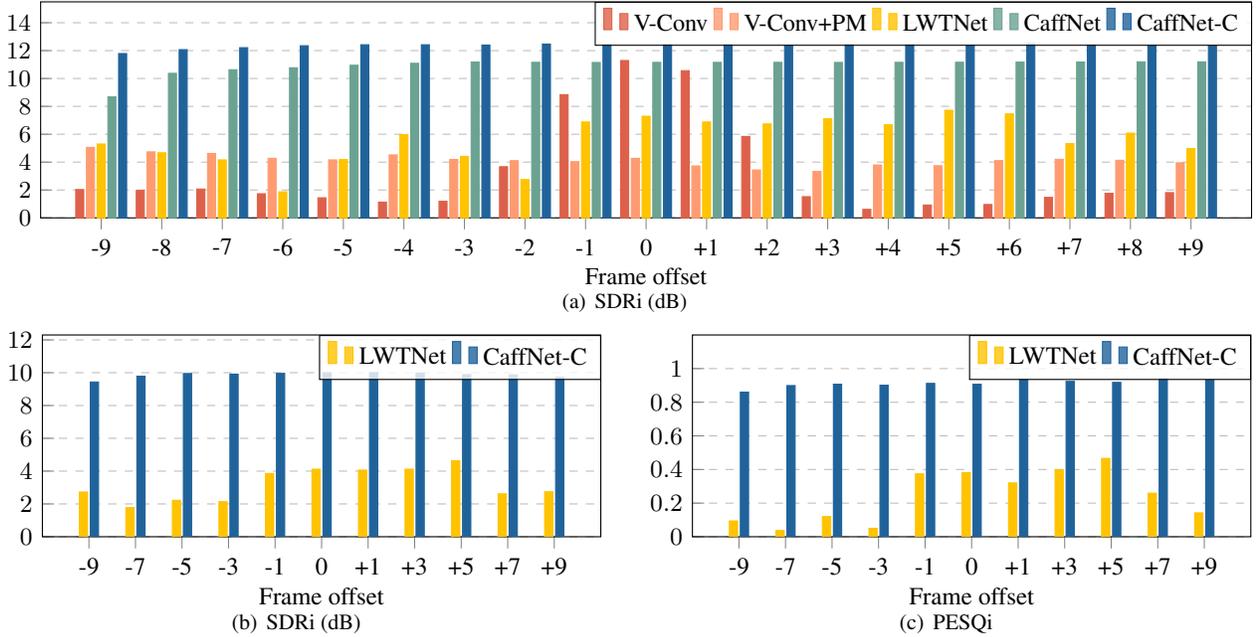
\begin{figure*}[t]
    \centering
    \resizebox{\textwidth}{!}{   
    \begin{tabular}{c}
        \subfigure[\vspace{-10pt}SDRi (dB)]{
        \centering
        \begin{tikzpicture}
        \tikzstyle{every node}=[font=\small]
            \begin{axis}[
                width=\textwidth, 
                height=0.53\columnwidth,
                xtick pos=left, 
                ytick pos=left,
                ybar=1pt,
                bar width=3pt,
                xtick distance=1000,
                xmin=-10, xmax=+10,
                ymin=0,   ymax=15.5,
                y label style={at={(0.1,0.5)},font=\small},
                xlabel={Frame offset\vspace{-10pt}},
                label style = {font=\small},
                ticklabel style = {font=\small},
                xticklabels={-9,-8,-7,-6,-5,-4,-3,-2,-1,0,+1,+2,+3,+4,+5,+6,+7,+8,+9},
                xtick={-9,-8,-7,-6,-5,-4,-3,-2,-1,0,+1,+2,+3,+4,+5,+6,+7,+8,+9},
                ytick={0,2,4,6,8,10,12,14,16,18},
                x tick label style={/pgf/number format/1000 sep=},
                ymajorgrids=true,
                xmajorgrids=false,
                grid style=dashed,
                legend style={
                at={(0.729, 1)},
                anchor=north, legend columns=-1, font=\small,
                fill opacity=0.8, draw opacity=1,text opacity=1},
                ]
                \addplot [style={color1,fill=color1,mark=none}] coordinates{%V-Conv
                (-9, 2.03) (-8, 1.97) (-7, 2.07) (-6, 1.73) (-5, 1.44) (-4, 1.13) (-3, 1.19) (-2, 3.67) (-1, 8.84) (0, 11.29) (+1, 10.56) (+2, 5.84) (+3, 1.52) (+4, 0.61) (+5, 0.92) (+6, 0.96) (+7, 1.47) (+8, 1.76) (+9, 1.81)};
                \addplot [style={color2,fill=color2,mark=none}] coordinates{%V-Conv + PM
                (-9, 5.055) (-8, 4.736) (-7, 4.623) (-6, 4.272) (-5, 4.159) (-4, 4.518) (-3, 4.19) (-2, 4.104) (-1, 4.03) (0, 4.272) (+1, 3.723) (+2, 3.438) (+3, 3.331) (+4, 3.795) (+5, 3.742) (+6, 4.109) (+7, 4.201) (+8, 4.133) (+9, 3.953)};
                \addplot [style={color3,fill=color3,mark=none}] coordinates{%LWTNet
                (-9, 5.30) (-8, 4.67) (-7, 4.16) (-6, 1.87) (-5, 4.19) (-4, 5.97) (-3, 4.41) (-2, 2.75) (-1, 6.88) (0, 7.29) (+1, 6.88) (+2, 6.74) (+3, 7.11) (+4, 6.68) (+5, 7.72) (+6, 7.46) (+7, 5.33) (+8, 6.08) (+9, 4.97)};
                \addplot [style={color4,fill=color4,mark=none}] coordinates{%CaffNet
                (-9, 8.68) (-8, 10.37) (-7, 10.63) (-6, 10.77) (-5, 10.95) (-4, 11.09) (-3, 11.18) (-2, 11.16) (-1, 11.15) (0, 11.157) (+1, 11.156) (+2, 11.159) (+3, 11.153) (+4, 11.160) (+5, 11.166) (+6, 11.175) (+7, 11.185) (+8, 11.196) (+9, 11.196)};
                \addplot [style={color5,fill=color5,mark=none}] coordinates{%CaffNet-C
                (-9, 11.79) (-8, 12.07) (-7, 12.21) (-6, 12.35) (-5, 12.42) (-4, 12.42) (-3, 12.40) (-2, 12.47) (-1, 12.44) (0, 12.47) (+1, 12.46) (+2, 12.4) (+3, 12.47) (+4, 12.47) (+5, 12.41) (+6, 12.52) (+7, 12.43) (+8, 12.36) (+9, 12.35)};
                \legend{V-Conv, V-Conv+PM, LWTNet, \basemodel, \complexmodel}\vspace{-15pt}
            \end{axis}\vspace{-15pt}
        \end{tikzpicture}\label{fig:result_gt_sync}
        }\hfill 
        \end{tabular}
    }
    
    \vspace{-7pt}
    \resizebox{\textwidth}{!}{
        \begin{tabular}{cc}
        \subfigure[SDRi (dB)]{ 
        \centering
        \begin{tikzpicture}
            \begin{axis}[
                width=0.5\textwidth, 
                height=0.5\columnwidth,
                scaled y ticks = false,
                scaled x ticks = false,
                xtick pos=left, 
                ytick pos=left,
                ybar=1pt,
                % interval=0.6,
                bar width=3pt,
                ymin=0,   ymax=12.3,
                y label style={at={(0.07,0.5)}},
                xlabel={Frame offset},
                label style = {font=\small},
                ticklabel style = {font=\small},
                symbolic x coords={-9,-7,-5,-3,-1,0,+1,+3,+5,+7,+9},
                xtick=data,
                xtick distance=100,
                ytick={0,2,4,6,8,10,12,14},
                minor tick length=1ex,
                x tick label style={/pgf/number format/1000 sep=},
                ymajorgrids=true,
                xmajorgrids=false,
                grid style=dashed,
                legend style={at={(0.748, 1)},
                anchor=north, legend columns=-3, font=\small,
                fill opacity=0.8, draw opacity=1,text opacity=1},
                ]
                \addplot [style={color3,fill=color3,mark=none}] coordinates{%LWTNet
                (-9, 2.727) (-7, 1.773) (-5, 2.22) (-3, 2.136) (-1, 3.855) (0, 4.116) (+1, 4.061) (+3, 4.118) (+5, 4.633) (+7, 2.616) (+9, 2.74)};
                \addplot [style={color5,fill=color5,mark=none}] coordinates{%CaffNet-C
                (-9, 9.432) (-7, 9.788) (-5, 9.95) (-3, 9.912) (-1, 9.959) (0, 9.992) (+1, 10.006) (+3, 9.974) (+5, 9.884) (+7, 9.869) (+9, 9.728)};
                \legend{LWTNet, \complexmodel}
            \end{axis}
        \end{tikzpicture}} &
        
        \subfigure[PESQi]{ 
        \centering
        \begin{tikzpicture}
            \begin{axis}[
                width=0.5\textwidth, 
                height=0.5\columnwidth,
                scaled y ticks = false,
                scaled x ticks = true,
                xtick pos=left, 
                ytick pos=left,
                ybar=1pt,
                bar width=3pt,
                ymin=0,   ymax=1.2,
                y label style={at={(0.06,0.5)}},
                xlabel={Frame offset},
                label style = {font=\small},
                ticklabel style = {font=\small},
                symbolic x coords={-9,-7,-5,-3,-1,0,+1,+3,+5,+7,+9},
                xtick=data,
                xtick distance=10,
                ytick={0,0.2,0.4,0.6,0.8,1.0},
                minor tick length=1ex,
                x tick label style={/pgf/number format/1000 sep=},
                ymajorgrids=true,
                xmajorgrids=false,
                grid style=dashed,
                legend style={at={(0.748, 1)},
                anchor=north, legend columns=-3, font=\small,
                fill opacity=0.8, draw opacity=1,text opacity=1},
                ]
                \addplot [style={color3,fill=color3,mark=none}] coordinates{%LWTNet
                (-9, 0.094) (-7, 0.037) (-5, 0.12) (-3, 0.05) (-1, 0.374) (0, 0.381) (+1, 0.32) (+3, 0.398) (+5, 0.466) (+7, 0.259) (+9, 0.142)};
                \addplot [style={color5,fill=color5,mark=none}] coordinates{%CaffNet-C
                (-9,0.86) (-7, 0.899) (-5, 0.907) (-3, 0.901) (-1, 0.912) (0, 0.907) (+1, 0.937) (+3, 0.924) (+5, 0.918) (+7, 0.94) (+9, 0.931)};
                \legend{LWTNet, \complexmodel}
            \end{axis}
        \end{tikzpicture}
        }
        \end{tabular}
        }
    \vspace{-7pt}
    \caption{Evaluation of AVSS performance with respect to each delay offset between audio and visual streams on LRS2 dataset. The frame offset unit is 40ms which is the duration length between consecutive video frames. (a) reports the SDRi evaluation using ground-truth phase with estimated magnitude spectrum. (b) and (c) report the SDRi and PESQi evaluation results on the predicted phase as well as estimated magnitude spectrum, respectively.}
    \label{fig:result_synchronisation}\vspace{-8pt}
\end{figure*}

\subsection{Setup}\vspace{-1pt}
\paragraph{Datasets.}
Our networks are evaluated on three commonly used AVSS benchmarks: Lip Reading Sentences 2 (LRS2)~\cite{chung16lip,chung2017lip,chung17Alip}, Lip Reading Sentences 3 (LRS3)~\cite{lrs3}, and VoxCeleb2~\cite{voxceleb2} datasets.
LRS2 and LRS3 include 224 and 475 hours of videos respectively, along with cropped face tracks of the speakers.
While LRS2 is sourced from British television broadcasts, LRS3 contains TED and TEDx videos.
Following~\cite{afouras2019my}, we remove the few speakers from the LRS3 training set that also appear in the test set, so that there is no overlap of identities between the two subsets. 
Hence, the test set includes only unseen and unheard speakers during training and is suitable for a speaker-agnostic evaluation of our methods.
VoxCeleb2 contains over 1 million utterances spoken by 5,994 speakers.
They provide the pre-train set and test set, and we follow this setting in our experiments.
It is assumed that all datasets are well-synchronized~\cite{afouras2018conversation,afouras2019my}, so we adapt them to our purposes by augmenting data.
Furthermore, VoxCeleb2 is divided into training and test sets according to the speaker's identity to explicitly assess whether our model can generalize to unseen speakers during training.
Thus training and testing sets are disjoint in terms of speaker's identities.\vspace{-13pt}

\paragraph{Data Sampling Protocol.}
% \paragraph{Training \& Evaluation Protocol.}
As mentioned in~\secref{sec:method}, we premise that audio-visual data is obtained in asynchronous circumstances.
We assume that any given frame possesses the same time shift, so the visual stream is randomly shifted by $-9$ to $9$ frames.
Although we randomly shift video frames to assume the discontinuities of training data, the corresponding video clips contain all the audio-response information.
For example, if the audio is sampled in the time duration $[T, T+\delta]$ and the video is shifted by $-9$ frames,
\ie 0.36s, we extract the video frames during the time duration $[T-0.36, T+\delta]$.
If the selected time offset is $9$, the video frames are extracted within the time duration $[T, T+\delta+0.36]$.
For consistency and fair evaluation, we follow the similar evaluation settings in the previous work~\cite{afouras2018conversation}.
To generate synthetic training examples, we first select a source pair consisting of visual and audio features by sampling 2 seconds randomly.
Source speech is mixed with randomly selected other speaker's speech signal in the time domain, to simulate multi-talker backgrounds signals.\vspace{-13pt}

\paragraph{Features.}
We use a recent audio-visual synchronization model~\cite{chung2019perfect} for extracting visual features to serve as its visual input.
The input to the visual stream is a video of cropped facial frames, with a frame rate of 25 fps. 
For every video frame, it outputs a compact 512-dimensional feature vector.
For audio features, we use a time-frequency representation via STFT with a 25ms window length and a 10ms hop length as a sampling rate of 16kHz.
Note that the extraction of face embeddings follows prior work~\cite{afouras2020self}.
\vspace{-13pt}

\begin{table*}[!th]
     \centering
     \resizebox{0.9\linewidth}{!}{
     \begin{tabular}{c|l| cccc | cccc | cccc}
            \hlinewd{0.8pt}
            \multirow{2}{*}{Dataset} & \multirow{2}{*}{Method} & \multicolumn{4}{c|}{SDRi $\uparrow$} & \multicolumn{4}{c|}{PESQi $\uparrow$} & \multicolumn{4}{c}{STOI $\uparrow$}  \tabularnewline   \cline{3-14}
            & & GT & GL & PR & MX & GT & GL & PR & MX & GT & GL & PR & MX   \tabularnewline \hline \hline
            \multirow{4}{*}{LRS2}
            & V-Conv~\cite{afouras2018conversation} & 11.28 & -4.36 & - & 6.73 & 1.35 & 0.63 & - & 0.75 & 0.89 & 0.85 & & 0.86 \tabularnewline 
            & LWTNet~\cite{afouras2020self} & 6.88 & -4.61 & 4.06 & 3.77 & 0.65 & 0.16 & 0.32 & 0.29 & 0.77 & 0.72 & 0.73 & 0.74\tabularnewline
            & \textbf{\basemodel(ours)} & 11.16 & -3.49 & - & 6.79 & 1.29 & 0.63 & - & 0.73 & 0.89 & 0.85 & - & 0.86 \tabularnewline
            & \textbf{\complexmodel (ours)} & 12.46 & -2.54 & 10.01 & 7.94 & 1.15 & 0.65 & 0.94 & 0.73 & 0.89 & 0.84 & 0.88 & 0.86\tabularnewline
            \hline 
            \multirow{4}{*}{LRS3}
            & V-Conv~\cite{afouras2018conversation} & 11.23 & -1.37 & - & 7.00 & 1.08 & 0.55 & - & 0.61 & 0.86 & 0.82 & - & 0.83 \tabularnewline 
            & LWTNet~\cite{afouras2020self} & 7.71 & -3.93 & 4.83 & 4.44 & 0.82 & 0.35 & 0.49 & 0.45 & 0.84 & 0.80 & 0.82 & 0.82\tabularnewline
            & \textbf{\basemodel(ours)} & 10.22 & -2.64 & - & 6.46 & 1.06 & 0.49 & - & 0.60 & 0.86 & 0.82 & -& 0.84  \tabularnewline
            & \textbf{\complexmodel (ours)} & 12.31 & -1.38 & 9.78 & 7.92 & 0.91 & 0.49 & 0.71 & 0.55 & 0.86 & 0.82 & 0.85 & 0.83 \tabularnewline
            \hlinewd{0.8pt}
            \end{tabular}
            }
            \vspace{-6pt}
      \caption{Evaluation of AVSS performance on the LRS2 and LRS3 datasets when audio and visual inputs are in synchronous condition. 
      Contrary to other methods~\cite{afouras2020self,afouras2018conversation}, \basemodel and \complexmodel are trained in unconditioned circumstance, \ie training with randomly given frame offsets. {GT}: ground-truth phase; {GL}: Griffin-Lim; {PR}: predicted phase; {MX}: mixture phase.}
      \label{tab:1}\vspace{-10pt}
\end{table*}

\paragraph{Evaluation Metrics.}
We use three metrics to compare the results of our method to previous methods~\cite{afouras2018conversation,afouras2020self}.
First, the signal-to-distortion ratio (SDR)~\cite{vincent2006performance} is commonly used metric in recent works~\cite{afouras2018conversation,afouras2019my,owens2018audio} to investigate the quality of enhanced speech.
Following the previous work~\cite{afouras2018conversation}, we also report results on the perceptual evaluation of speech quality (PESQ)~\cite{rix2001perceptual} varying from -0.5 to 4.5 and the short-time objective intelligibility (STOI)~\cite{taal2011algorithm}, which is correlated with the intelligibility of degraded speech signals.
In the following experiments, we report SDR improvement (SDRi) and PESQ improvement (PESQi) for a fair comparison with other methods since the testing samples are randomly generated by combinations of the test set.\vspace{-13pt}

\paragraph{Baseline Models.}
For the fair comparison with \basemodel, we reproduce the magnitude network of `V-Conv'~\cite{afouras2018conversation}, which is trained with the magnitude loss only.
Also, since V-Conv model only assumes the synchronization circumstances between audio and visual streams, we combine V-Conv with a cutting-edge synchronization method proposed in~\cite{chung2019perfect,chung2020perfect} to deal with asynchronous samples, referred to as 'V-Conv+PM'.
Furthermore, we examine the performance of LWTNet~\cite{afouras2020self} using delayed test samples, where this method includes an independent synchronization module.

\subsection{Results}\vspace{-1pt}
\paragraph{LRS2 and LRS3.} 
In \figref{fig:result_synchronisation}, the proposed models, \basemodel and \complexmodel, show robust performance in asynchronized environment while the previous methods have significant degradation in performance when the two frames were delayed.
Although V-Conv+PM system has the synchronization step before the separation, it is clear that our methods are more robust to these temporal shifts.
Because V-Conv+PM is a cascaded-step system, errors in the first step have a negative impact on the second.
For that reason, V-Conv+PM even shows less effective results, even compared to V-Conv when audio and visual streams are synchronized.
Furthermore, despite the alignment step in LWTNet~\cite{afouras2020self}, its accuracy is poorer than our methods on delayed samples.

When the two streams are well-aligned, our method provides competitive performance compared to baseline methods.
In \tabref{tab:1}, we summarize the comparison results on the LRS2 and LRS3 datasets without a random shift on video streams (\ie, frame offset is 0).
Even though our network is trained in an unconditioned environment where audio and visual streams are not synchronized, our method outperforms existing methods~\cite{afouras2018conversation, afouras2020self} which are trained with the synchronized audio-visual streams.

To validate the generality of our models, we investigate the performance on the LRS3 dataset, summarized in~\tabref{tab:1}.
Each model is only trained on the LRS2 dataset and evaluated on the LRS3 dataset without an additional adaptation process.
Overall performances are similar to those on the LRS2 dataset.
\complexmodel achieves the best SDRi of $12.31$ and still outperforms all the others, where the reconstructions are obtained using the magnitudes predicted by our network and either the ground truth phase. 
This demonstrates that our methods can be generalized to other datasets.
Furthermore, the results show that phase estimation helps reconstruct the magnitude.
Compared to \basemodel, \complexmodel improves the SDRi by 2.09dB using ground-truth phase on the LRS3 dataset, respectively.\vspace{-13pt}

\paragraph{VoxCeleb2.}
% We evaluated the generality of our models over languages on the VoxCeleb2 dataset.
In order to explicitly investigate whether our model can be generalized to speakers unseen during training, we fine-tune and test on the VoxCeleb2 in~\tabref{tab:vox2}. 
The training and test sets are disjoint in terms of speaker identities.
We evaluate performance based on gender combinations, as many previous speech separation methods have shown significant performance degradation when mixtures involve same-gender speech~\cite{hershey2016deep,chung2020facefilter}.
Although our method shows slight drops in performance on the female-female set, the result on the male-male set is similar to those of subsets including different genders. 

\begin{table}[t]
\centering
\resizebox{\linewidth}{!}{
            \begin{tabular}{llcccc}
            \hlinewd{0.8pt}
            \multirow{2}{*}{Speaker} & Mixture & \multicolumn{4}{c}{Phase estimation} 
            \tabularnewline 
            \cline{3-6}
            & (Source-Reference) & GT & GL & PR & MX \tabularnewline
            \hline
            \hline
            \multirow{4}{*}{Seen} 
            & Male-Female   & 7.35 & -5.46 & 3.54 & 3.33  \tabularnewline
            & Female-Male   & 7.51 & -3.63 & 3.99 & 3.71   \tabularnewline
            & Male-Male     & 8.21 & -4.61 & 4.24 & 3.91   \tabularnewline
            & Female-Female & 7.24 & -3.41 & 3.91 & 3.58   \tabularnewline
            \hline
            \multirow{4}{*}{Unseen} 
            & Male-Female   & 8.22 & -5.12 & 4.80 & 4.42 \tabularnewline
            & Female-Male   & 7.34 & -3.42 & 3.99 & 3.71   \tabularnewline
            & Male-Male     & 7.35 & -5.46 & 3.55 & 3.33 \tabularnewline
            & Female-Female & 6.37 & -4.23 & 3.15 & 2.92   \tabularnewline
            \hlinewd{0.8pt}
            \end{tabular}
           }
            \vspace{-6pt}
      \caption{Evaluation of SDRi on \complexmodel with regarding to gender combinations on the VoxCeleb2 dataset}
      \label{tab:vox2}\vspace{-10pt}
\end{table}

\subsection{Analysis}\vspace{-1pt}
We conduct extensive experiments to describe the contribution of our method.
Note that all experiments in this section are performed on the LRS2 dataset with \complexmodel.\vspace{-13pt}

\begin{figure}[t]
\begin{center}
    \subfigure[-9 offset (12.2dB)]{\includegraphics[width=0.32\linewidth]{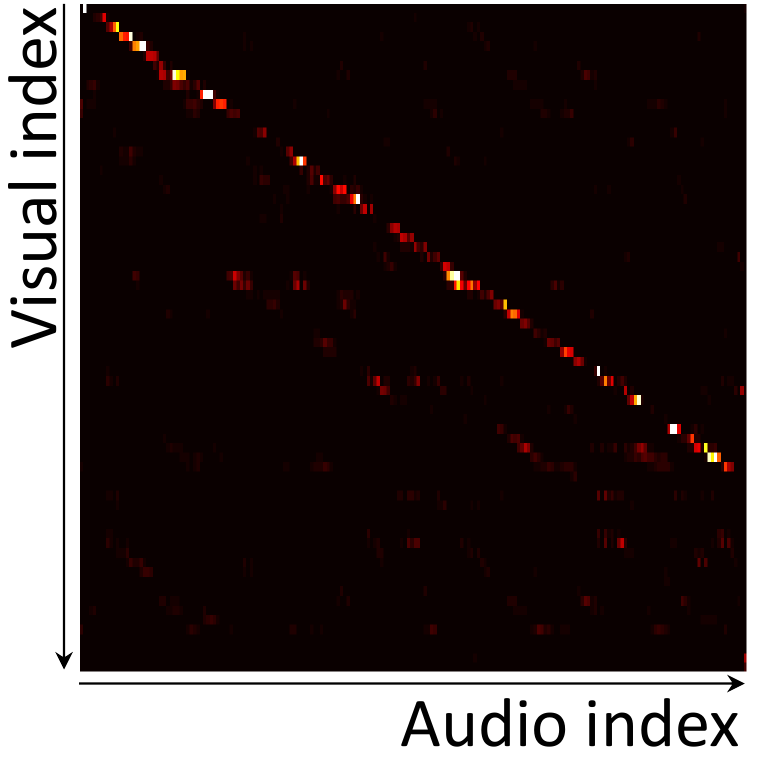}}\hfill
    \subfigure[0 offset (12.8dB)]{\includegraphics[width=0.32\linewidth]{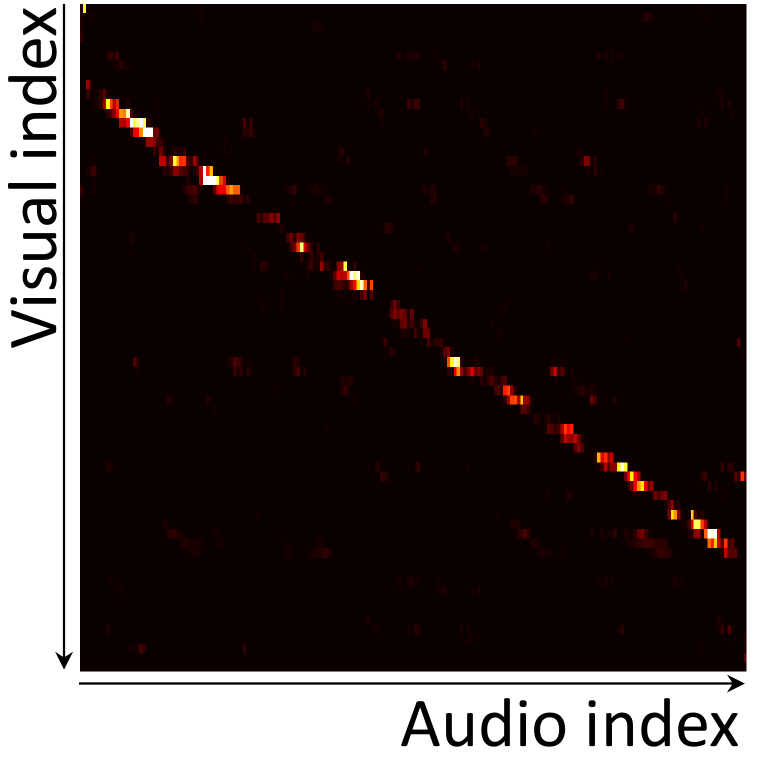}}\hfill 
    \subfigure[+9 offset (12.6dB)]{\includegraphics[width=0.32\linewidth]{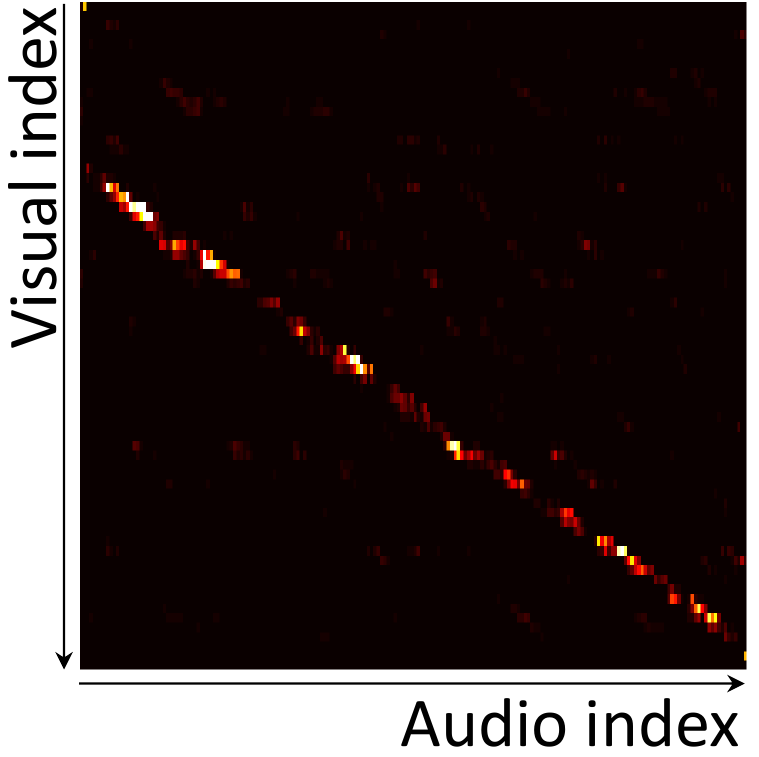}}\hfill \\ \vspace{-10pt}
    \caption{Qualitative results of affinity matrices according to the frame delay offsets. If the frame offset is negative, a linear pattern along the diagonal of the affinity matrix is displaced higher than zero offset case, and if it is positive, the pattern is shown lower than (b). ($\cdot$) is SDR.}
    \label{fig:analysis_delay}\vspace{-10pt}
\end{center}
\end{figure}

\paragraph{Channel Latency.}
Our premise is that temporal alignment of cross-modal can be achieved in our networks without additional supervision for ground-truth mapping between each pair of input streams.
As shown in~\figref{fig:analysis_delay}, \complexmodel surprisingly finds appropriate frame offsets without any extra supervision.
It demonstrates that our method reliably infers clean speech with well-aligned visual streams.
Although we obtain the highest SDR in the synchronous setting, there are only slight differences between the performance despite delays in audio and video.\vspace{-13pt}

\begin{table}[t]
     \centering
     \resizebox{0.85\linewidth}{!}{
            \begin{tabular}{cll ccc}
            \hlinewd{0.8pt}
            \multirow{2}{*}{AR} & \multirow{2}{*}{Magnitude} & \multirow{2}{*}{Phase} & \multicolumn{3}{c}{Delay offset} 
            \tabularnewline 
            \cline{4-6}
            & & & -5 & 0 & 5 \tabularnewline
            \hline
            \hline
            \xmark & Prediction & GT & 8.01 & 7.77 & 7.71 \tabularnewline
            \xmark & Prediction & GL & -3.93 & -4.24 & -3.96 \tabularnewline
            \xmark & Prediction & MX & 4.37 & 3.99 & 3.86 \tabularnewline
            \xmark & Prediction & PR &  5.13 & 4.75 & 4.62 \tabularnewline
            \hline
            \cmark & Prediction & GT & 12.42 & 12.48 & 12.41 \tabularnewline
            \cmark & Prediction & GL & -3.05 & -2.67 & -2.54 \tabularnewline
            \cmark & Prediction & MX & 7.96 & 7.97 & 7.85 \tabularnewline
            \cmark & Prediction & PR & 9.95 & 9.92 & 9.88 \tabularnewline
            \hlinewd{0.8pt}
            \end{tabular}
      }\vspace{-6pt}
      \caption{Evaluation of SDRi to demonstrate the effectiveness of affinity regularization. `AR' refers to the presence or absence of regularization.}
      \label{tab:analysis_reg}\vspace{-10pt}
\end{table}

\paragraph{Affinity Regularization.}
One might also ask whether affinity regularization of our method is helpful.
To verify its efficacy, we conducted an ablation study with \complexmodel in~\tabref{tab:analysis_reg}.
Since the affinity regularization has induced the global correspondence term for reliable speech separation, SDRi increases about 4 dB.
It provides reasonable evidence for the effects of the affinity regularization.\vspace{-13pt}

\begin{figure}[!t]
\begin{center}
    \hfill
    \subfigure[With jitter]{\includegraphics[width=0.32\linewidth]{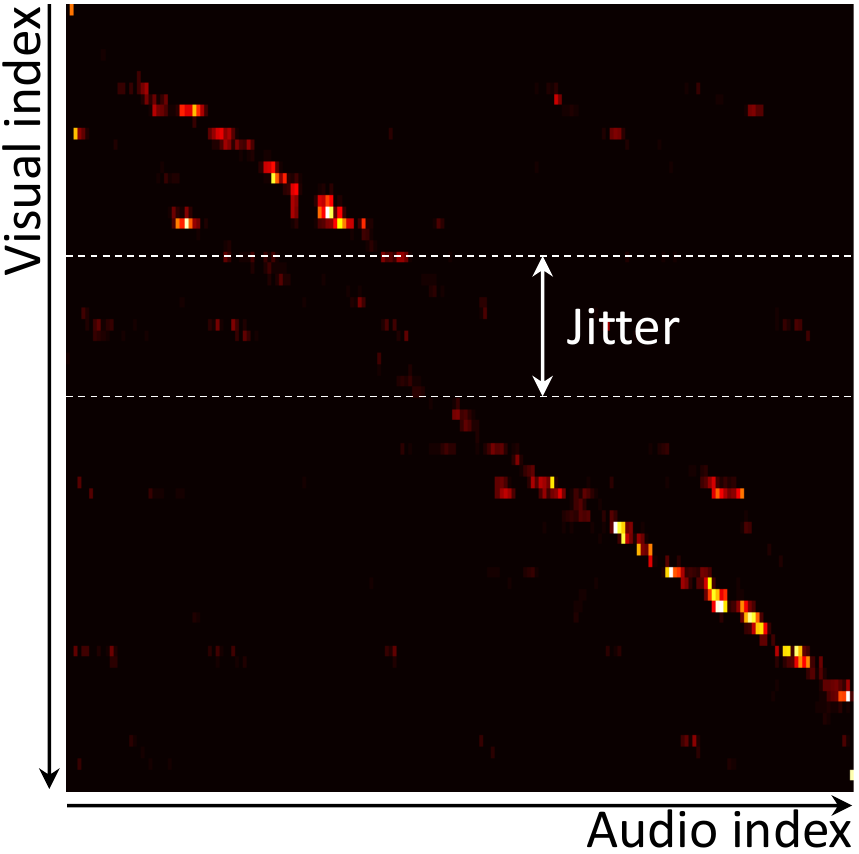}}\hfill
    \subfigure[Without jitter]{\includegraphics[width=0.32\linewidth]{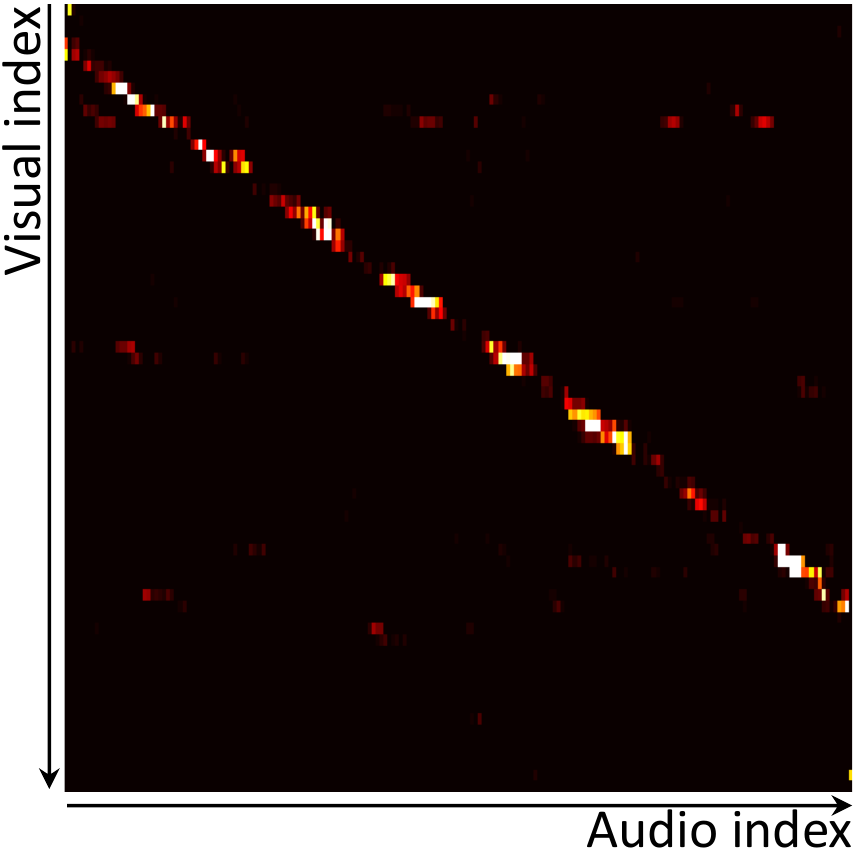}}\hfill
    \vspace{-10pt}
    \caption{Example of affinity matrices with jitter problem. In this case, the jitter is happen for 0.4s on the 25th frame, thus diagonal components are disconnected and reappeared from the 35th frame.}
    \label{fig:jitter}\vspace{-13pt}
\end{center}
\end{figure}

\paragraph{Jitter.}
We then assessed the impact of our method from jitter, another well-known discontinuity problem frequently observed due to independent packetization of the audio-video streams.
As a proof of concept, we assumed that no video frames are transmitted from $t$ to $t$+$\tau$ frames, where $\tau$ is chosen as a random integer number such that $\tau\leq8$ ($\approx$ 0.3 s).
Considering a simple frame repetition method that can be used in the jitter situation, we replaced the missing video frames with the ($t-1$)-th frame.
\complexmodel shows compliant performance with 5.64 dB even in these challenging situations, in terms of SDRi on the testing set of LRS2 dataset.
As exemplified in~\figref{fig:jitter}, it shows that our network clearly distinguishes two global terms in the affinity matrix when the frame jitter arises in the visual stream.\vspace{-13pt}

\begin{table}[t]
     \centering
     \resizebox{0.9\linewidth}{!}{
            \begin{tabular}{ll cccc}
            \hlinewd{0.8pt}
            \multirow{2}{*}{Magnitude} & \multirow{2}{*}{Phase} & \multicolumn{3}{c}{Delay offset} & \multirow{2}{*}{Avg. $\downarrow$}
            \tabularnewline 
            \cline{3-5}
            & & -5 & 0 & 5 \tabularnewline
            \hline
            \hline
            Mixture & MX & - & 84.91 & - & 84.91 \tabularnewline
            Ground-truth & GT & - & 17.74 & - & 17.74 \tabularnewline \hline
            Prediction & GT & 31.70 & 32.96 & 31.90 & 32.18 \tabularnewline
            Prediction & GL & 40.54 & 39.38 & 39.74 & 39.88\tabularnewline
            Prediction & MX & 38.35 & 37.01 & 37.73 & 37.69\tabularnewline
            Prediction & PR &  35.88 & 35.08 & 34.83 & {35.26}\tabularnewline
            \hlinewd{0.8pt}
            \end{tabular}
      }\vspace{-6pt}
      \caption{Automatic speech recognition on the LRS2 dataset. 
      }
      \label{tab:asr}\vspace{-10pt}
\end{table}

\paragraph{Speech Recognition.}
To verify the intelligibility of the outputs, we further exploit the estimated speech signals on another task.
Specifically, we conduct an additional experiment for automatic speech recognition with enhanced speech signals.
To do this, we utilize the speech recognition API of Google cloud system~\footnote{https://cloud.google.com/speech-to-text} and compute the word error rate (WER) as an automatic metric to evaluate the accuracy of recognition.
Firstly, we obtain the WER of $17.74\%$ on the clean ground-truth set, which is the best result that we can achieve in this setup, while the WER on the mixture set is $84.95\%$.
In~\tabref{tab:asr}, \complexmodel achieves $35.26\%$ error rate when we use separated speech signals with the network setup of phase prediction~(PR). It clearly shows that there is no meaningful difference by varying delay offset.

\section{Conclusion and Future Work}
We presented a novel framework to separate a target speaker's speech from audio in the wild. 
To deal with the local matching problem in AVSS, we effectively established a cross-modal affinity between a pair of audio-visual features by modeling relative timing dependencies without supervision.
We further suggested a way of affinity regularization to employ global alignment while addressing the label permutation problem.
Moreover, we generalized the proposed networks to the complex domain to reconstruct both magnitude and phase masks, enhancing the separation performance.
A direction for future work is to explore applications of this generic framework which can be widely plugged into many audio-visual systems.
\blfootnote{
This research was supported by the Yonsei University Research Fund~(Yonsei Signature Research Cluster Program).}

{\small
\bibliographystyle{ieee_fullname}
\bibliography{egbib}
}

\newpage

\appendix
\section{Appendix}

In this document, we describe architecture details of \basemodel and \complexmodel, details of implementation and training, and provide more quantitative and qualitative results on LRS2, LRS3, and VoxCeleb2 datasets.
\subsection{Network}

\paragraph{Complex-valued Networks}
Here we present a brief review of complex-valued networks, which can handle complex computations in deep networks~\cite{choi2019phase,lee2017fully,trabelsi2017deep}.
The complex-valued convolution operation on the intermediate feature representation $\mathbf{h}=\mathbf{h}_{r}+i\mathbf{h}_{i}$ with a complex-valued convolutional filter $\mathbf{w} = \mathbf{w}_{r}+i\mathbf{w}_{i}$:
\begin{equation}
\mathbf{w} * \mathbf{h} = (\mathbf{h}_{r}*\mathbf{x}_{r} - \mathbf{h}_{i}*\mathbf{w}_{i}) + i(\mathbf{h}_{r}*\mathbf{w}_{i} + \mathbf{h}_{i}*\mathbf{w}_{r}),
\end{equation}
where $\mathbf{w}_{r},\mathbf{w}_{i}$ are real-valued matrices of filter and $\mathbf{h}_{r}, \mathbf{h}_{r}$ are real-valued matrices of complex feature representation.
In practice, complex convolutions can be implemented as two different real-valued convolution operations with shared real-valued convolution filters as follows~\cite{choi2019phase}, and activation functions like ReLU were also adapted to the complex domain.
To establish the complex-valued convolutional layer in our networks, we modified 2D complex-valued convolutional layers~\cite{choi2019phase} into 1D complex-valued convolutional layers.

\paragraph{Network Architecture}
Our networks are based on V-Conv~\cite{afouras2018conversation} architecture, which consists of encoder-decoder architecture.
For the audio-visual encoder, it contains two coupled encoders; an audio encoder and a visual encoder.
In~\tabref{tab:caffnet_configuration}, \basemodel follows basic configuration of V-Conv~\cite{afouras2018conversation}.
Different from that, the affinity module is added for audio-visual fusion considering global and local correspondence between cross-modal streams. 
As demonstrated in~\tabref{tab:caffnetc_configuration}, \complexmodel leverages complex-valued convolutional layers ($\mathbb{C}\text{onv}$) in audio encoder and mask decoder.
For the simplicity, we denote $\Pi(\mathcal{E}_{f}{(I^{1:5})}, \cdots, \mathcal{E}_{f}{(I^{T\text{-}4:T})})$ as $\mathcal{E}_{f}(I)$ in \tabref{tab:caffnet_configuration} and \tabref{tab:caffnetc_configuration}.

\begin{table*}[t!]
	\centering
        \begin{tabular}{c|l cccccccc}
			\hlinewd{0.8pt}
% 			\multicolumn{6}{c}{Visual Encoder} \tabularnewline
% 			\hlinewd{0.8pt}
			Module & Layer & K. & Ch. I/O & S. & P. & B.N. & Act. & Input & Output \tabularnewline
			\hline
			\hline
			\multirow{3}{*}{Visual Encoder}
			& V-Conv1D   & 5 & 512/1536 & 1 & 2 & \cmark & ReLU & $\mathcal{E}_{f}(I)$ &  v-feat1 \tabularnewline
			& V-Conv1D ($\times$8) & 5 & 1536/1536 & 1 & 2 & \cmark & ReLU &  v-feat1 & v-feat9 \tabularnewline
			& V-Conv1D  & 5 & 1536/1536 & 1 & 2 & - & - & v-feat9 & $\mathbf{V}$ \tabularnewline
			\hline
			\multirow{4}{*}{Audio Encoder}
			& Decompose & - & 257/257 & - & - & - & - & $\mathbf{X}$ & $\vert \mathbf{X} \vert$ \tabularnewline
			& A-Conv1D   & 5 & 257/1536 & 1 & 2 & \cmark & ReLU & $\vert \mathbf{X} \vert$ & a-feat1 \tabularnewline
			& A-Conv1D ($\times$3) & 5 & 1536/1536 & 1 & 2 & \cmark & ReLU & a-feat1 & a-feat4 \tabularnewline
			& A-Conv1D   & 5 & 1536/1536 & 1 & 2 & - & - & a-feat4 & $\mathbf{S}$ \tabularnewline
			\hline
			\multirow{4}{*}{Affinity Module}
			& V-Nonlocal ($\times$2) & 5 & 1536/1536 & 1 & 2 & \cmark & ReLU & $\mathbf{V}$ & $\mathbf{\bar{V}}$ \tabularnewline
			& A-Nonlocal ($\times$2) & 5 & 1536/1536 & 1 & 2 & \cmark & ReLU & $\mathbf{S}$ & $\mathbf{\bar{S}}$ \tabularnewline
			& AV-Aff.       & - & 1536/1536 & - & - & - & - & $\mathbf{\bar{S}}$, $\mathbf{\bar{V}}$ & $\mathbf{\hat{V}}$ \tabularnewline
			& Concat. ($\Pi$) & - & 1536/3072 & - & - & - & - & $\mathbf{\bar{S}}$, $\mathbf{\hat{V}}$ & $\mathbf{\Psi}$ \tabularnewline
			\hline
			\multirow{4}{*}{Mask Decoder}
			& Conv1D & 5 & 3072/1536 & 1 & 2 & \cmark & ReLU & $\mathbf{\Psi}$ & m-feat1 \tabularnewline
			& Conv1D ($\times$13) & 5 & 1536/1536 & 1 & 2 & \cmark & ReLU & m-feat1 & m-feat14 \tabularnewline
			& Conv1D & 5 & 1536/257 & 1 & 2 & \cmark & Sigmoid & m-feat14 & $\mathbf{M}$ \tabularnewline
			& Mult. ($\odot$) & - & - & - & - & - & - & $\vert \mathbf{X} \vert $, $\mathbf{M}$ & $\mathbf{\hat{Y}}$ \tabularnewline
			\hlinewd{0.8pt}
		\end{tabular}\vspace{-7pt}
% 	}
	\caption{Network configuration of \basemodel. ‘K’, ‘S’ and ‘P’ represents the kernel, stride, and padding size of convolution layer, and ‘Ch. I/O’ represents channels of input and output relative to the input, respectively. 'B.N' is the batch normalization layer and 'Act.' is the activation function. }
	\label{tab:caffnet_configuration}
\end{table*}

\begin{table*}[t!]
	\centering
		\begin{tabular}{c|l cccccccc}
			\hlinewd{0.8pt}
% 			\multicolumn{6}{c}{Visual Encoder} \tabularnewline
% 			\hlinewd{0.8pt}
			Module & Layer & K. & Ch. I/O & S. & P. & B.N. & Act. & Input & Output \tabularnewline
			\hline
			\hline
			\multirow{3}{*}{Visual Encoder}
			& V-Conv1D & 5 & 512/1536 & 1 & 2 & \cmark & ReLU & $\mathcal{E}_{f}(I)$  & v-feat1 \tabularnewline
			& V-Conv1D ($\times$8) & 5 & 1536/1536 & 1 & 2 & \cmark & ReLU &  v-feat1 & v-feat9 \tabularnewline
			& V-Conv1D & 5 & 1536/1536 & 1 & 2 & - & - & v-feat9 & $\mathbf{V}$ \tabularnewline
			\hline
			\multirow{4}{*}{Audio Encoder}
			& A-$\mathbb{C}$onv1D & 5 & 257/1536 & 1 & 2 & \cmark & LReLU & $\mathbf{X}$ & a-feat1 \tabularnewline
			& A-$\mathbb{C}$onv1D ($\times$3) & 5 & 1536/1536 & 1 & 2 & \cmark & LReLU & a-feat1 & a-feat4 \tabularnewline
			& A-$\mathbb{C}$onv1D & 5 & 1536/1536 & 1 & 2 & - & - & a-feat4 & $\mathbf{S}$ \tabularnewline
			& Decompose & - & - & - & - & - & - & $\mathbf{S}$ & $\vert \mathbf{S} \vert$, $e^{i \theta_{\mathbf{S}}}$\tabularnewline
			\hline
			\multirow{5}{*}{Affinity Module}
			& V-Nonlocal ($\times$2) & 5 & 1536/1536 & 1 & 2 & \cmark & ReLU & $\mathbf{V}$ & $\mathbf{\bar{V}}$ \tabularnewline
			& A-Nonlocal ($\times$2) & 5 & 1536/1536 & 1 & 2 & \cmark & ReLU & $\vert \mathbf{S} \vert$ & $\vert \mathbf{\bar{S}} \vert$ \tabularnewline
			& AV-Aff. & - & 1536/1536 & - & - & - & - & $\vert \mathbf{\bar{S}} \vert$, $\mathbf{\bar{V}}$ & $\mathbf{\hat{V}}$ \tabularnewline
			& Concat. ($\Pi$) & - & 1536/3072 & - & - & - & - & $\vert \mathbf{\bar{S}} \vert$, $\mathbf{\hat{V}}$ & $\vert \mathbf{\Psi} \vert $ \tabularnewline
			& Reconstruct & - & - & - & - & - & - & $\vert \mathbf{\Psi} \vert $, $e^{i \theta_{\mathbf{S}}}$ & $\mathbf{\Psi}$\tabularnewline
			\hline
			\multirow{4}{*}{Mask Decoder}
			& $\mathbb{C}$onv1D & 5 & 3072/1536 & 1 & 2 & \cmark & LReLU & $\mathbf{\Psi}$ & m-feat1 \tabularnewline
			& $\mathbb{C}$onv1D ($\times$13) & 5 & 1536/1536 & 1 & 2 & \cmark & LReLU & m-feat1 & m-feat14 \tabularnewline
			& $\mathbb{C}$onv1D & 5 & 1536/257 & 1 & 2 & - & Tanh & m-feat14 & $\mathbf{M}$ \tabularnewline
			& Mult. ($\odot$) & - & - & - & - & - & - & $\mathbf{X} $, $\mathbf{M}$ & $\mathbf{\hat{Y}}$ \tabularnewline
			\hlinewd{0.8pt}
		\end{tabular}\vspace{-7pt}
% 	}
	\caption{Network configuration of \complexmodel. ‘K’, ‘S’ and ‘P’ represents the kernel, stride, and padding size of convolution layer, and ‘Ch. I/O’ represents channels of input and output relative to the input, respectively. 'B.N' is the batch normalization layer and 'Act.' is the activation function. ‘LReLU’ indicates LeakyReLU function with a slope 0.2. $\mathbb{C}$onv denotes complex-valued convolutional layer. At the last layer of mask decoder, we adopt tanh activation for the magnitude of the estimated mask.}
	\label{tab:caffnetc_configuration}
\end{table*}

\subsection{Implementation and Training Details}

\paragraph{Visual Feature Extraction.}
Visual features for audio-visual speech separation represent implicit information correspondent to target speech, \ie linguistic representation on lip movements, called \textit{viseme}.
We train the feature extractor using the strategy of audio-to-video synchronization in self-supervision.
In~\cite{chung16lip,chung2019perfect}, they build two-stream networks to embed audio and visual sequences onto a latent space.
In their strategies, when input audio and visual segments are taken from the same timestamps, the distance between audio-visual embeddings is minimized, whereas it is maximized for the segments from different offsets.
As a result, the embedding learns linguistic information commonly existing on speech sound and lip movements.
Especially, we adopt $M$-way matching method proposed in~\cite{chung2019perfect,chung2020perfect} for the representation learning by providing multiple negative samples over a single positive input,
\begin{equation}
\label{eq:xent}
    \begin{gathered}
        E=-\frac{1}{2N}\sum_{n=1}^{N}\sum_{m=1}^{M}y_{n,m}\text{log}(p_{n,m})\\
        p_{n,m}=\frac{\exp(d^{\minus 1}_{n,m})}{\sum_{m=1}^{M}\exp(d^{\minus 1}_{n,m})}
    \end{gathered}
\end{equation}
where $d_{n,m}=||v_{n,m}-a_{n,m}||_2$ is similarity distance between embedding pairs and $y_{n,m}\in\{0,1\}$ denotes similarity label.
In this experiment, we stacked 5 consecutive visual frames of $224\times224$ pixels with RGB channels as the visual input, and 20 audio frames as the audio input, where they are 0.2-seconds length.
The architecture configuration is same as described in~\cite{chung2019perfect}.
It shows powerful performance compared to using contrastive loss, and we prepare two visual feature extractors; one is for LRS2 and LRS3 dataset and the other is for VoxCeleb2 dataset.
We pre-trained visual feature extractors and not fine-tuned on the separation task; thus there is still a margin to be improved with joint training of the extractor and the separator.

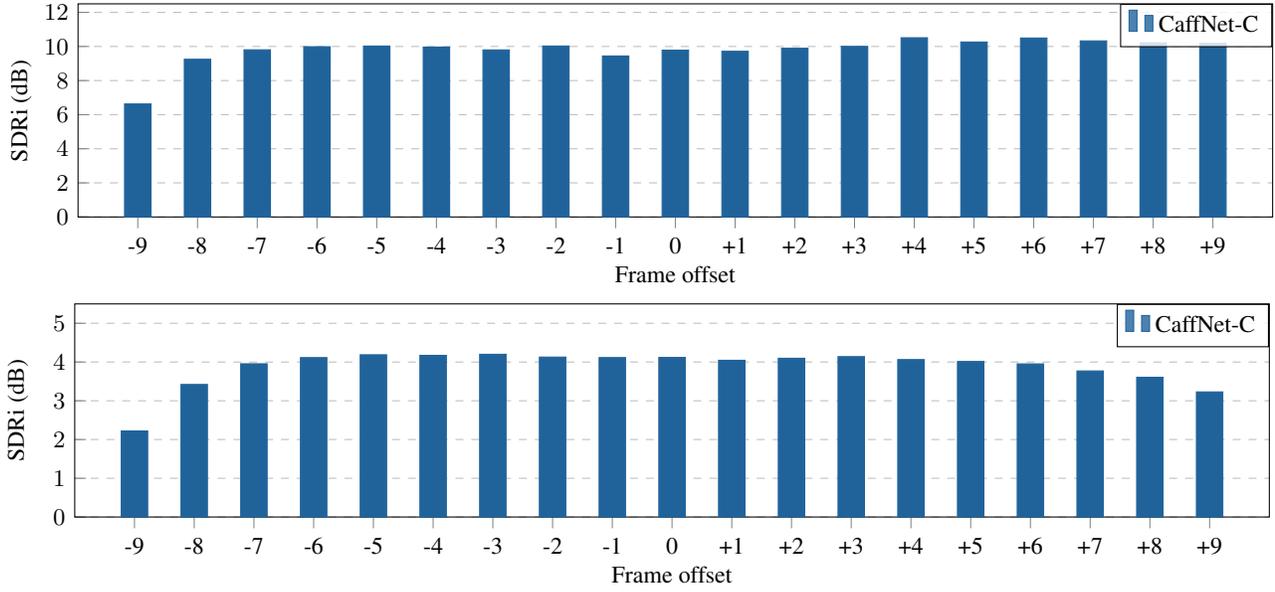
\begin{figure*}[!ht]
    \centering
    % \resizebox{\linewidth}{!}{   
    \begin{tabular}{c}
        
        \centering
        \begin{tikzpicture}
        \tikzstyle{every node}=[font=\small]
            \begin{axis}[
                width=\textwidth, 
                height=0.53\columnwidth,
                xtick pos=left, 
                ytick pos=left,
                ybar=6pt,
                bar width=10pt,
                xtick distance=1000,
                xmin=-10, xmax=+10,
                ymin=0,   ymax=12.5,
                ylabel={SDRi (dB)},
                y label style={at={(0.03,0.5)},font=\small},
                xlabel={Frame offset\vspace{-10pt}},
                label style = {font=\small},
                ticklabel style = {font=\small},
                % symbolic x coords={-9,-8,-7,-6,-5,-4,-3,-2,-1,0,+1,+2,+3,+4,+5,+6,+7,+8,+9},
                xticklabels={-9,-8,-7,-6,-5,-4,-3,-2,-1,0,+1,+2,+3,+4,+5,+6,+7,+8,+9},
                xtick={-9,-8,-7,-6,-5,-4,-3,-2,-1,0,+1,+2,+3,+4,+5,+6,+7,+8,+9},
                ytick={0,2,4,6,8,10,12,14,16,18},
                % minor tick length=1ex,
                % major x tick style = {opacity=1},
                x tick label style={/pgf/number format/1000 sep=},
                ymajorgrids=true,
                xmajorgrids=false,
                grid style=dashed,
                % legend style={at={(0.72, 0.95)},
                legend style={
                at={(0.936, 1)},
                anchor=north, legend columns=-1, font=\small,
                fill opacity=0.8, draw opacity=1,text opacity=1},
                ]
                \addplot [style={color5,fill=color5,mark=none}] coordinates{%CaffNet-C
                (-9, 6.63) (-8, 9.25) (-7, 9.805) (-6, 9.986) (-5, 10.024) (-4, 9.965) (-3, 9.8) (-2, 10.027) (-1, 9.443) (0, 9.782) (+1, 9.718) (+2, 9.895) (+3, 10.005) (+4, 10.512) (+5, 10.262) (+6, 10.496) (+7, 10.327) (+8, 10.213) (+9, 10.18)};
                \legend{\complexmodel}\vspace{-15pt}
            \end{axis}\vspace{-15pt}
        \end{tikzpicture}\label{fig:result_lrs3}
        \end{tabular}
    \\
    
    \begin{tabular}{c}
        
        \centering
        \begin{tikzpicture}
        \tikzstyle{every node}=[font=\small]
            \begin{axis}[
                width=\textwidth, 
                height=0.53\columnwidth,
                xtick pos=left, 
                ytick pos=left,
                ybar=6pt,
                bar width=10pt,
                xtick distance=1000,
                xmin=-10, xmax=+10,
                ymin=0,   ymax=5.5,
                ylabel={SDRi (dB)},
                y label style={at={(0.03,0.5)},font=\small},
                xlabel={Frame offset\vspace{-10pt}},
                label style = {font=\small},
                ticklabel style = {font=\small},
                % symbolic x coords={-9,-8,-7,-6,-5,-4,-3,-2,-1,0,+1,+2,+3,+4,+5,+6,+7,+8,+9},
                xticklabels={-9,-8,-7,-6,-5,-4,-3,-2,-1,0,+1,+2,+3,+4,+5,+6,+7,+8,+9},
                xtick={-9,-8,-7,-6,-5,-4,-3,-2,-1,0,+1,+2,+3,+4,+5,+6,+7,+8,+9},
                ytick={0,1,2,3,4,5,6,7,8,9,10,12,14,16,18},
                % minor tick length=1ex,
                % major x tick style = {opacity=1},
                x tick label style={/pgf/number format/1000 sep=},
                ymajorgrids=true,
                xmajorgrids=false,
                grid style=dashed,
                % legend style={at={(0.72, 0.95)},
                legend style={
                at={(0.936, 1)},
                anchor=north, legend columns=-1, font=\small,
                fill opacity=0.8, draw opacity=1,text opacity=1},
                ]
                \addplot [style={color5,fill=color5,mark=none}] coordinates{%CaffNet-C
                (-9, 2.221) (-8, 3.421) (-7, 3.955) (-6, 4.114) (-5, 4.187) (-4, 4.172) (-3, 4.200) (-2, 4.128) (-1, 4.115) (0, 4.118) (+1, 4.045) (+2, 4.097) (+3, 4.139) (+4, 4.063) (+5, 4.014) (+6, 3.952) (+7, 3.770) (+8, 3.606) (+9, 3.229)};
                \legend{\complexmodel}\vspace{-15pt}
            \end{axis}\vspace{-15pt}
        \end{tikzpicture}\label{fig:result_vox2}
       
        \end{tabular}
        % }
    \vspace{-13pt}
    \caption{Speech separation performance with respect to each delay offset between audio and visual streams on LRS3 (top) and VoxCeleb2 (bottom) datasets. The frame offset unit is 40ms which is the duration length between consecutive video frames. 
    % (a) reports the SDRi evaluation using ground-truth phase with estimated magnitude spectrum. (b) and (c) report the SDRi and PESQi evaluation on predicted phase as well as estimated magnitude spectrum, respectively.
    }
    \label{fig:results_vox2lrs3}\vspace{-10pt}
\end{figure*}

\paragraph{Training and Optimization}
We use PyTorch library~\cite{paszke2017automatic} and single RTX 24GB to train our networks.
The network parameters are optimized using the mini-batch stochastic gradient descent (SGD) method.
All networks are trained from scratch and we applied the Adam optimizer~\cite{kingma2014adam} with learning rate $1\times 10^{-5}$, beta1 $\beta_1 = 0.9$, beta2 $\beta_2 = 0.999$ and batch size 32. 
We use PyTorch’s ReduceLROnPlateau learning rate scheduler with a reduction factor of 0.8 and a patience parameter of 2 to adapt the learning rate during training.
At each training iteration, we randomly sampled one audio-video pair and another audio that has a different identity.
We randomly take segments from each audio signal and add them to make mixture input.
Hence, we use 64K training samples for each epoch and all networks are trained for a total of 50 epochs.

\subsection{More Results}

\paragraph{Quantitative Results}
We provide more quantitative results for AVSS performance regarding SDR improvement~(SDRi) metric concerning varying delay in \figref{fig:results_vox2lrs3} on LRS3 and VoxCeleb2 datasets.
The results are obtained from \complexmodel using predicted magnitude and phase.
On the top of \figref{fig:results_vox2lrs3}, we report the average SDRi of unseen 1000 speaker samples on LRS3 dataset.
In this experiment, we train \complexmodel using LRS2 dataset only, so the network cannot see any samples of LRS3 at the training stage.
At the bottom of \figref{fig:results_vox2lrs3}, we provide the average SDRi of unseen 1000 speaker samples on VoxCeleb2 dataset.

\begin{figure}[!ht]
\centering
    \renewcommand{\thesubfigure}{}
    \subfigure[(a) With jitter]{\includegraphics[width=0.45\linewidth]{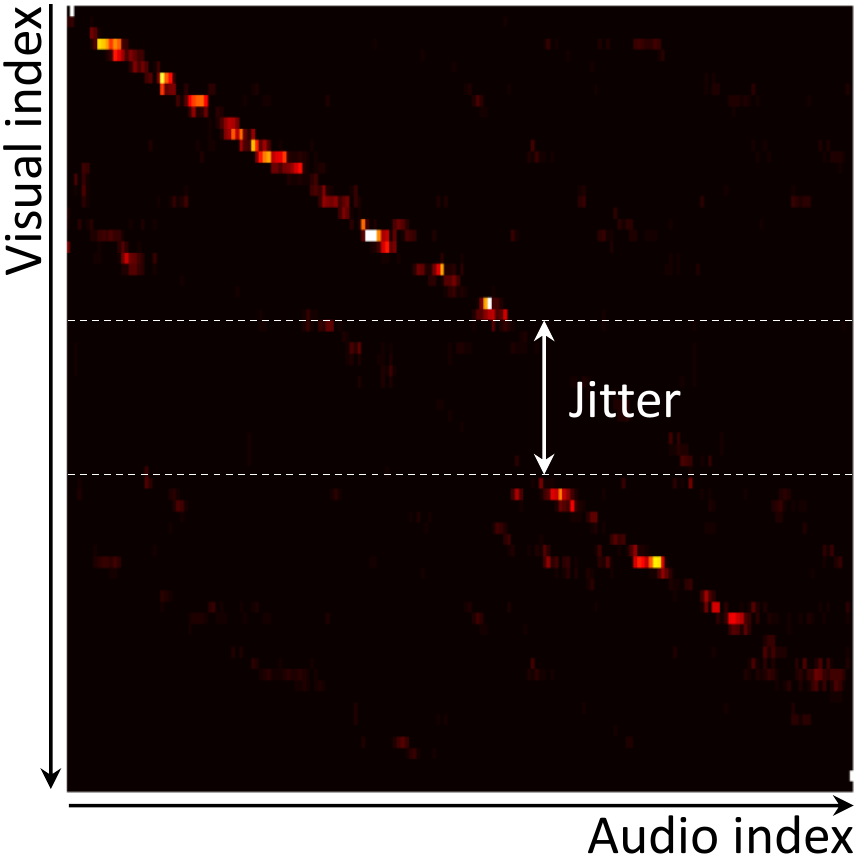}}
    \subfigure[(b) Without jitter]{\includegraphics[width=0.45\linewidth]{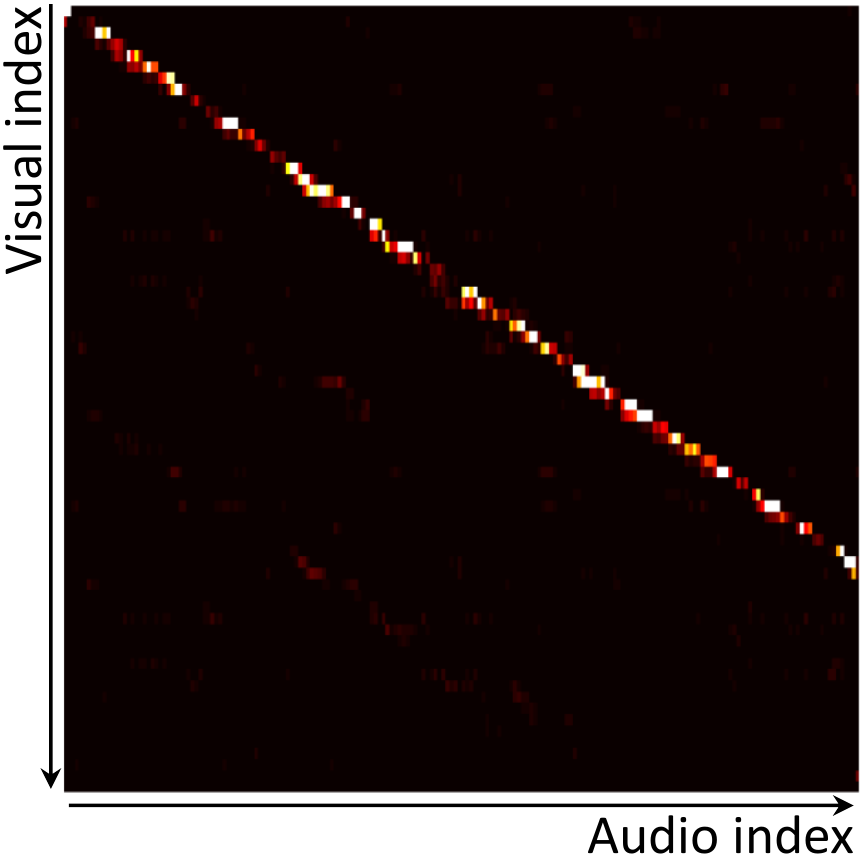}}\hfill 
    \\ \vspace{-7pt}
    \subfigure[(c) Mixture speech ]{\includegraphics[width=0.45\linewidth]{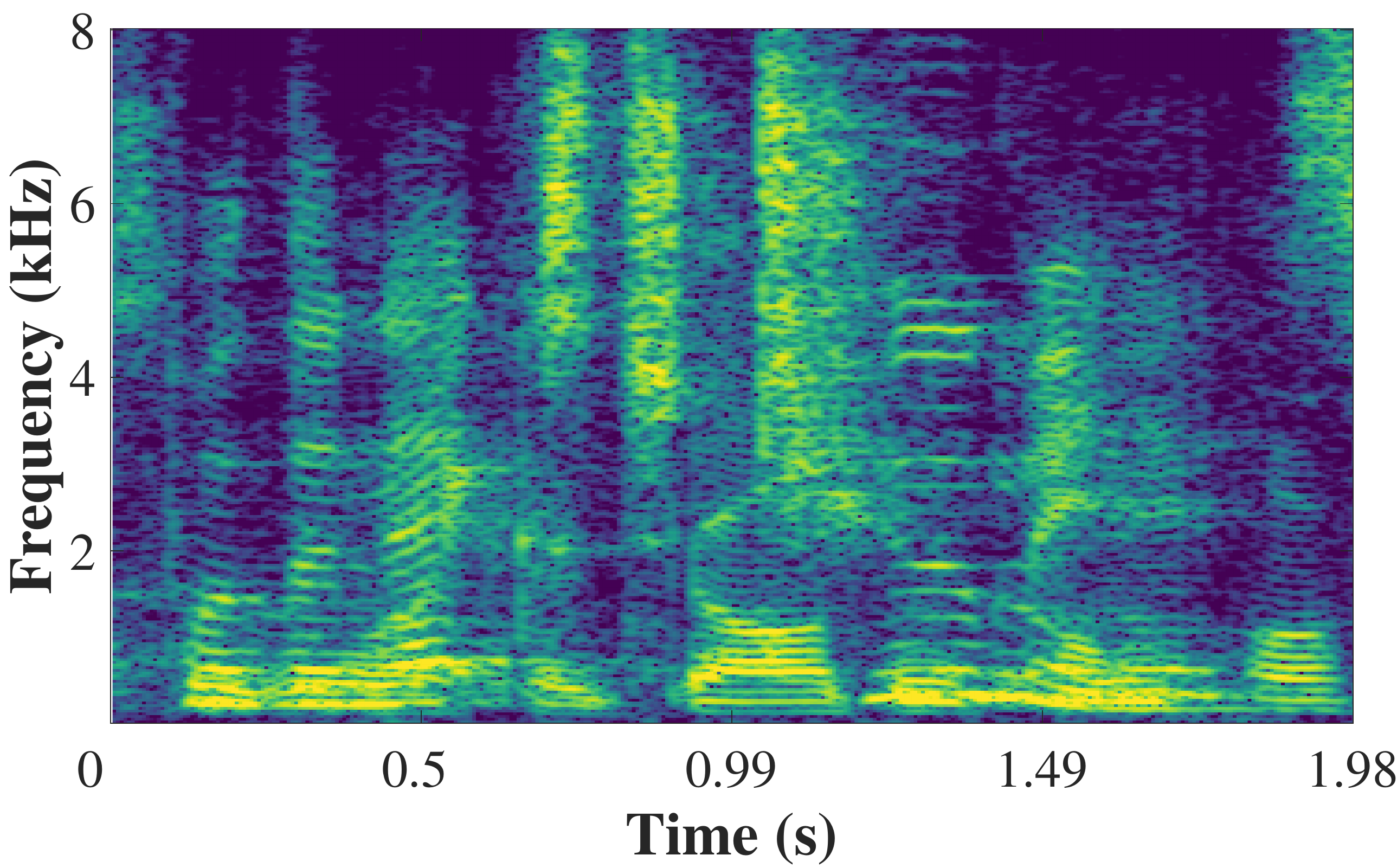}}
    \subfigure[(d) Source speech (GT)]{\includegraphics[width=0.45\linewidth]{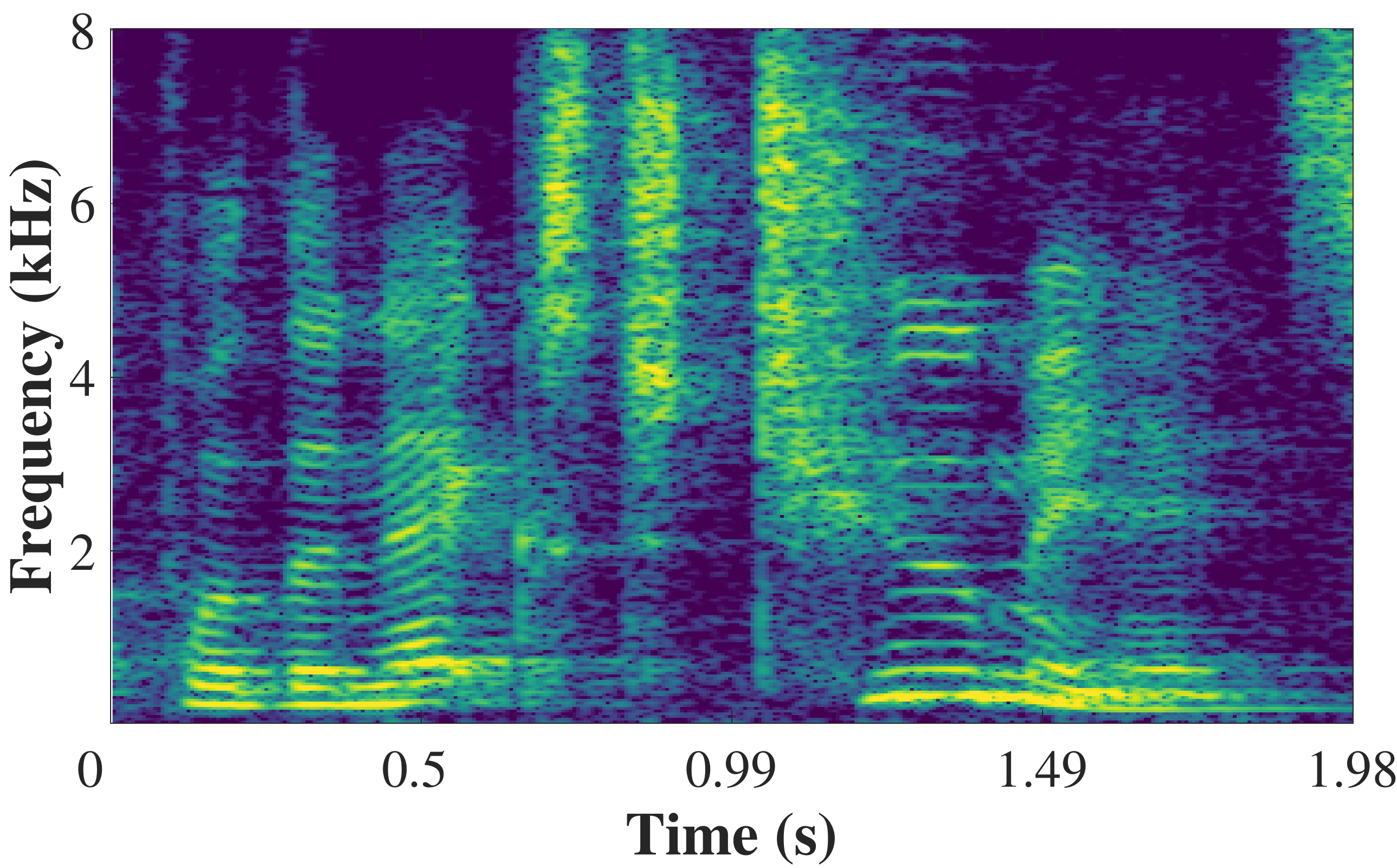}}\hfill 
    \\ \vspace{-7pt}
    \subfigure[(e) Enhanced speech with (a)]{\includegraphics[width=0.45\linewidth]{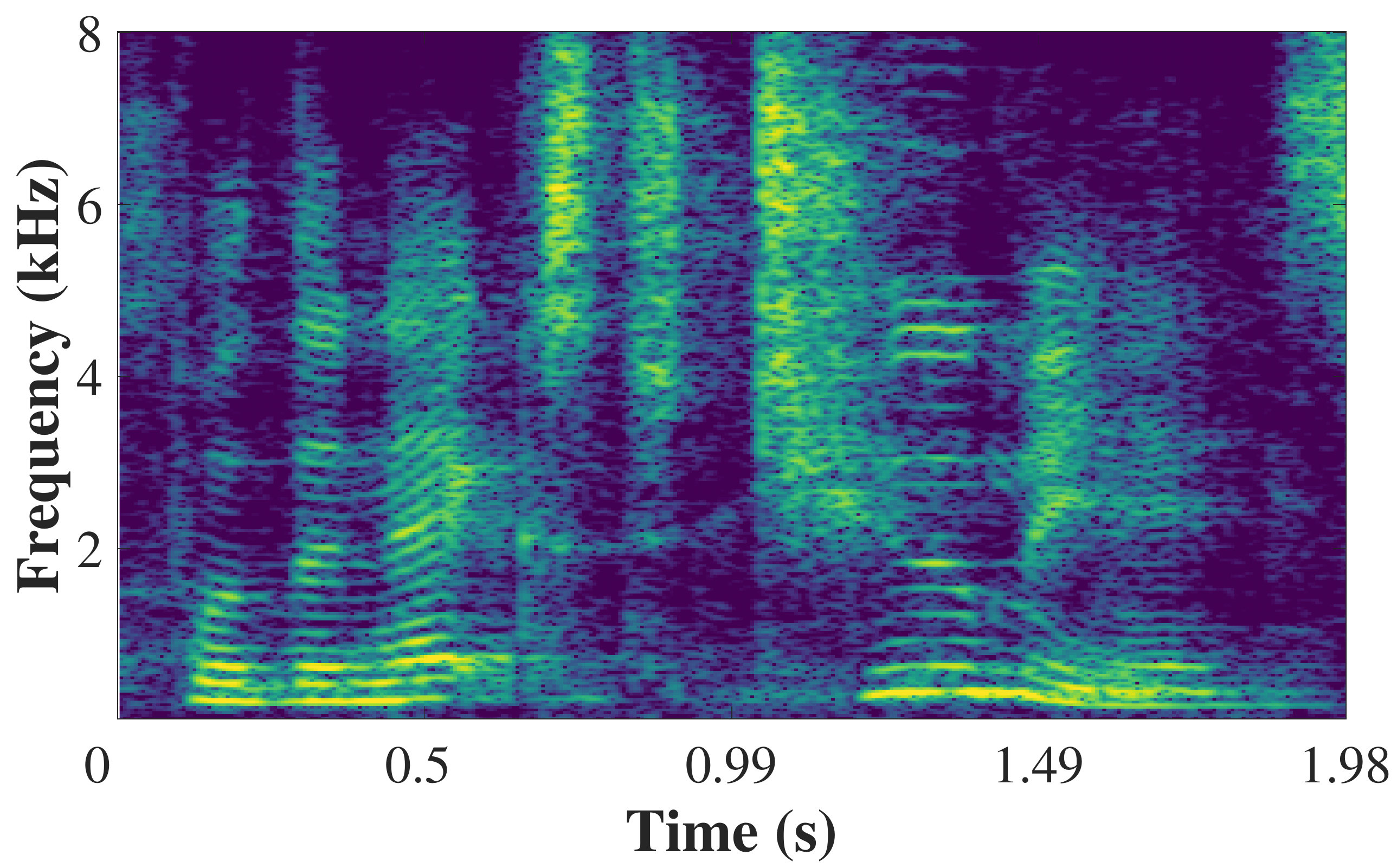}}
    \subfigure[(f) Enhanced speech with (b)]{\includegraphics[width=0.45\linewidth]{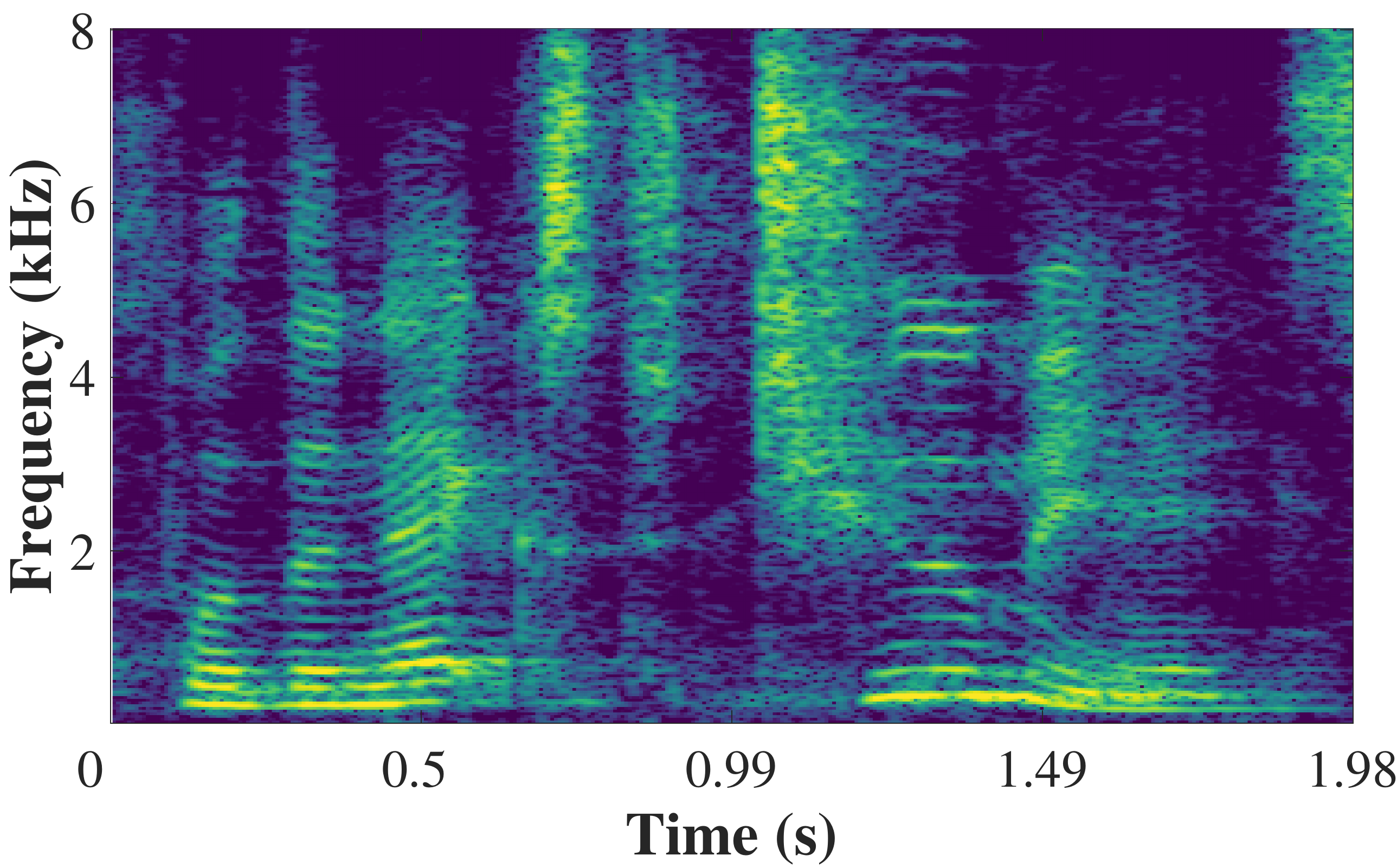}}\hfill \\
    \vspace{-10pt}
    \caption{Qualitative evaluation of affinity matrices and spectrograms with respect to jitter effect on LRS2 dataset. The results are obtained from \complexmodel.}
    \label{fig:lrs2jitter}\vspace{-10pt}
\end{figure}

\begin{figure*}[!ht]
\centering
    \renewcommand{\thesubfigure}{}
    \subfigure{\includegraphics[width=0.2\linewidth]{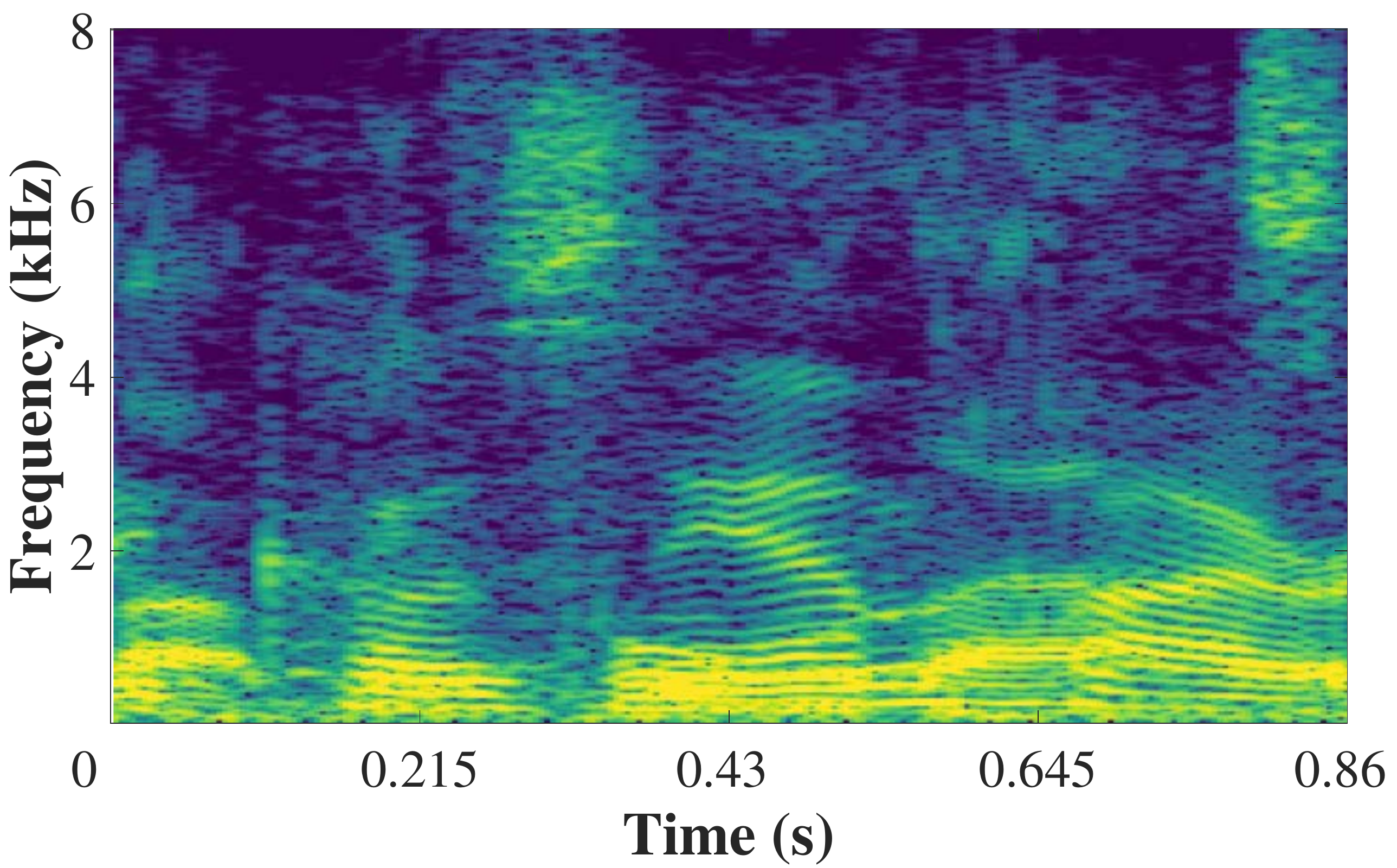}}\hfill
    \subfigure{\includegraphics[width=0.2\linewidth]{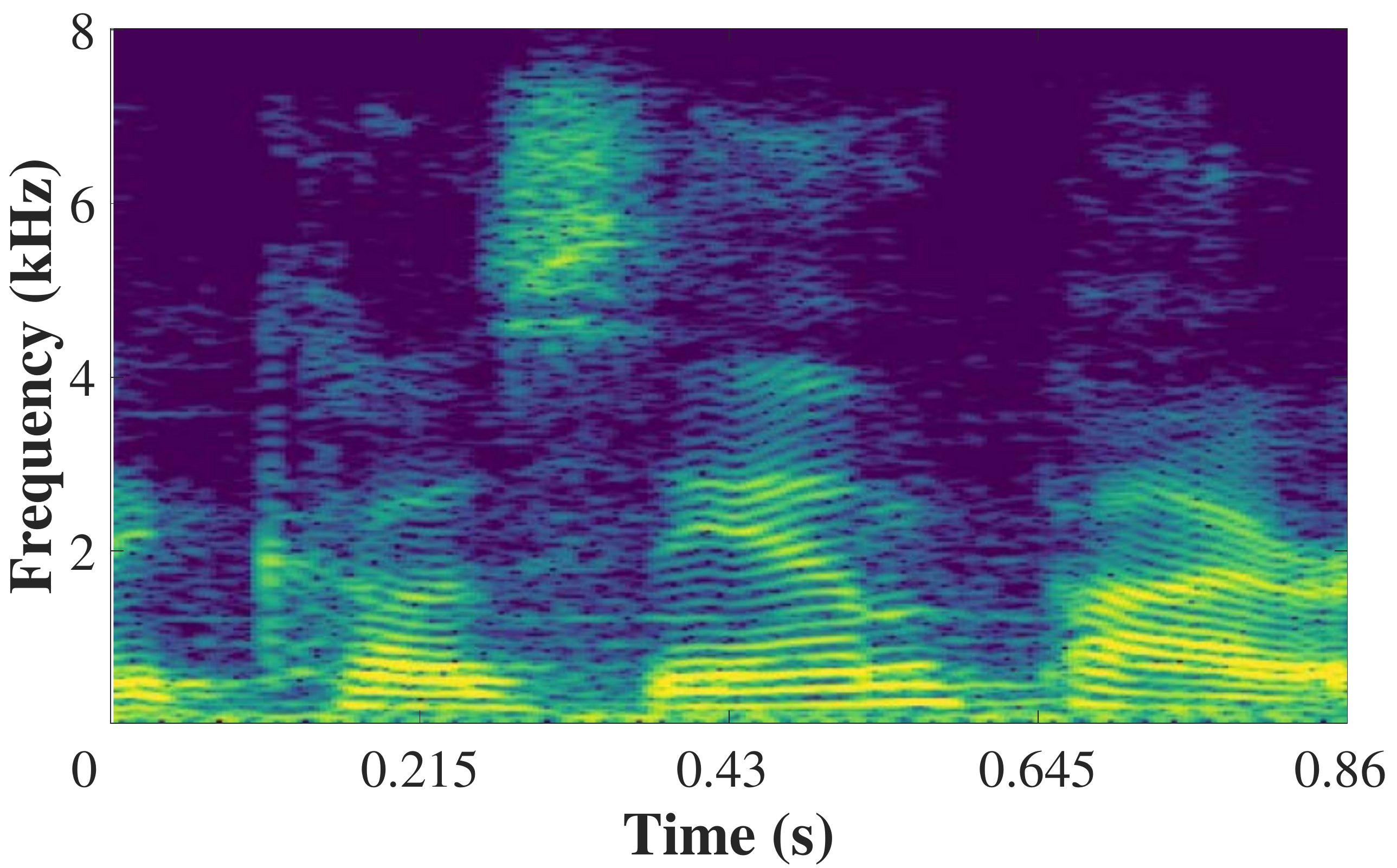}}\hfill
    \subfigure{\includegraphics[width=0.2\linewidth]{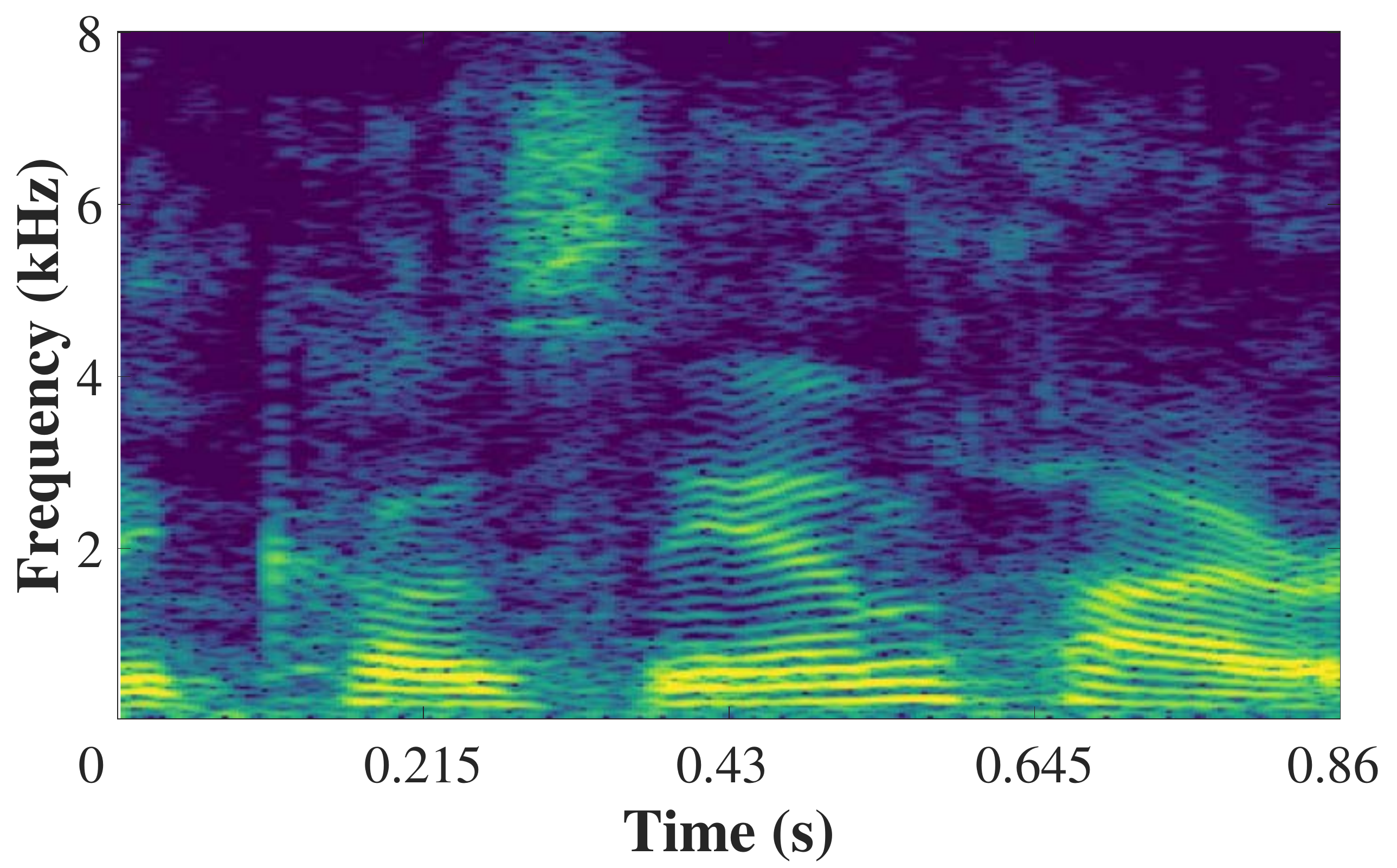}}\hfill
    \subfigure{\includegraphics[width=0.2\linewidth]{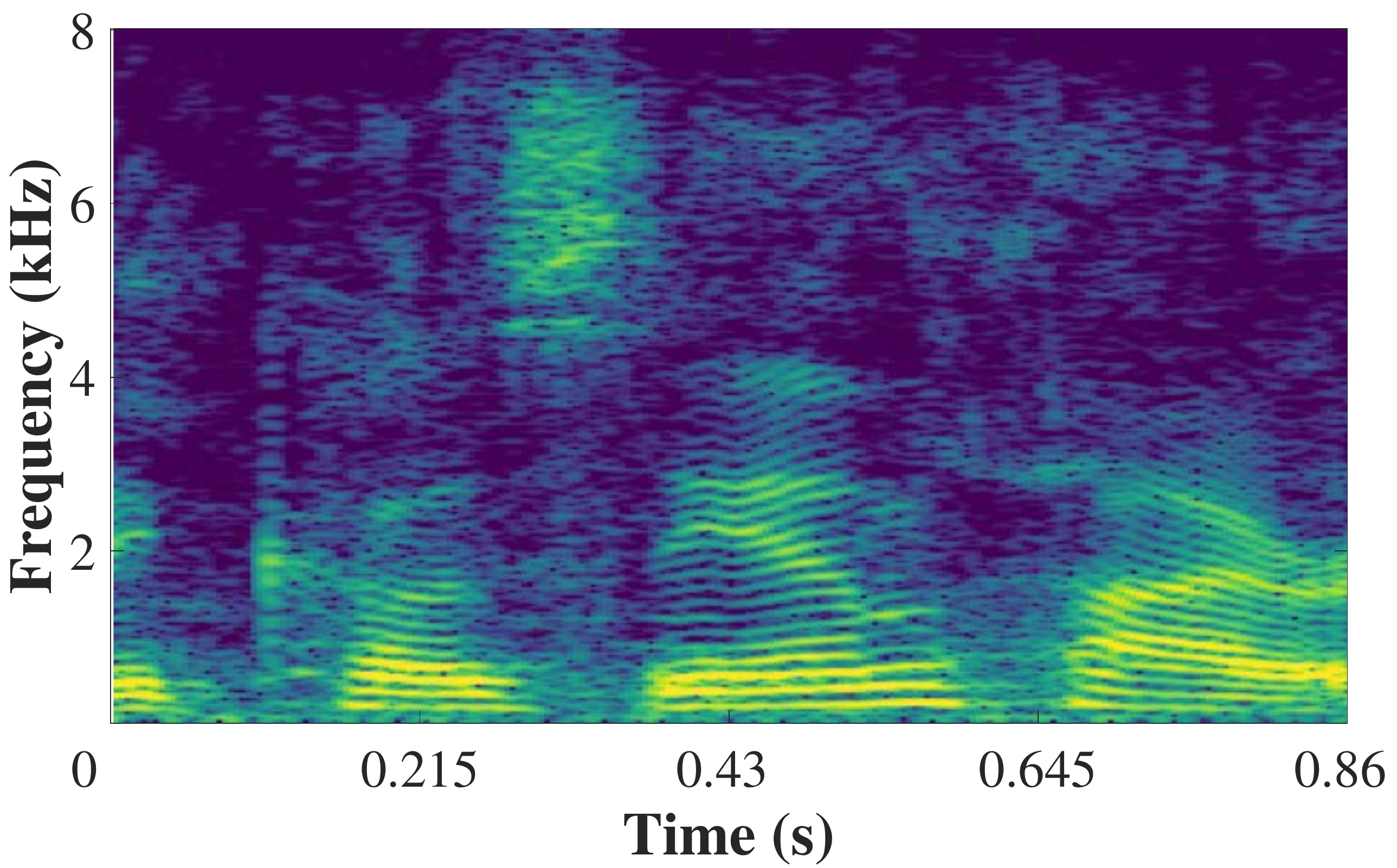}}\hfill
    \subfigure{\includegraphics[width=0.2\linewidth]{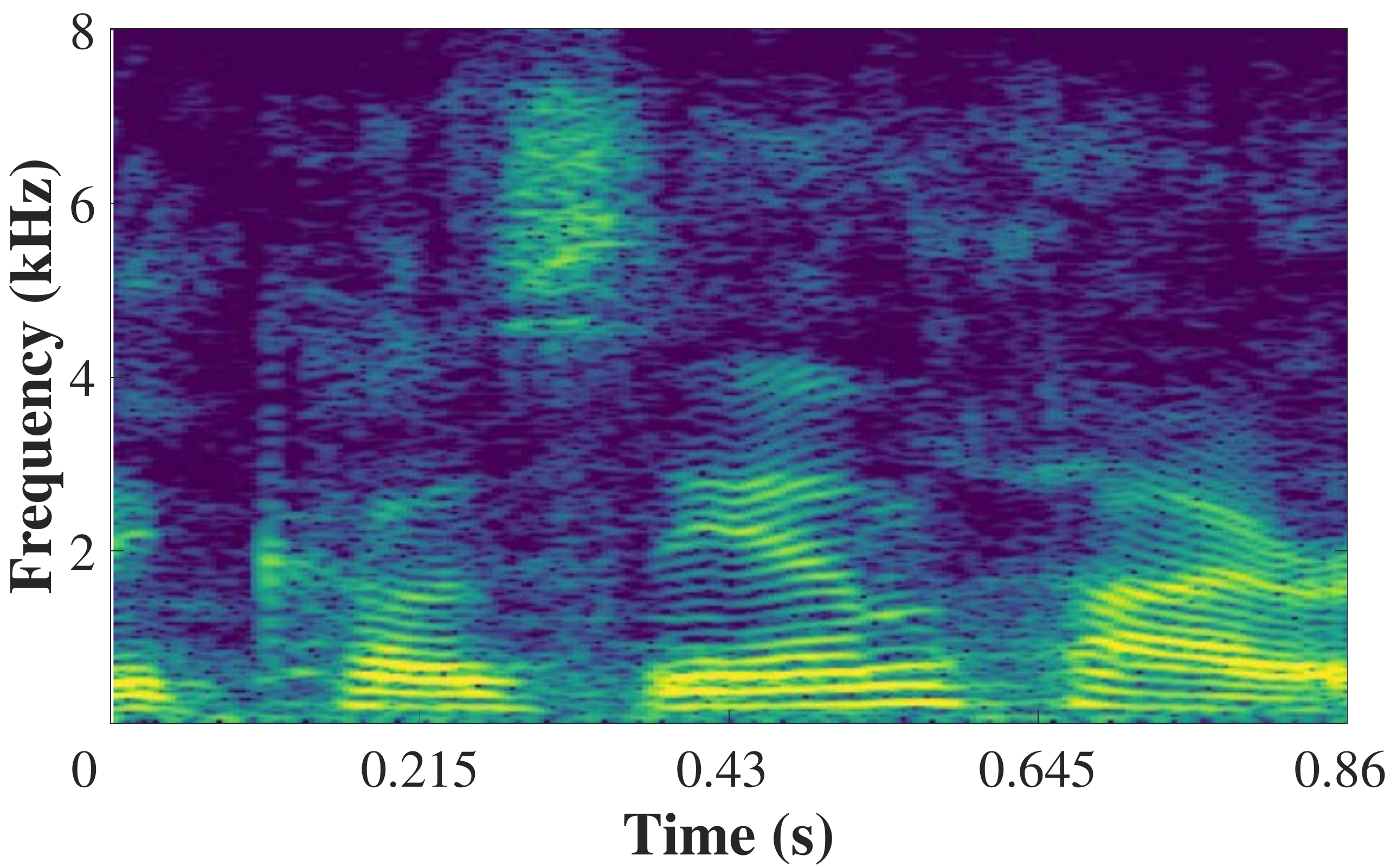}}\hfill
    \\ \vspace{-7pt}
    \subfigure{\includegraphics[width=0.2\linewidth]{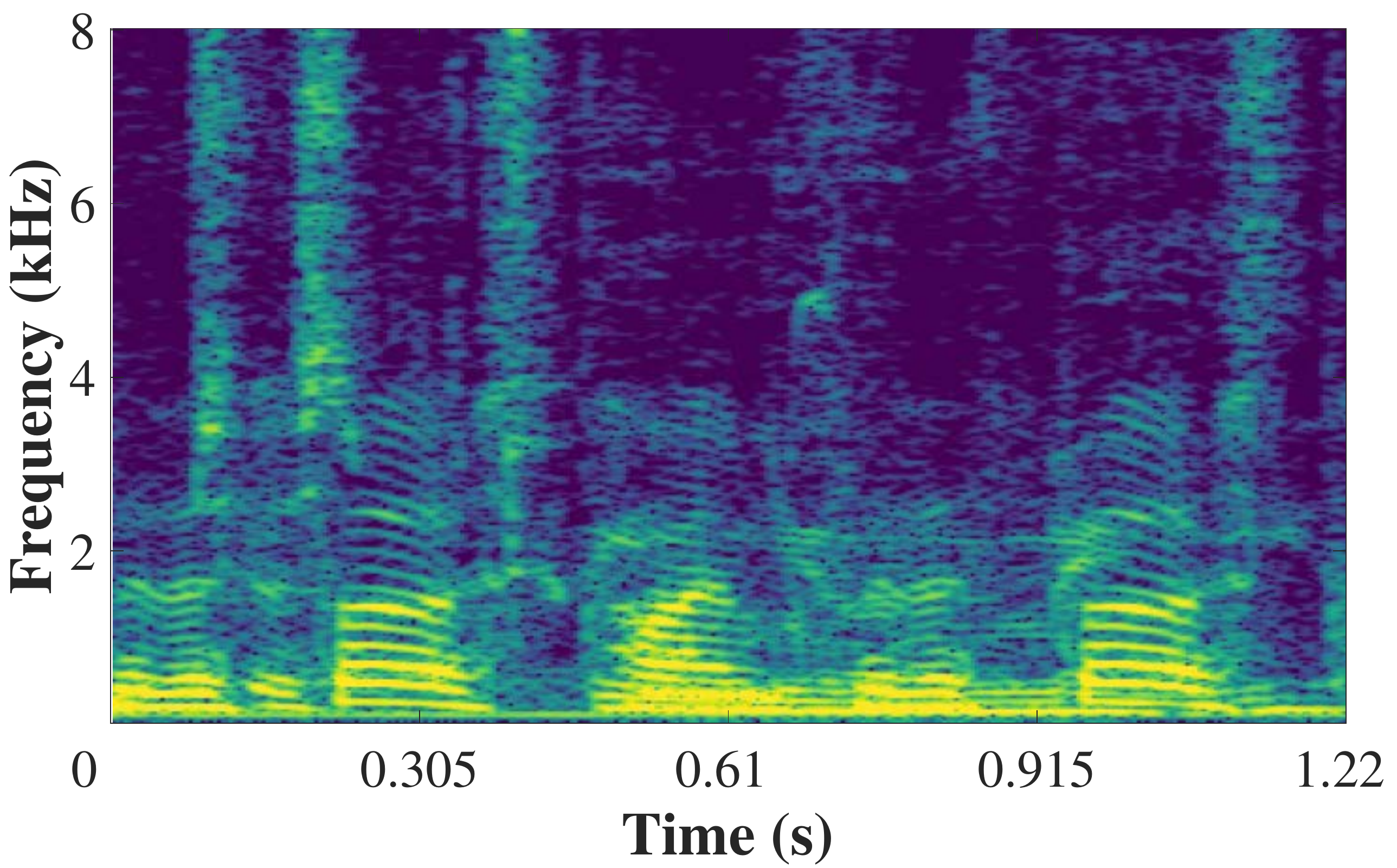}}\hfill
    \subfigure{\includegraphics[width=0.2\linewidth]{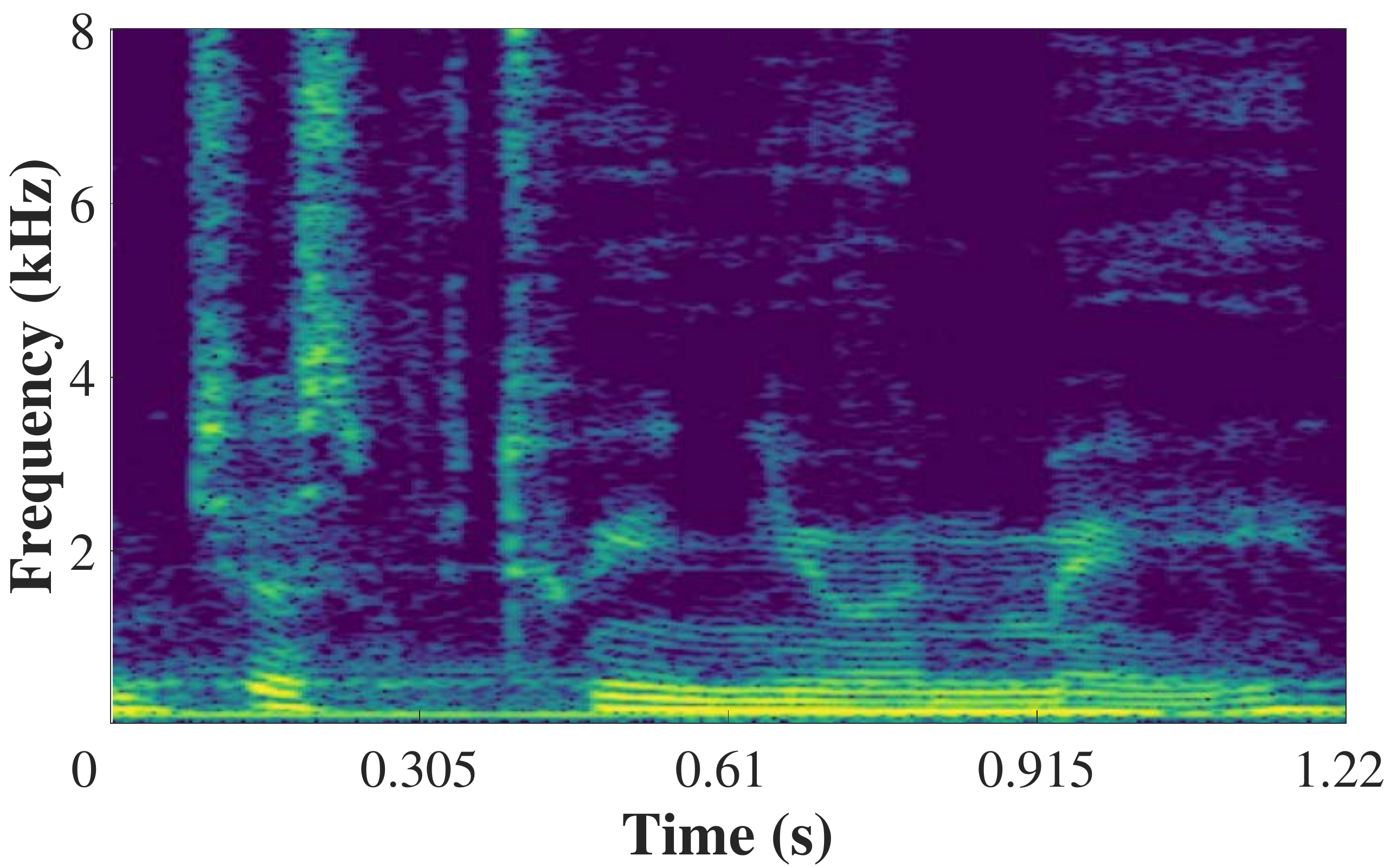}}\hfill
    \subfigure{\includegraphics[width=0.2\linewidth]{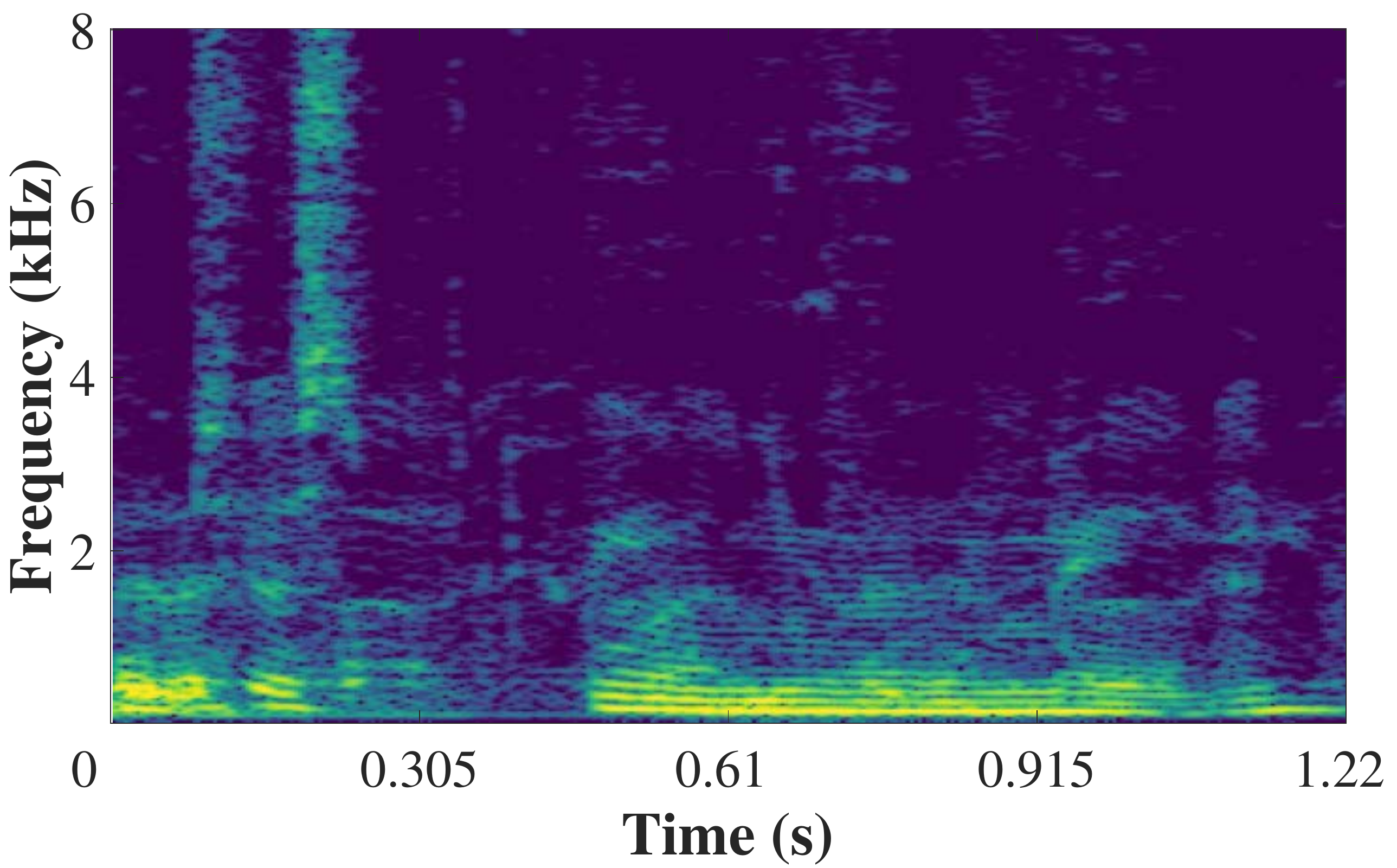}}\hfill
    \subfigure{\includegraphics[width=0.2\linewidth]{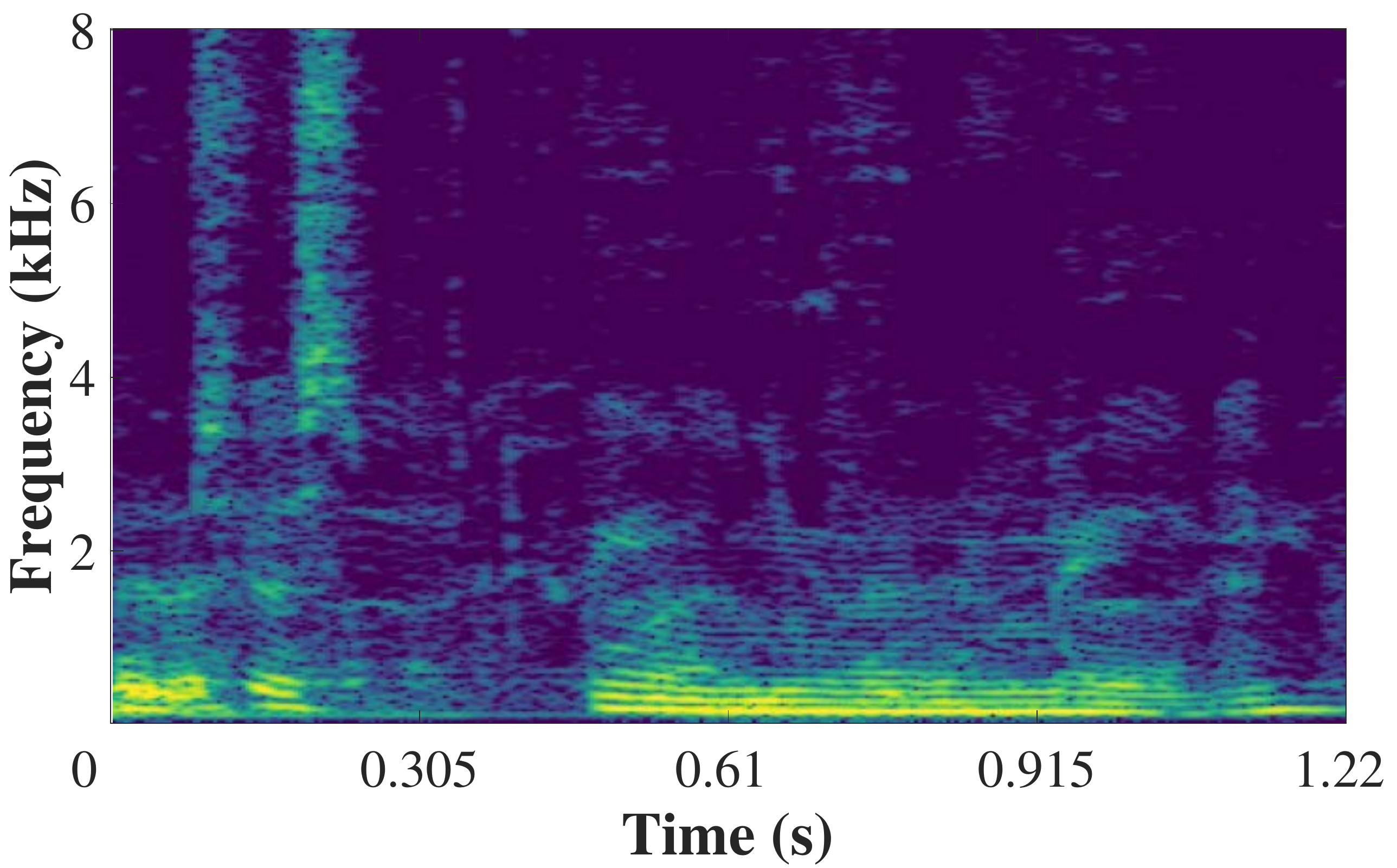}}\hfill
    \subfigure{\includegraphics[width=0.2\linewidth]{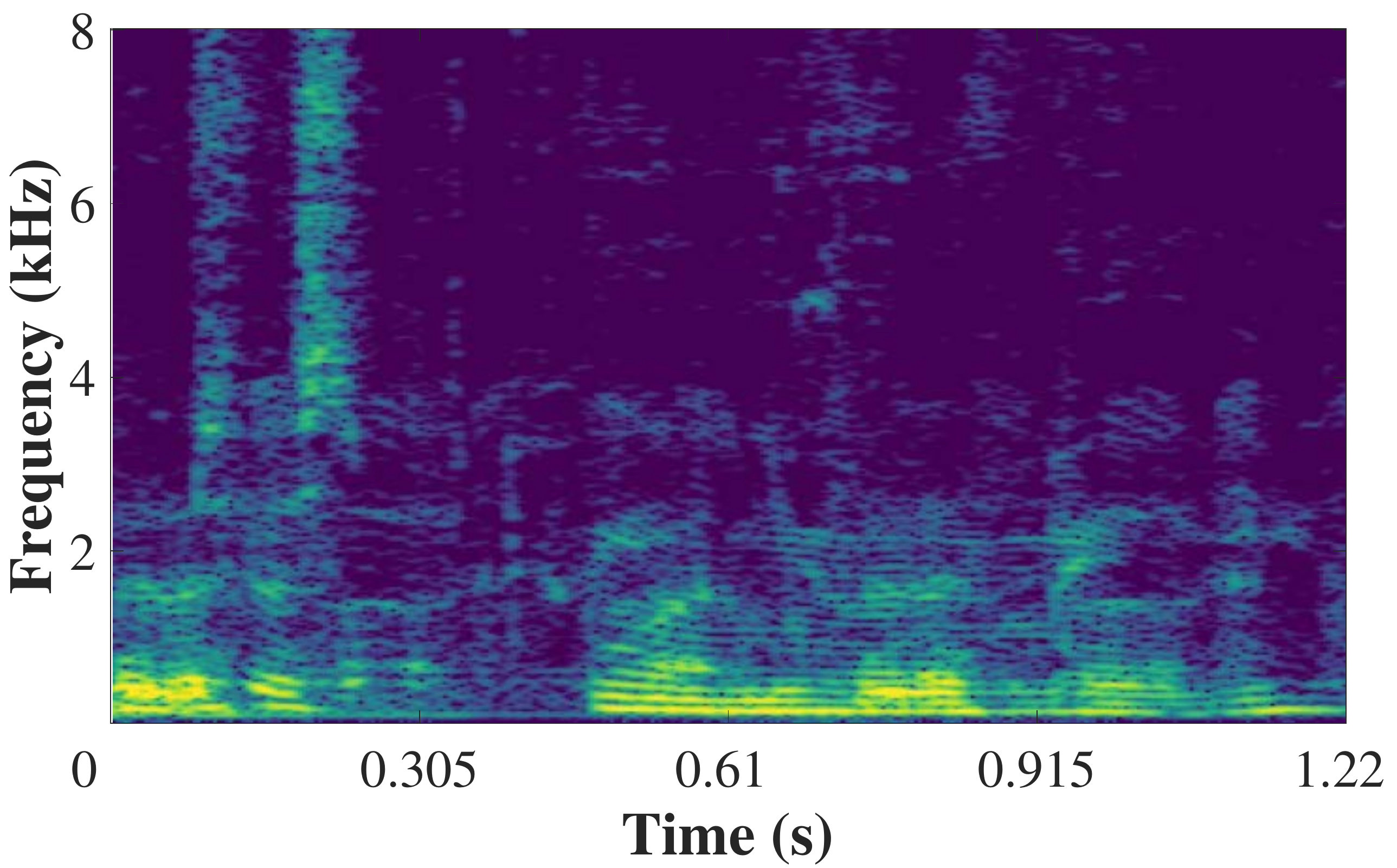}}\hfill
    \\ \vspace{-7pt}
    \subfigure[(a) Mixture]{\includegraphics[width=0.2\linewidth]{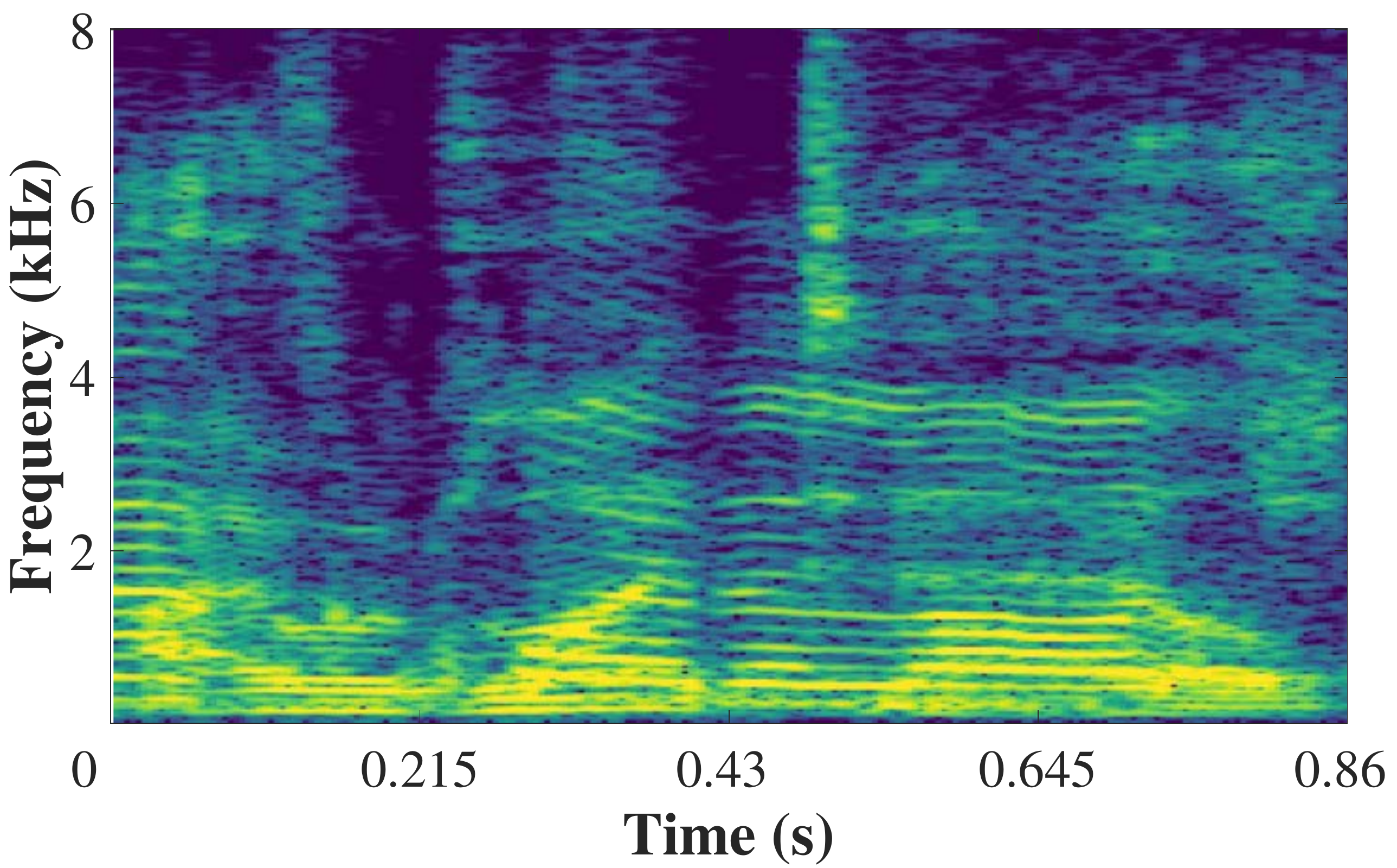}}\hfill
    \subfigure[(b) Source (GT)]{\includegraphics[width=0.2\linewidth]{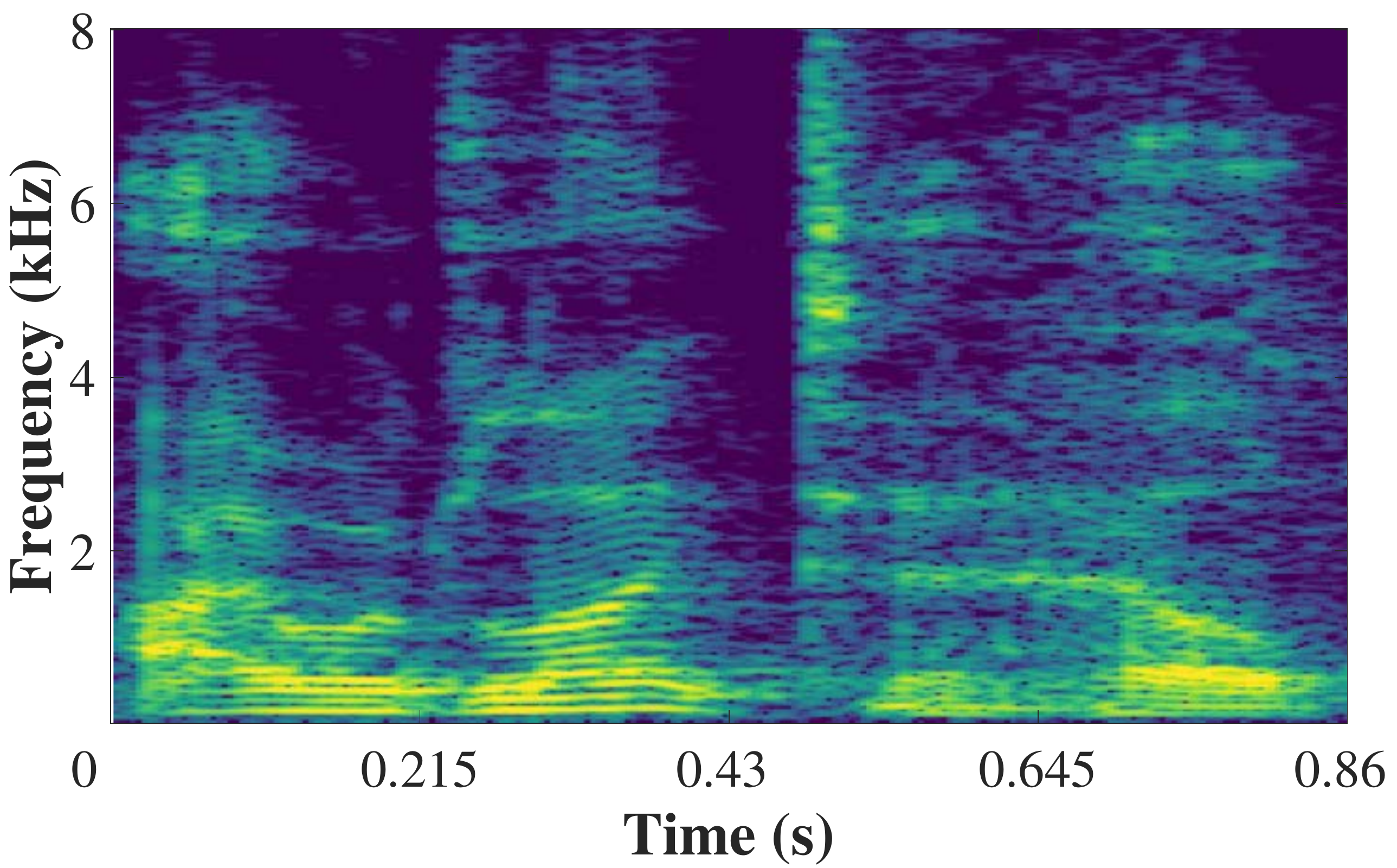}}\hfill
    \subfigure[(c) -5 delay]{\includegraphics[width=0.2\linewidth]{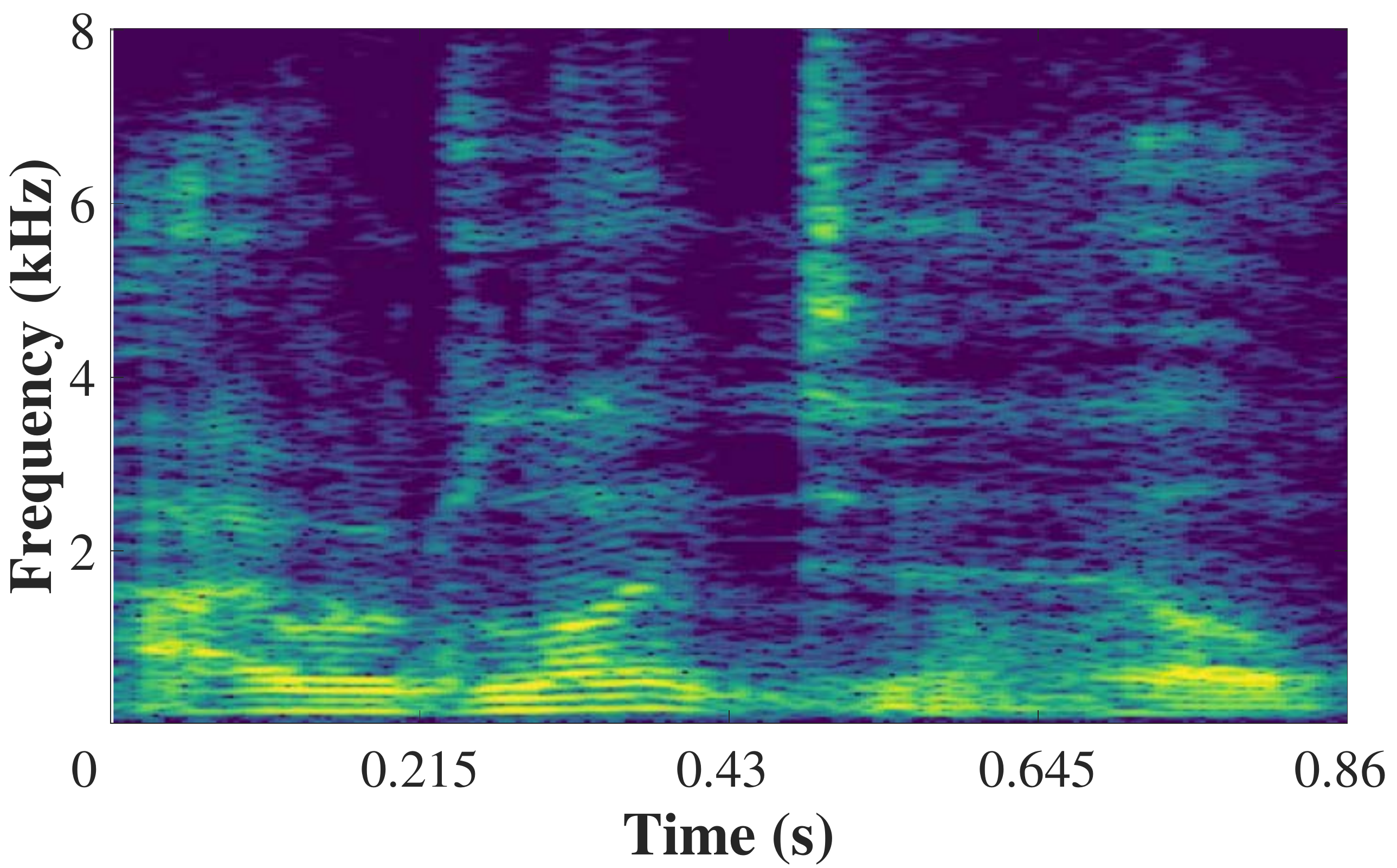}}\hfill
    \subfigure[(d) 0 delay]{\includegraphics[width=0.2\linewidth]{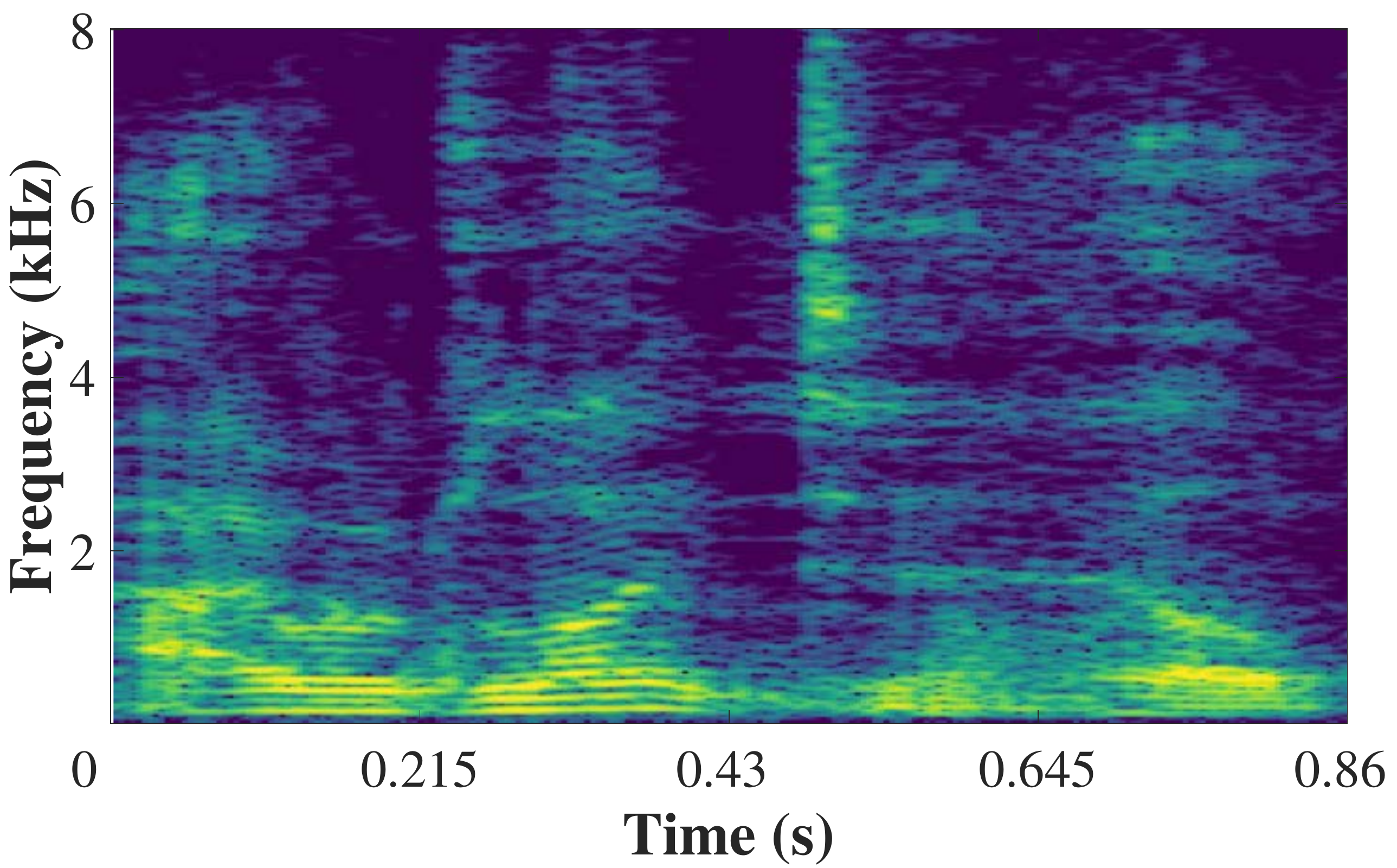}}\hfill
    \subfigure[(e) +5 delay]{\includegraphics[width=0.2\linewidth]{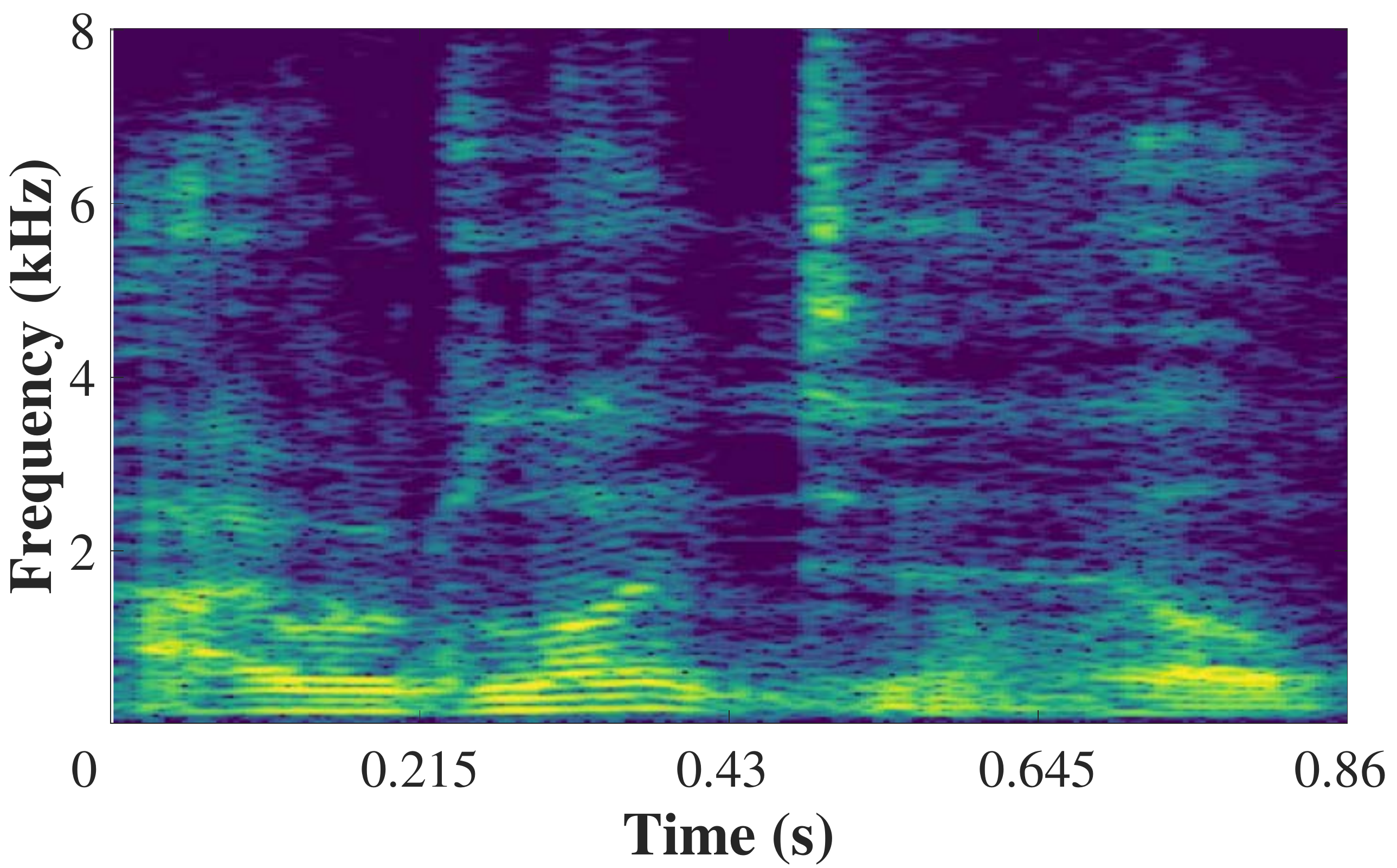}}\hfill
    \\ 
    \vspace{-10pt}
    \caption{Qualitative evaluation of spectrograms with respect to the frame delay on LRS2 dataset. (c), (d) and (e) are results of \complexmodel. Each row is different testing sample and all results are reconstructed from the predicted magnitude and phase.}
    \label{fig:lrs2}\vspace{-10pt}
\end{figure*}

\begin{figure*}[!t]
\centering
    \renewcommand{\thesubfigure}{}
    \subfigure[(a) Mixture]{\includegraphics[width=0.25\linewidth]{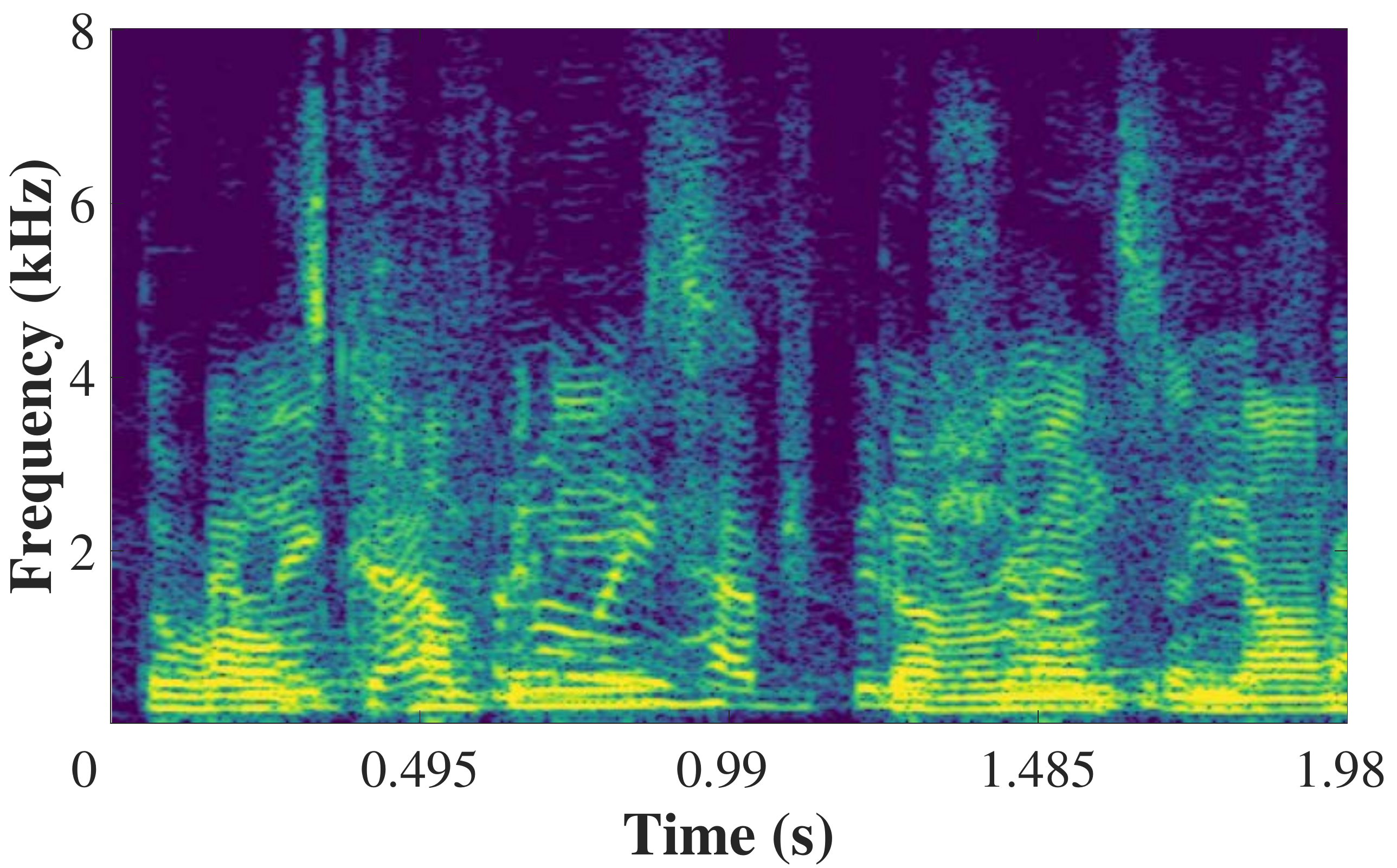}}\hfill
    \subfigure[(b) V-Conv~\cite{afouras2018conversation} (-5 delay)]{\includegraphics[width=0.25\linewidth]{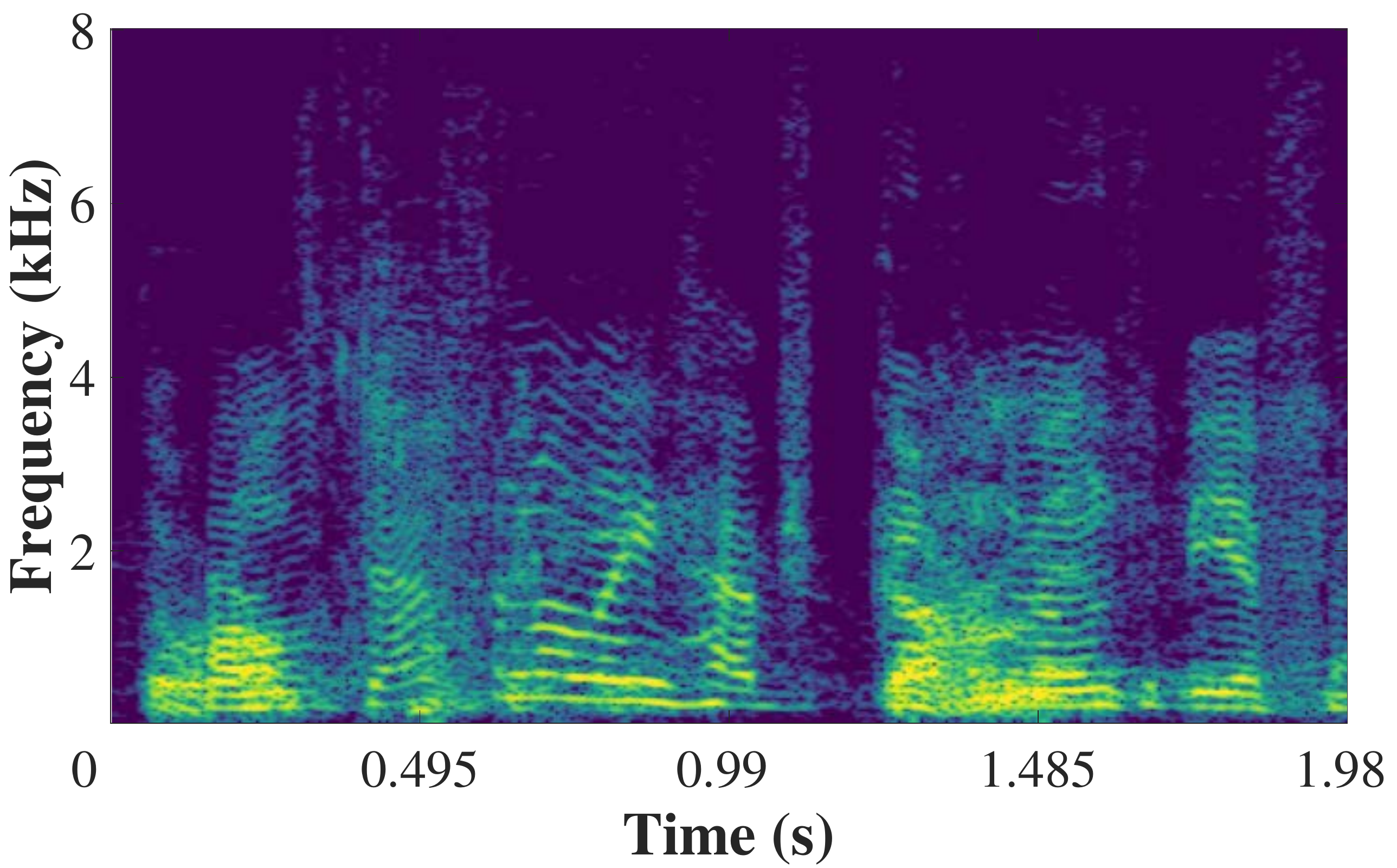}}\hfill
    \subfigure[(c) V-Conv~\cite{afouras2018conversation} (0 delay)]{\includegraphics[width=0.25\linewidth]{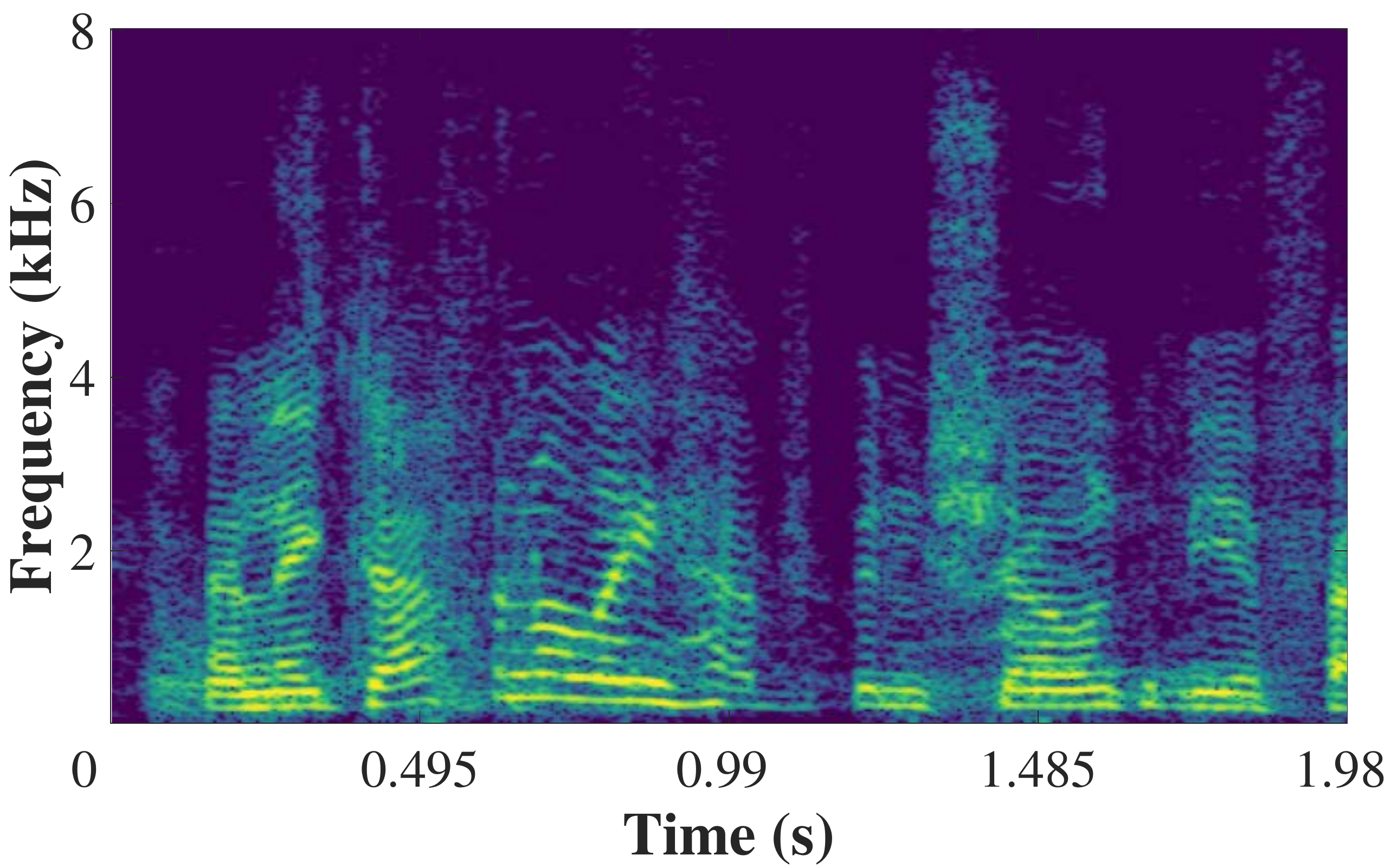}}\hfill
    \subfigure[(d) V-Conv~\cite{afouras2018conversation} (+5 delay)]{\includegraphics[width=0.25\linewidth]{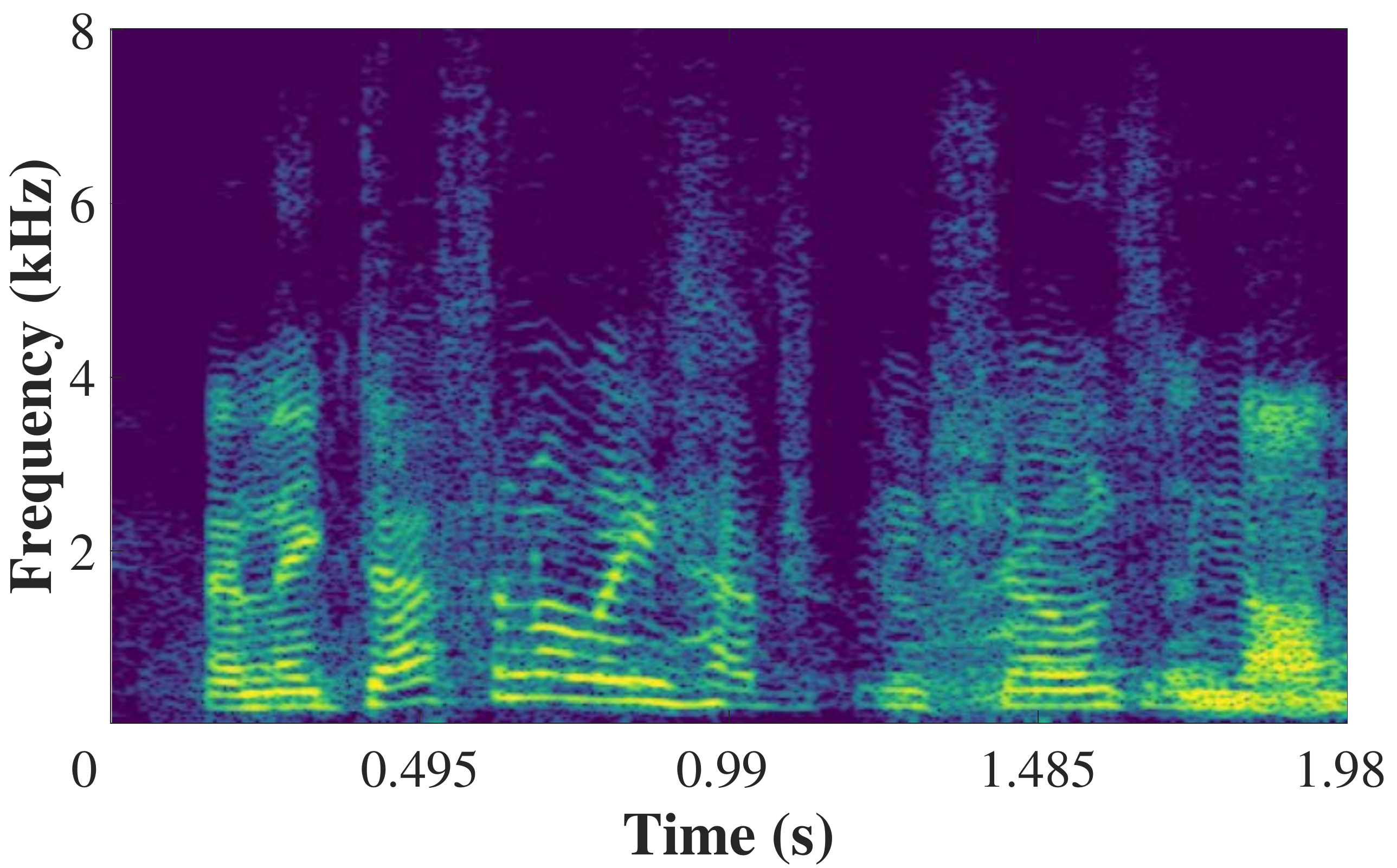}}\hfill\\
    \vspace{-7pt}
    \subfigure[(e) Source (GT)]{\includegraphics[width=0.25\linewidth]{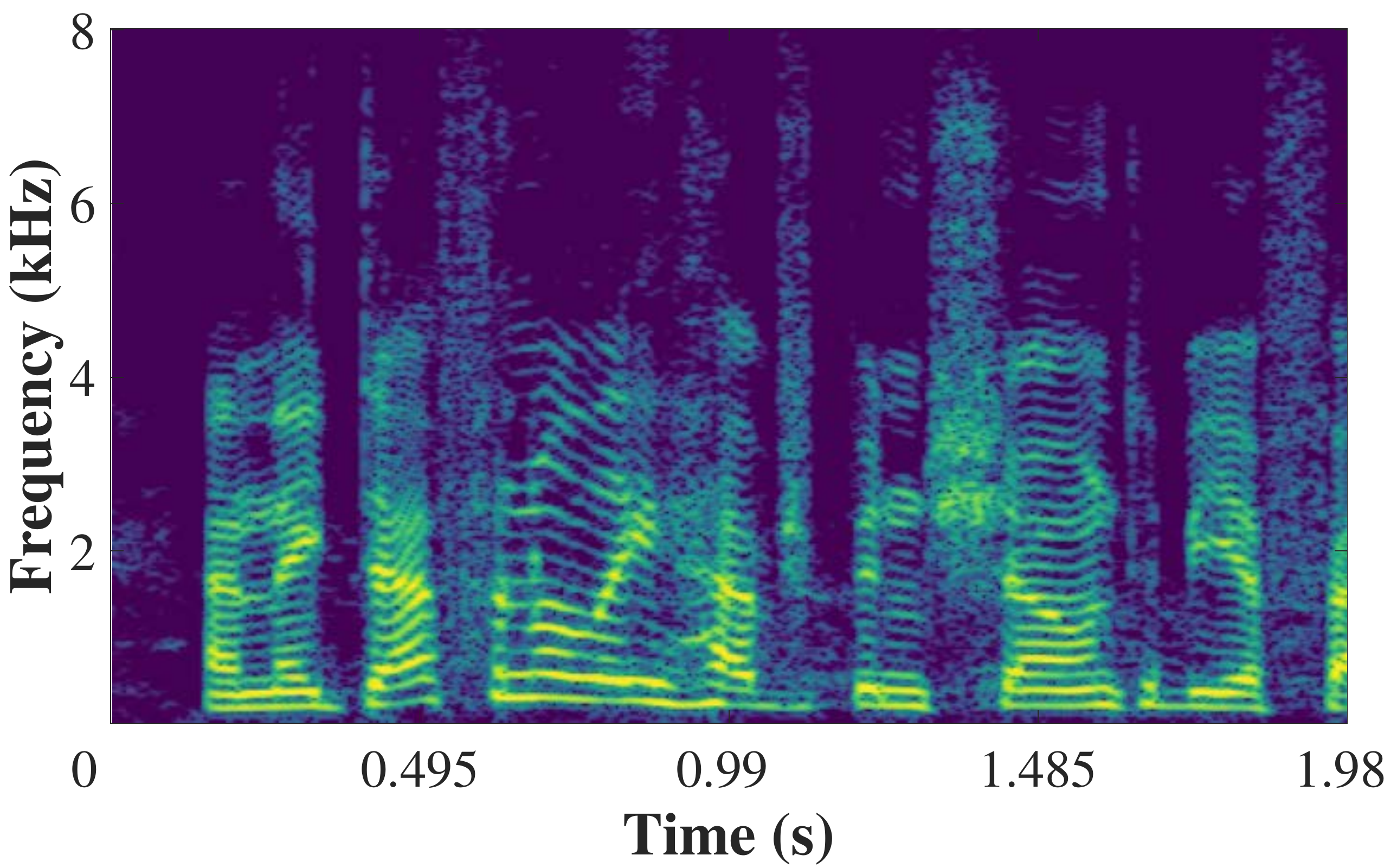}}\hfill
    \subfigure[(f) \basemodel (-5 delay)]{\includegraphics[width=0.25\linewidth]{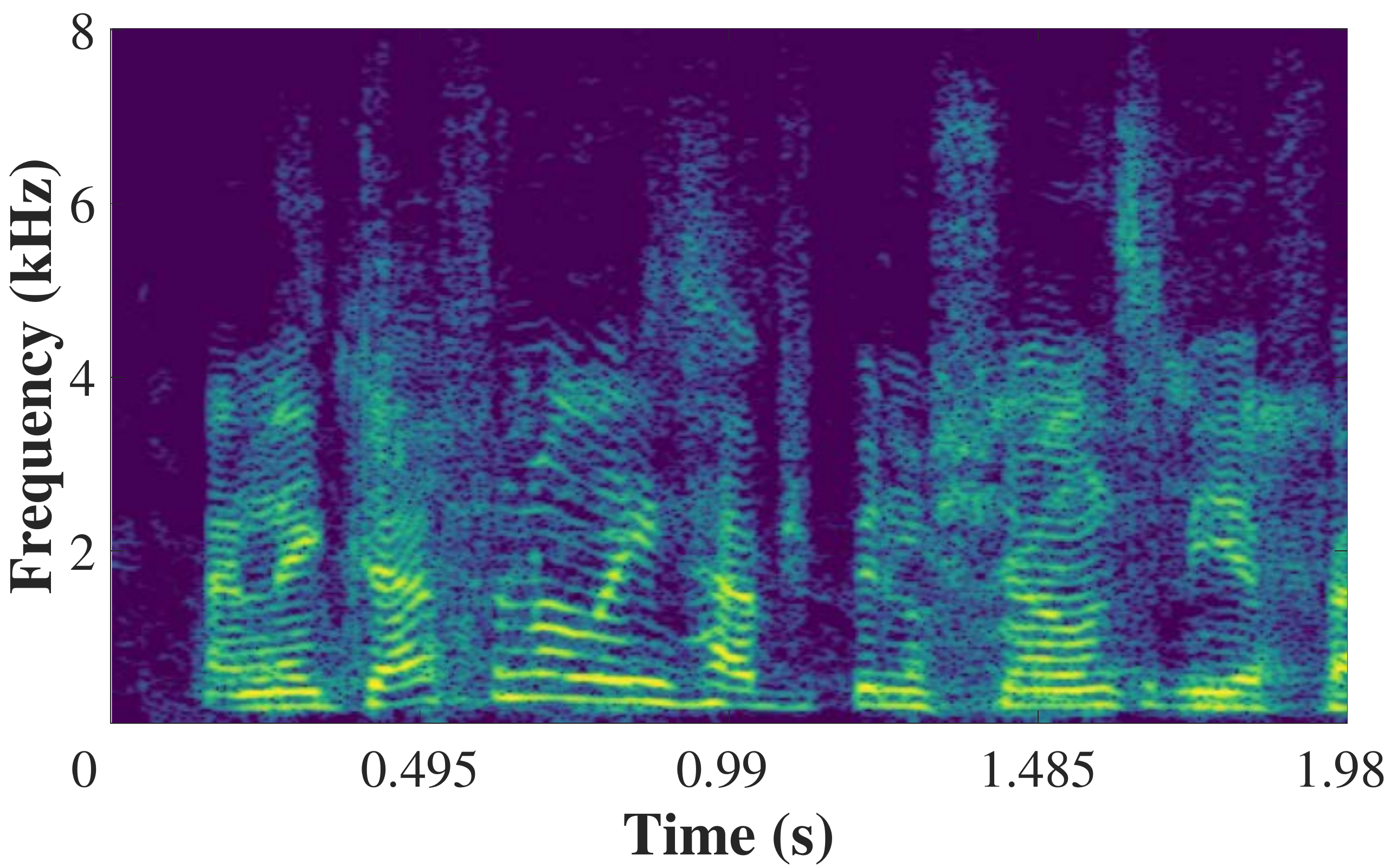}}\hfill
    \subfigure[(g) \basemodel (0 delay)]{\includegraphics[width=0.25\linewidth]{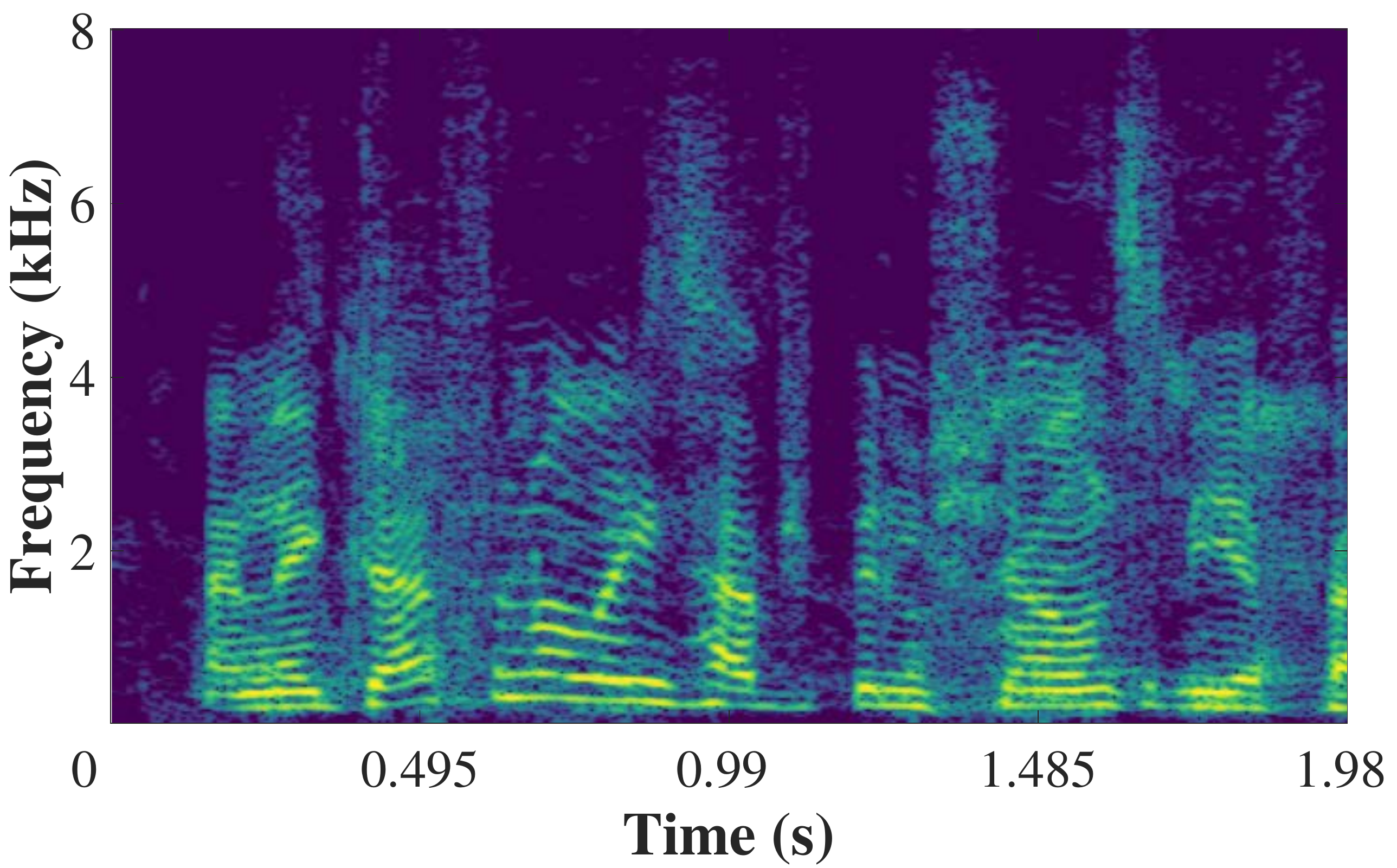}}\hfill
    \subfigure[(h) \basemodel (+5 delay)]{\includegraphics[width=0.25\linewidth]{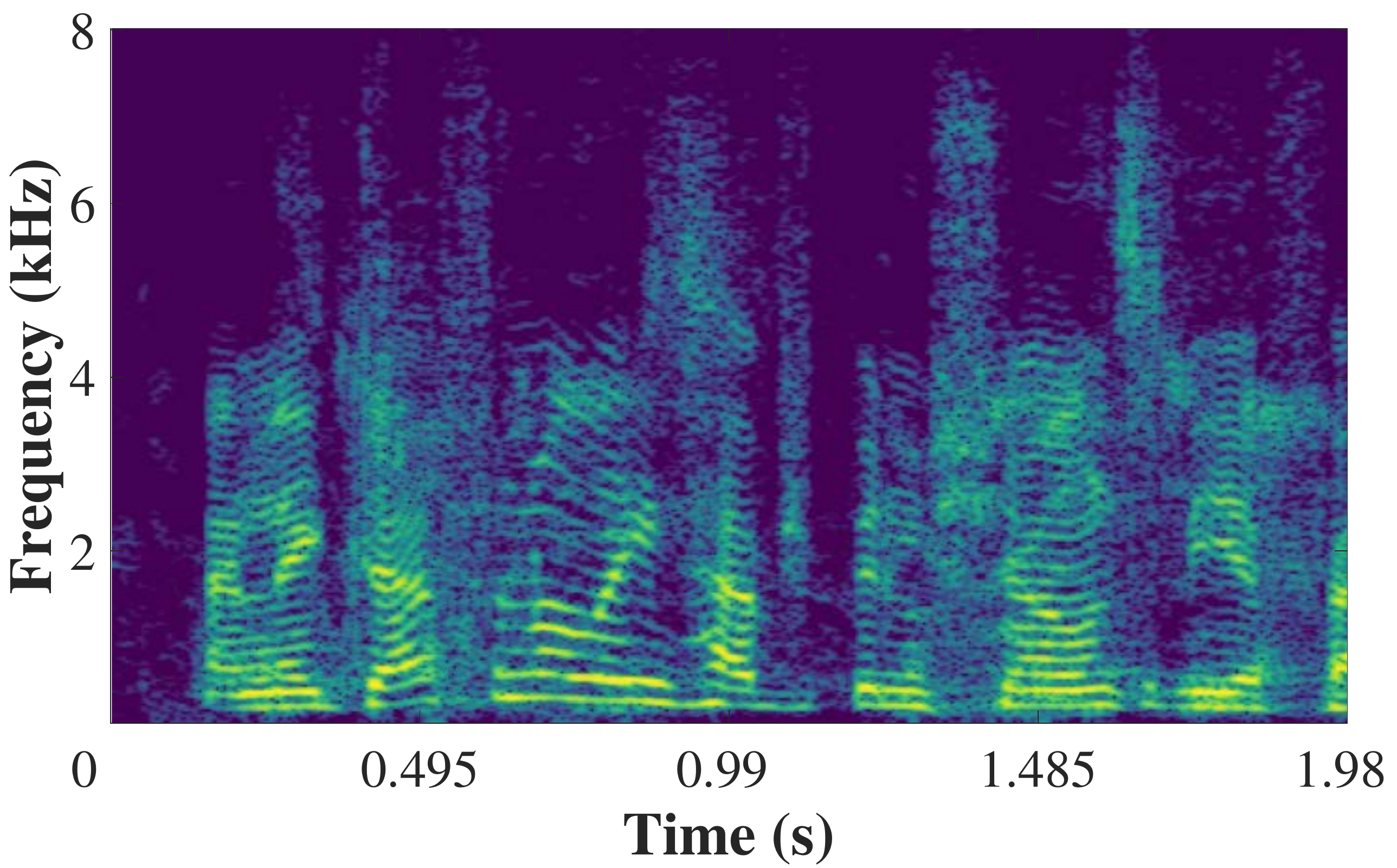}}\hfill\\
    \vspace{-10pt}
    \caption{Qualitative comparison results of spectrogram estimated by V-Conv~\cite{afouras2018conversation} and \basemodel on LRS3 dataset. All results are reconstructed from the combination of estimated magnitude and ground-truth phase.}
    \label{fig:lrs3}\vspace{-10pt}
\end{figure*}

\begin{figure*}[!ht]
\centering
    \renewcommand{\thesubfigure}{}
    \subfigure[]{
    \begin{minipage}{0.075\linewidth}\centering
        \includegraphics[width=\columnwidth]{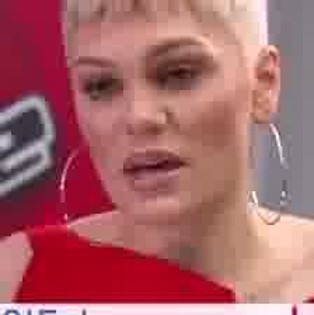} \\ \vspace{-3pt}
        \includegraphics[width=\columnwidth]{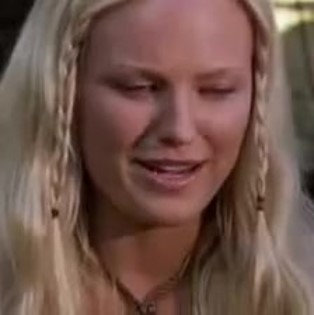}
    \end{minipage}\hfill
    \begin{minipage}{0.245\linewidth}\centering
        \includegraphics[width=\columnwidth]{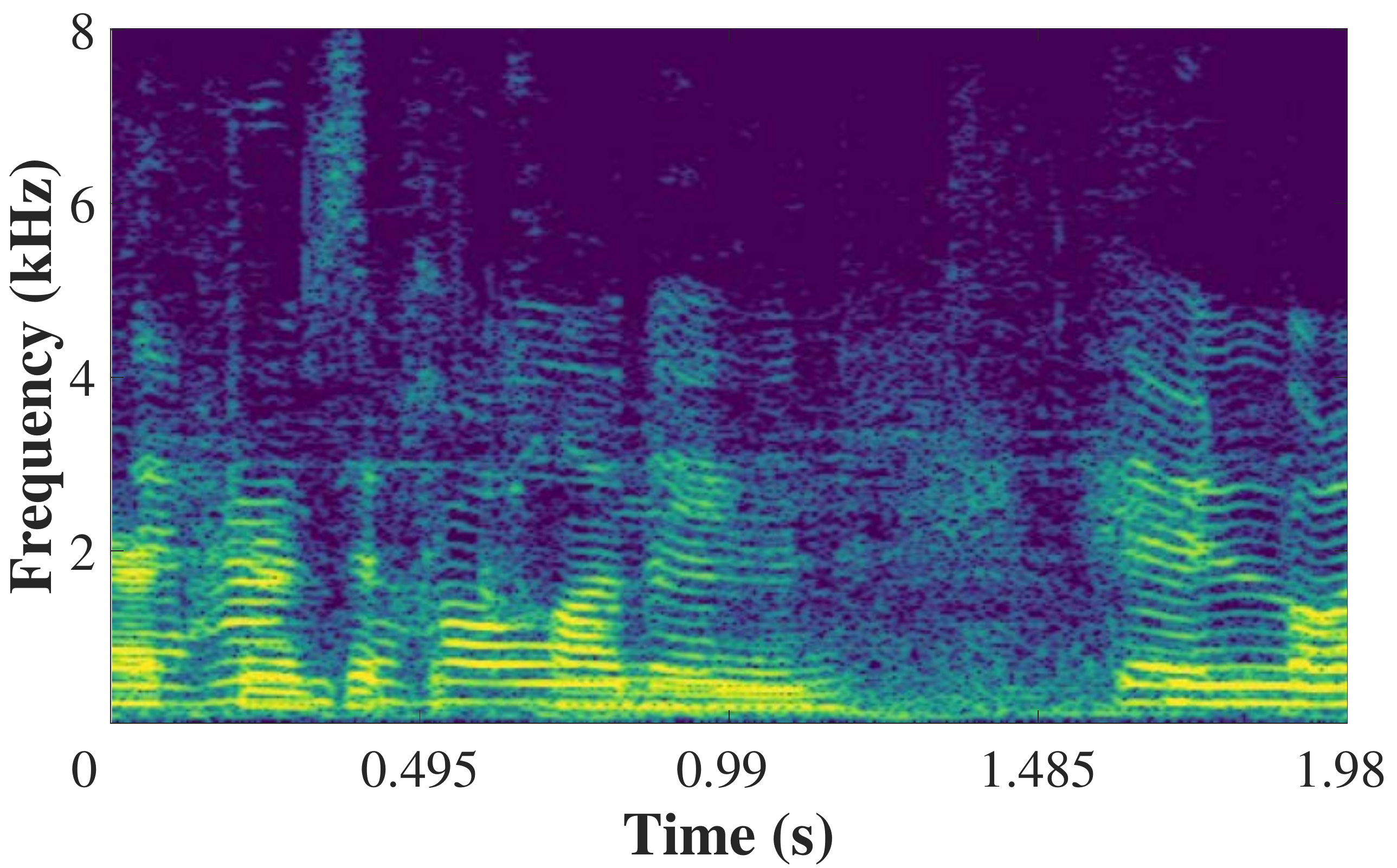}
    \end{minipage}\hfill
    }\hfill
    \subfigure[]{
    \begin{minipage}{0.075\linewidth}\centering
        \includegraphics[width=\columnwidth]{figsup_12_FF1_src.jpg}
    \end{minipage}\hfill
    \begin{minipage}{0.245\linewidth}\centering
        \includegraphics[width=\columnwidth]{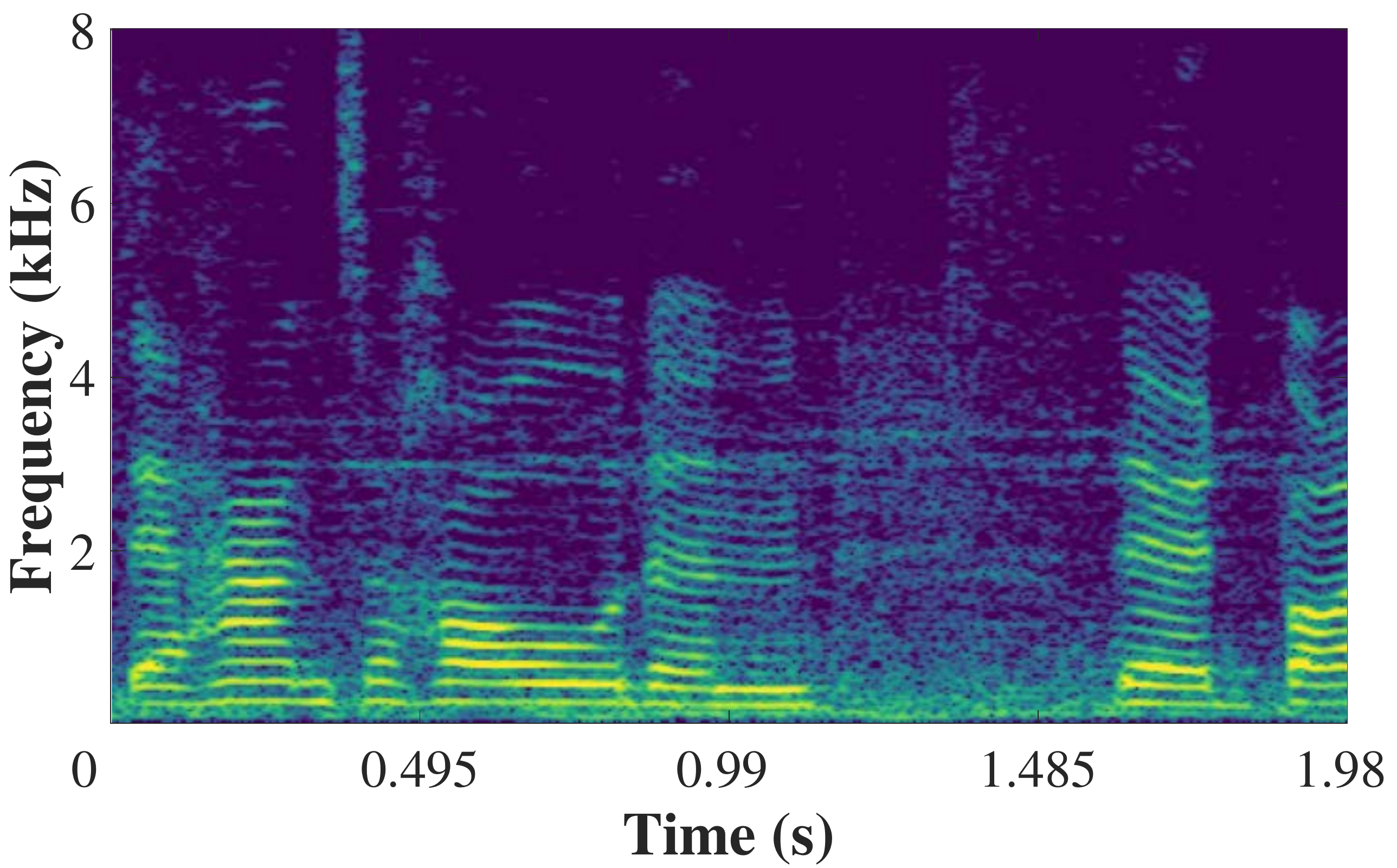}
    \end{minipage}\hfill
    }\hfill
    \subfigure[]{
    \begin{minipage}{0.075\linewidth}\centering
        \includegraphics[width=\columnwidth]{figsup_12_FF1_src.jpg} \\ \vspace{-3pt}
        \includegraphics[width=\columnwidth]{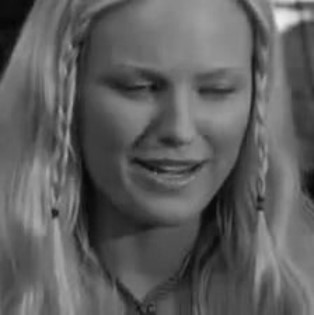}
    \end{minipage}\hfill
    \begin{minipage}{0.245\linewidth}\centering
        \includegraphics[width=\columnwidth]{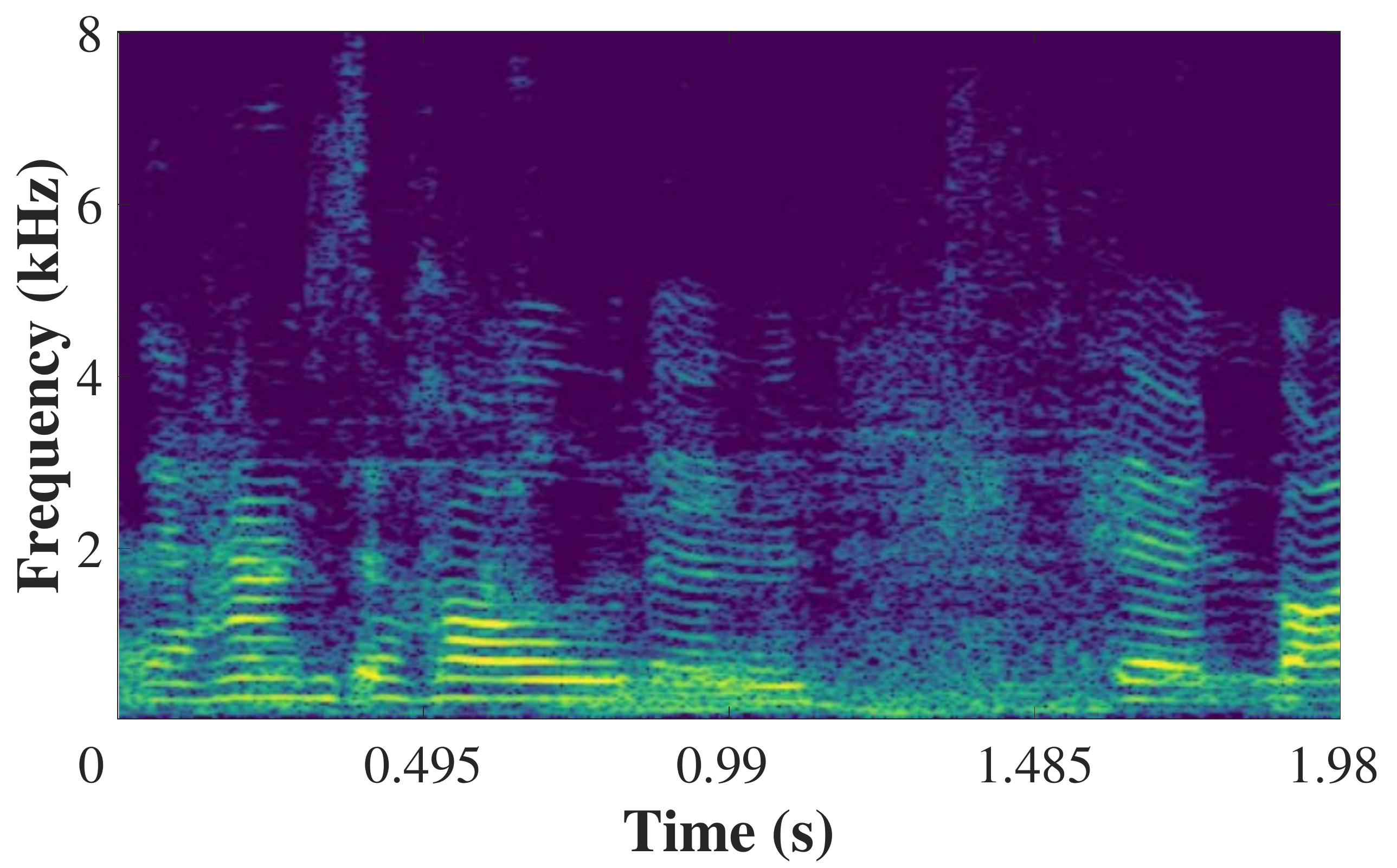}
    \end{minipage}\hfill
    }\hfill\\ \vspace{-11pt}
    
    \subfigure[]{
    \begin{minipage}{0.075\linewidth}\centering
        \includegraphics[width=\columnwidth]{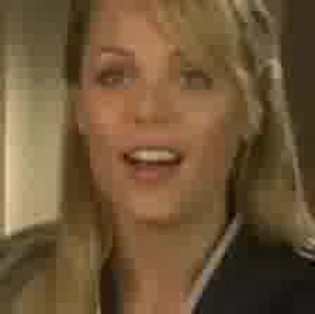} \\ \vspace{-3pt}
        \includegraphics[width=\columnwidth]{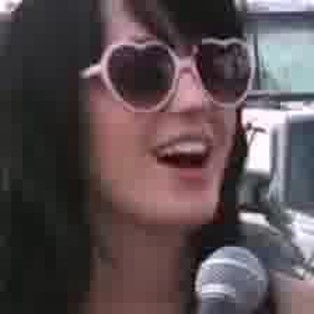}
    \end{minipage}\hfill
    \begin{minipage}{0.245\linewidth}\centering
        \includegraphics[width=\columnwidth]{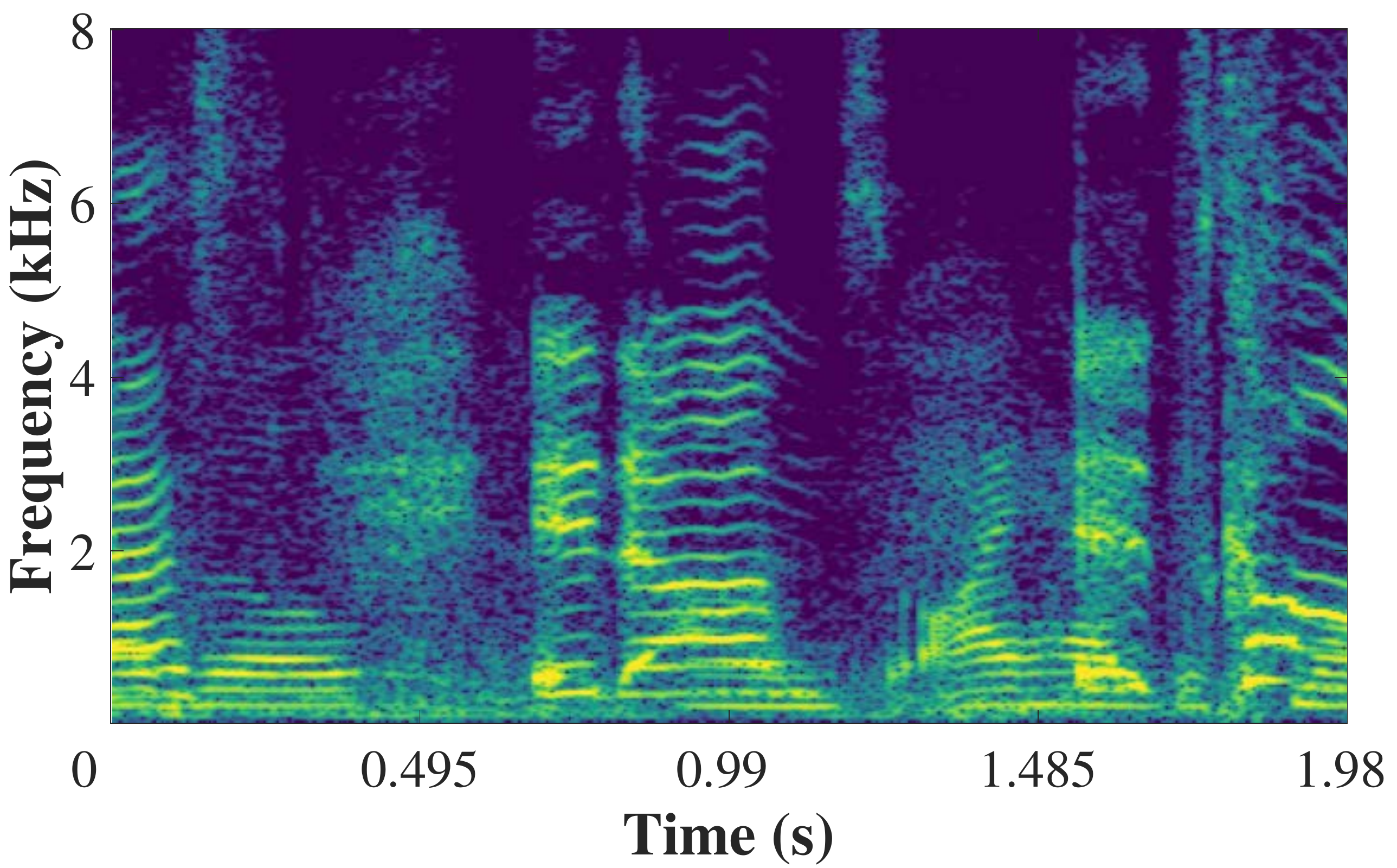}
    \end{minipage}\hfill
    }\hfill
    \subfigure[]{
    \begin{minipage}{0.075\linewidth}\centering
        \includegraphics[width=\columnwidth]{figsup_12_FF2_src.jpg}
    \end{minipage}\hfill
    \begin{minipage}{0.245\linewidth}\centering
        \includegraphics[width=\columnwidth]{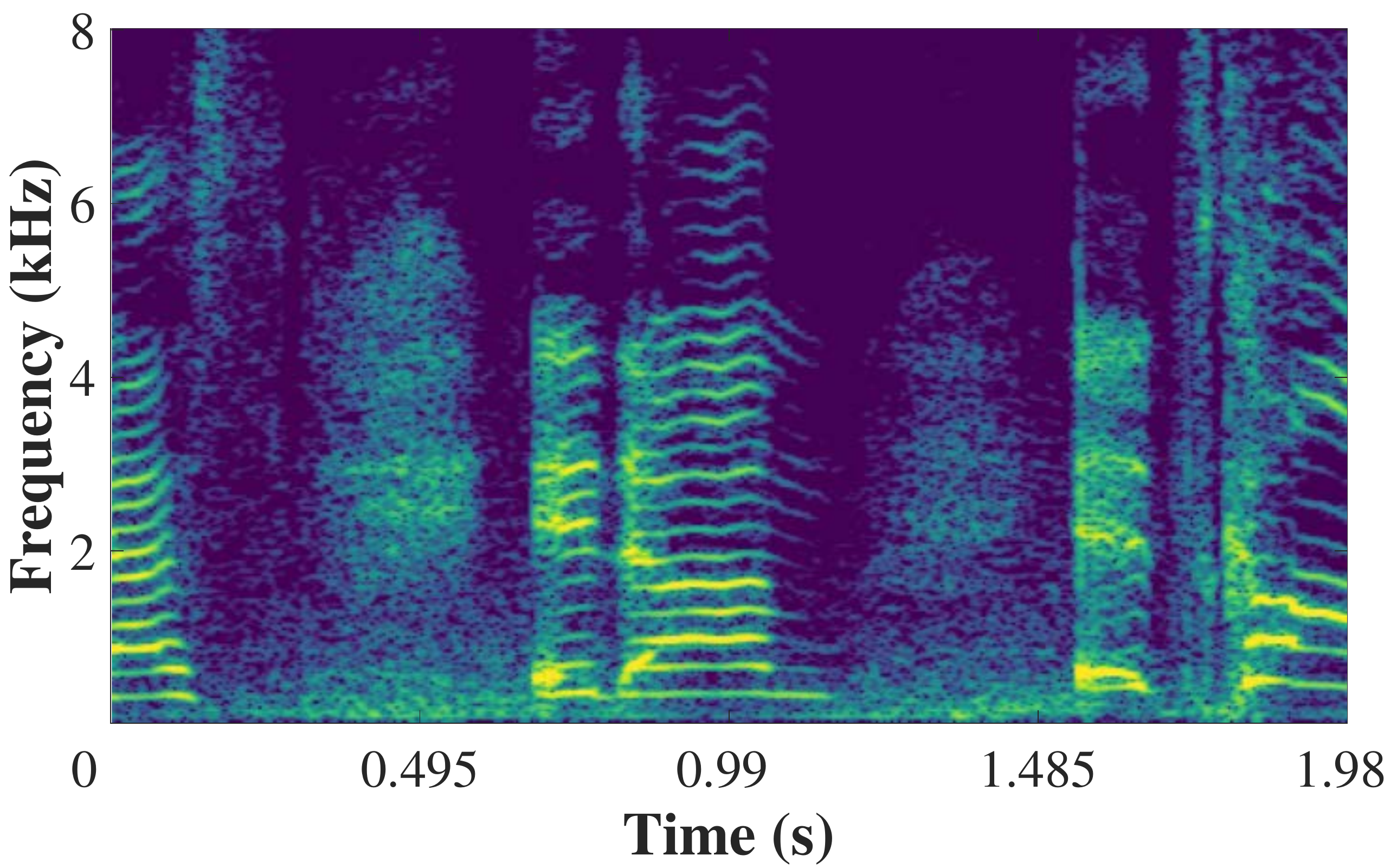}
    \end{minipage}\hfill
    }\hfill
    \subfigure[]{
    \begin{minipage}{0.075\linewidth}\centering
        \includegraphics[width=\columnwidth]{figsup_12_FF2_src.jpg} \\ \vspace{-3pt}
        \includegraphics[width=\columnwidth]{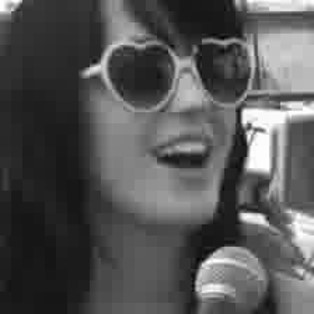}
    \end{minipage}\hfill
    \begin{minipage}{0.245\linewidth}\centering
        \includegraphics[width=\columnwidth]{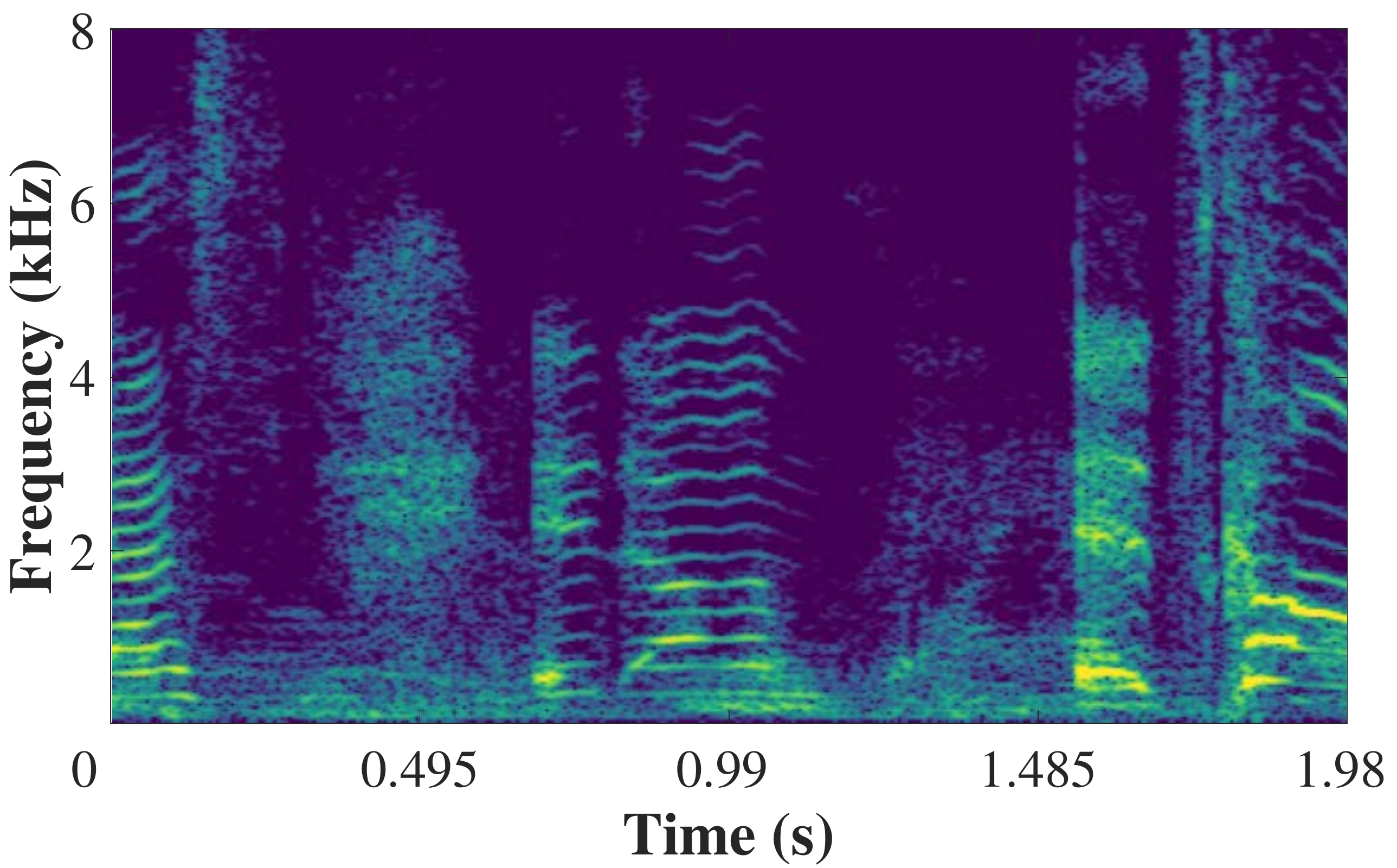}
    \end{minipage}\hfill
    }\hfill\\ \vspace{-11pt}
    
    \subfigure[]{
    \begin{minipage}{0.075\linewidth}\centering
        \includegraphics[width=\columnwidth]{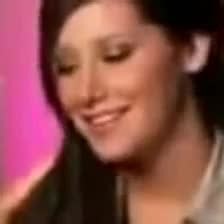} \\ \vspace{-3pt}
        \includegraphics[width=\columnwidth]{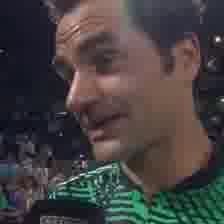}
    \end{minipage}\hfill
    \begin{minipage}{0.245\linewidth}\centering
        \includegraphics[width=\columnwidth]{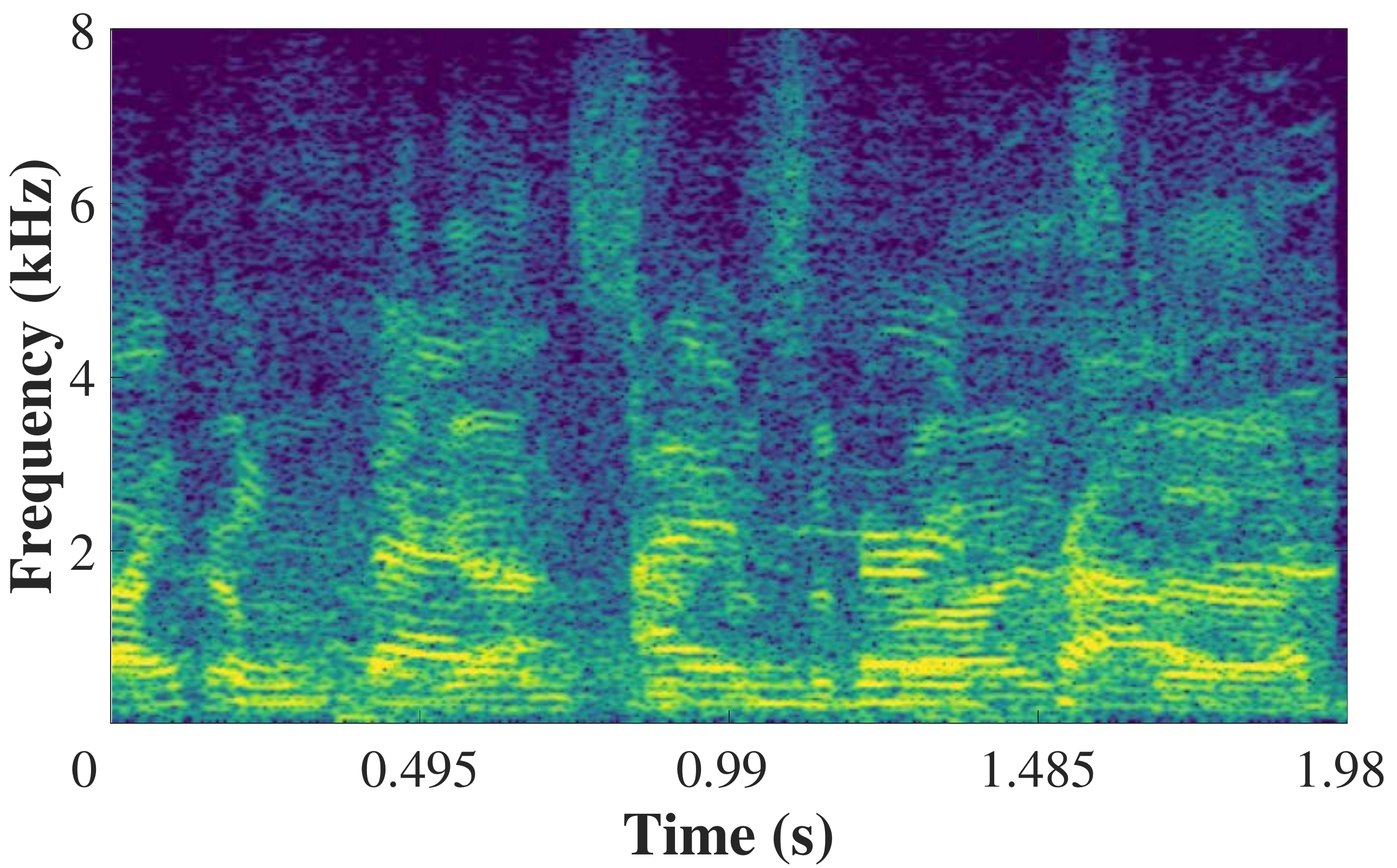}
    \end{minipage}\hfill
    }\hfill
    \subfigure[]{
    \begin{minipage}{0.075\linewidth}\centering
        \includegraphics[width=\columnwidth]{figsup_12_FM1_src.jpg}
    \end{minipage}\hfill
    \begin{minipage}{0.245\linewidth}\centering
        \includegraphics[width=\columnwidth]{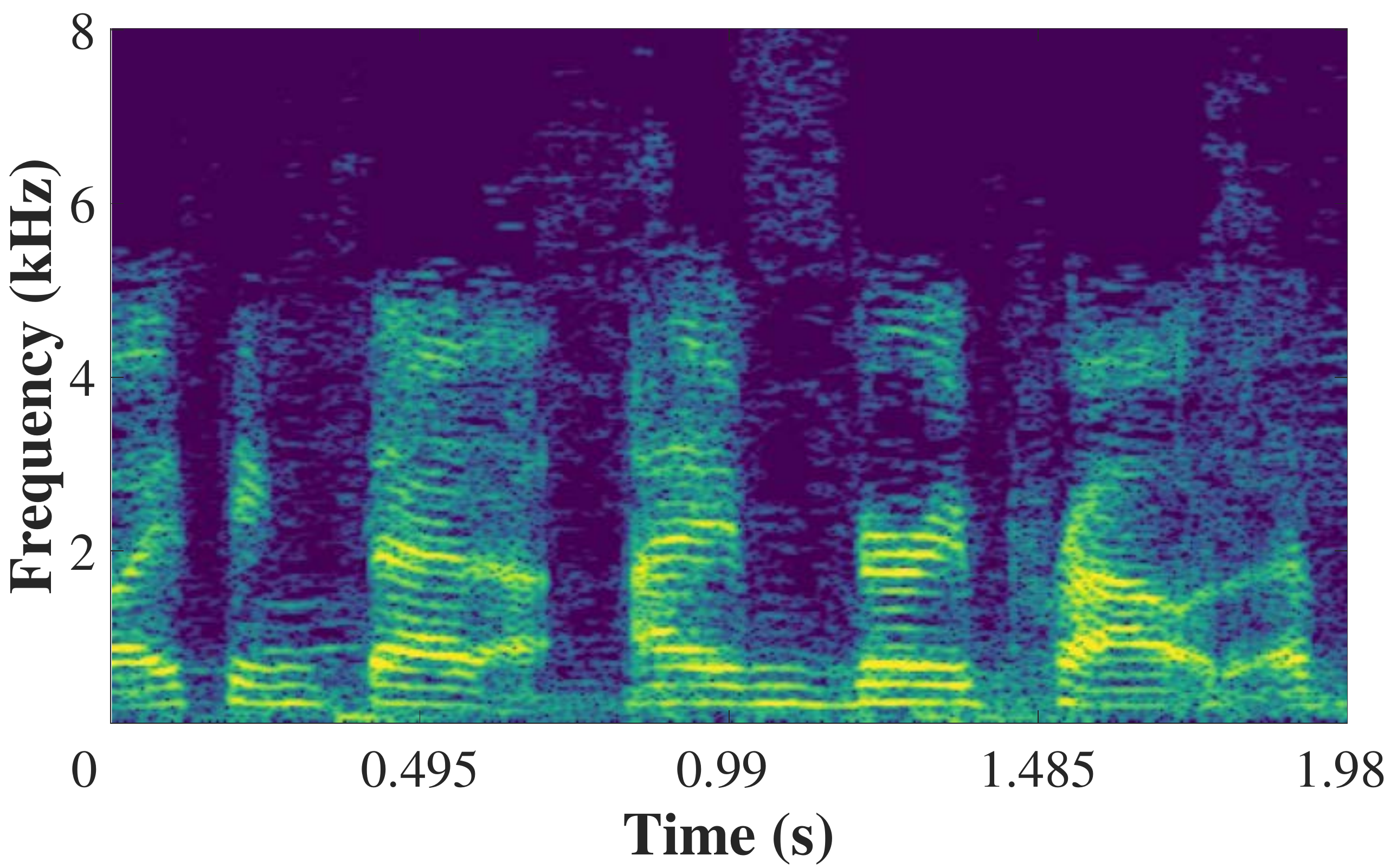}
    \end{minipage}\hfill
    }\hfill
    \subfigure[]{
    \begin{minipage}{0.075\linewidth}\centering
        \includegraphics[width=\columnwidth]{figsup_12_FM1_src.jpg} \\ \vspace{-3pt}
        \includegraphics[width=\columnwidth]{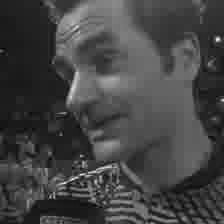}
    \end{minipage}\hfill
    \begin{minipage}{0.245\linewidth}\centering
        \includegraphics[width=\columnwidth]{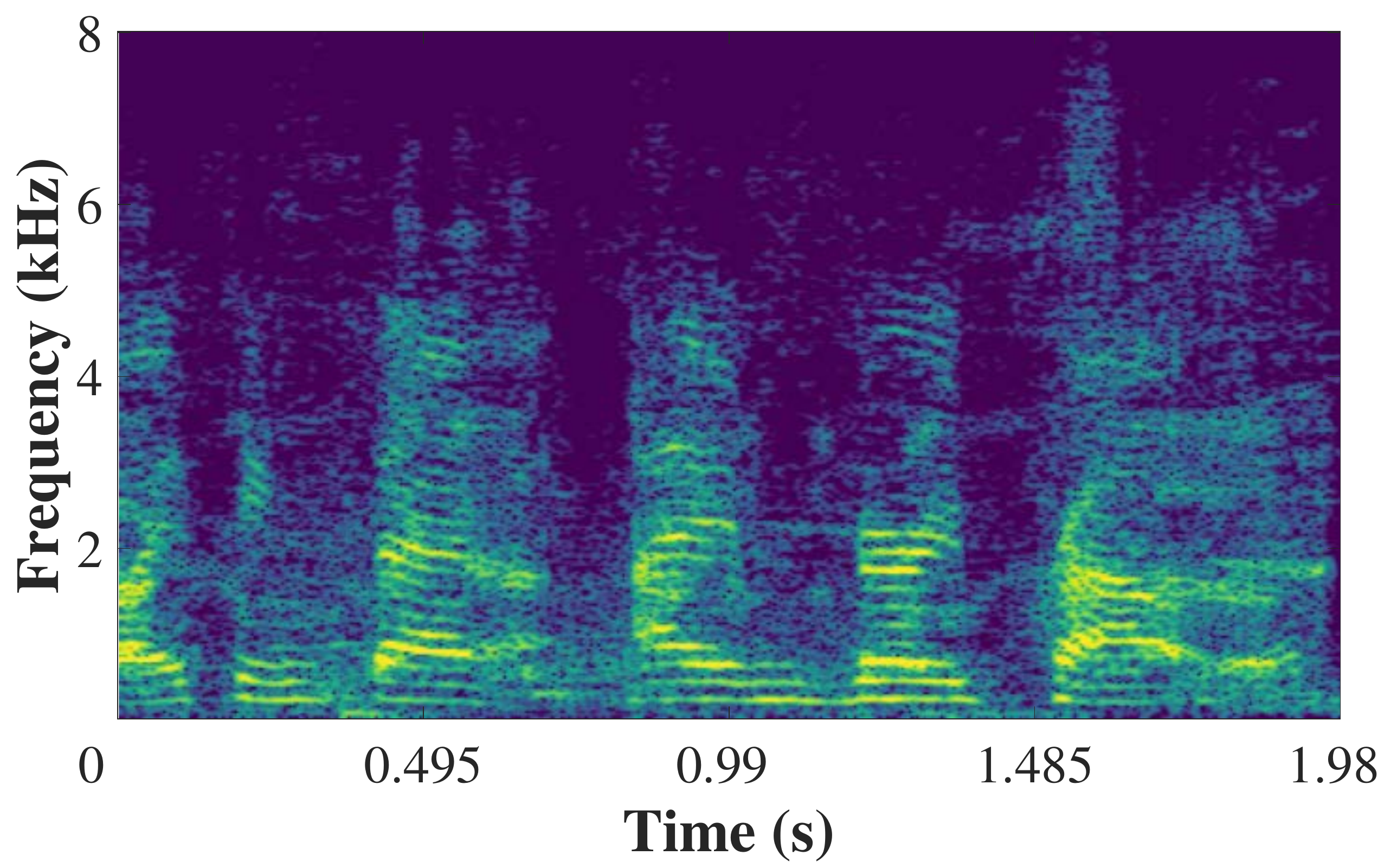}
    \end{minipage}\hfill
    }\hfill\\ \vspace{-11pt}
    
    \subfigure[]{
    \begin{minipage}{0.075\linewidth}\centering
        \includegraphics[width=\columnwidth]{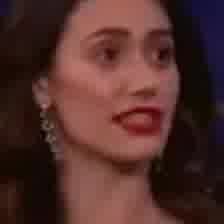} \\ \vspace{-3pt}
        \includegraphics[width=\columnwidth]{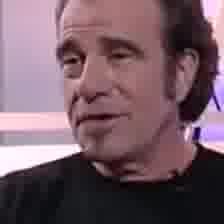}
    \end{minipage}\hfill
    \begin{minipage}{0.245\linewidth}\centering
        \includegraphics[width=\columnwidth]{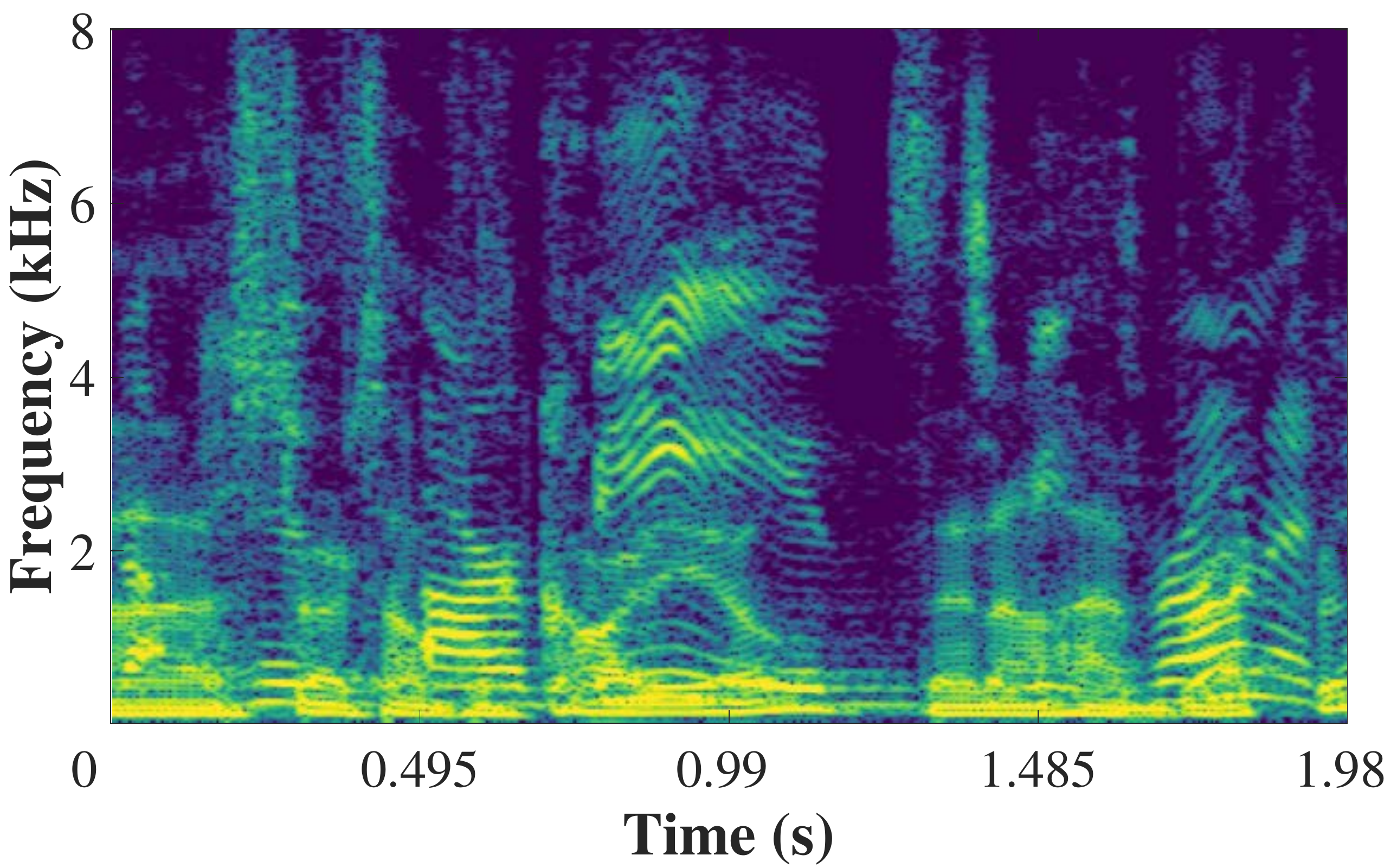}
    \end{minipage}\hfill
    }\hfill
    \subfigure[]{
    \begin{minipage}{0.075\linewidth}\centering
        \includegraphics[width=\columnwidth]{figsup_12_FM2_src.jpg}
    \end{minipage}\hfill
    \begin{minipage}{0.245\linewidth}\centering
        \includegraphics[width=\columnwidth]{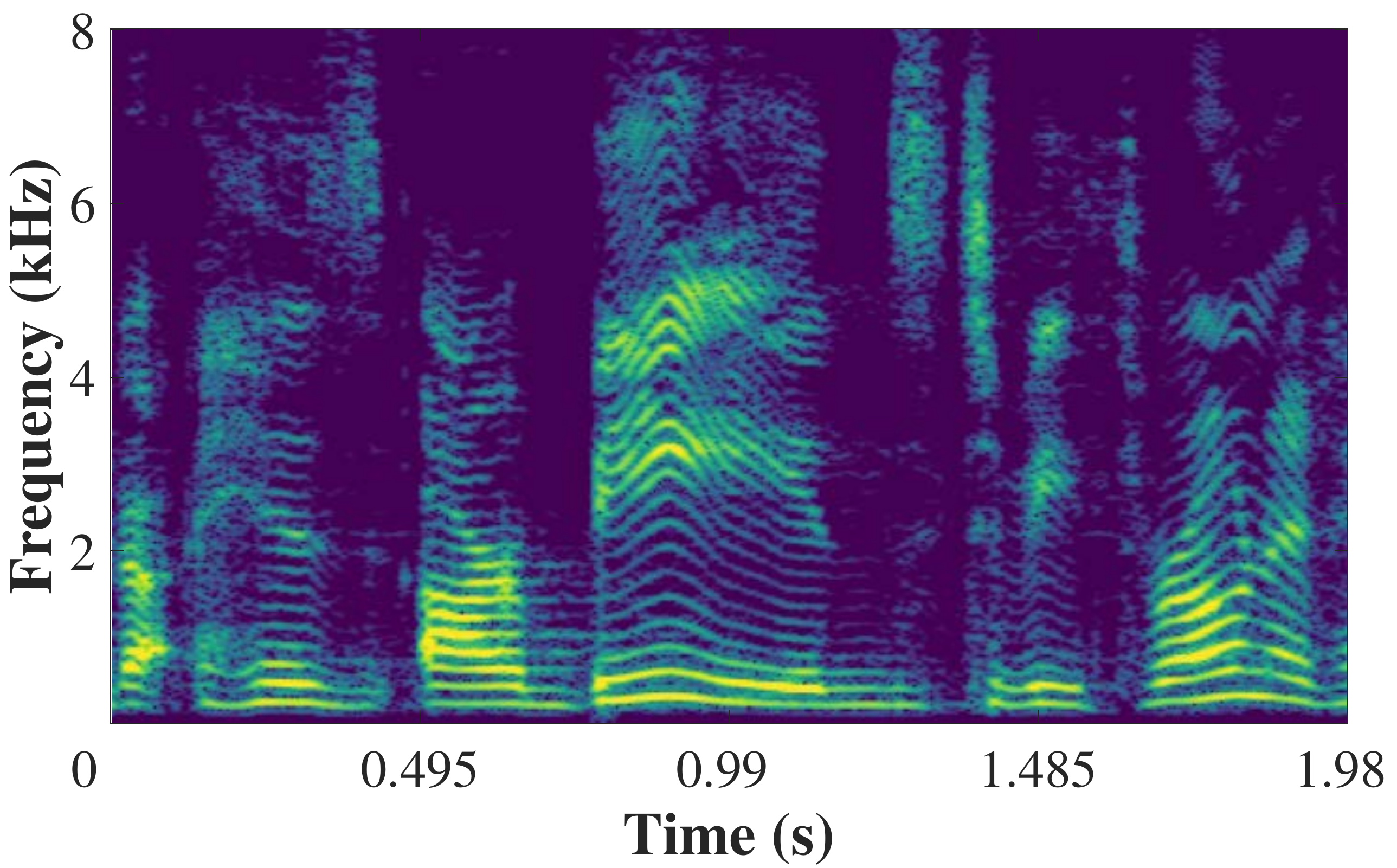}
    \end{minipage}\hfill
    }\hfill
    \subfigure[]{
    \begin{minipage}{0.075\linewidth}\centering
        \includegraphics[width=\columnwidth]{figsup_12_FM2_src.jpg} \\ \vspace{-3pt}
        \includegraphics[width=\columnwidth]{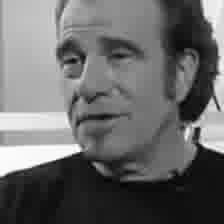}
    \end{minipage}\hfill
    \begin{minipage}{0.245\linewidth}\centering
        \includegraphics[width=\columnwidth]{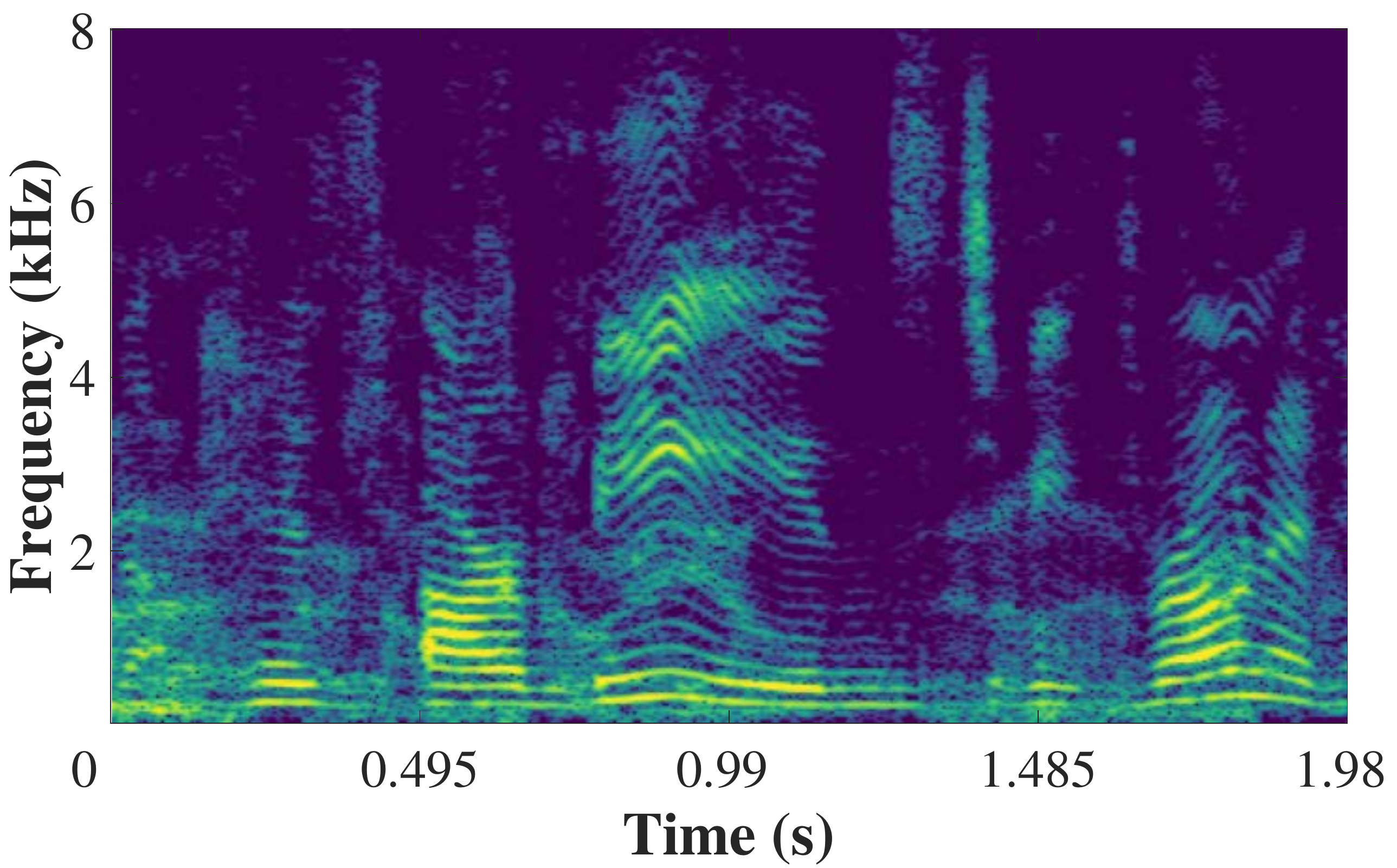}
    \end{minipage}\hfill
    }\hfill\\ \vspace{-11pt}
    
    \subfigure[]{
        \begin{minipage}{0.075\linewidth}\centering
            \includegraphics[width=\columnwidth]{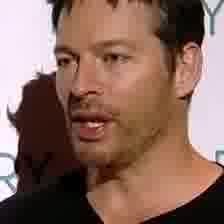} \\ \vspace{-3pt}
            \includegraphics[width=\columnwidth]{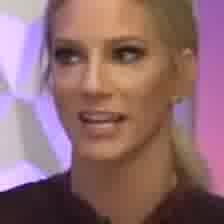}
        \end{minipage}\hfill
        \begin{minipage}{0.245\linewidth}\centering
            \includegraphics[width=\columnwidth]{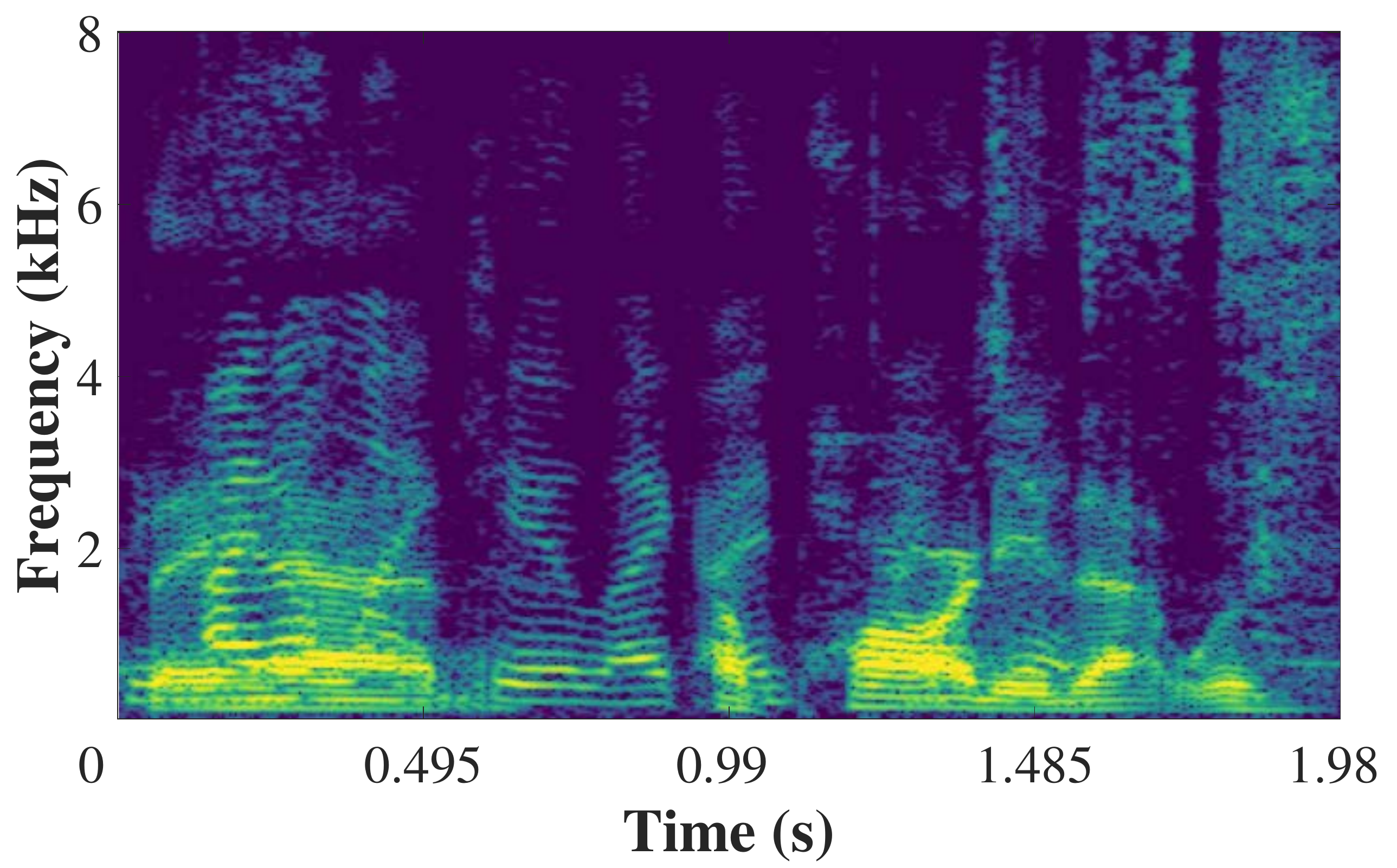}
        \end{minipage}\hfill
        }\hfill
        \subfigure[]{
        \begin{minipage}{0.075\linewidth}\centering
            \includegraphics[width=\columnwidth]{figsup_12_MF1_src.jpg}
        \end{minipage}\hfill
        \begin{minipage}{0.245\linewidth}\centering
            \includegraphics[width=\columnwidth]{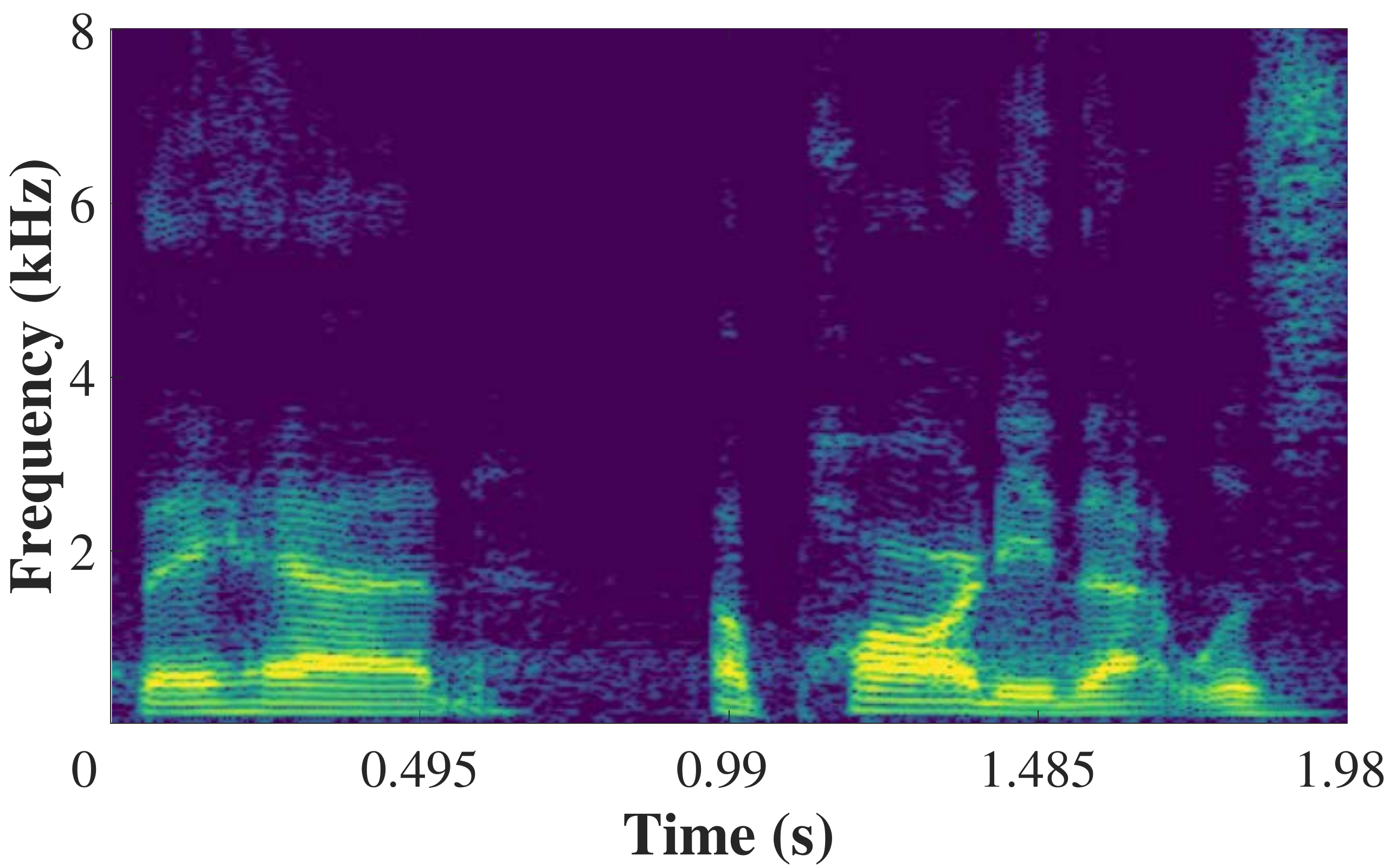}
        \end{minipage}\hfill
        }\hfill
        \subfigure[]{
        \begin{minipage}{0.075\linewidth}\centering
            \includegraphics[width=\columnwidth]{figsup_12_MF1_src.jpg} \\ \vspace{-3pt}
            \includegraphics[width=\columnwidth]{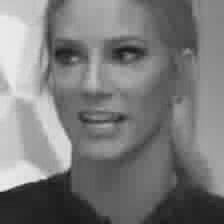}
        \end{minipage}\hfill
        \begin{minipage}{0.245\linewidth}\centering
            \includegraphics[width=\columnwidth]{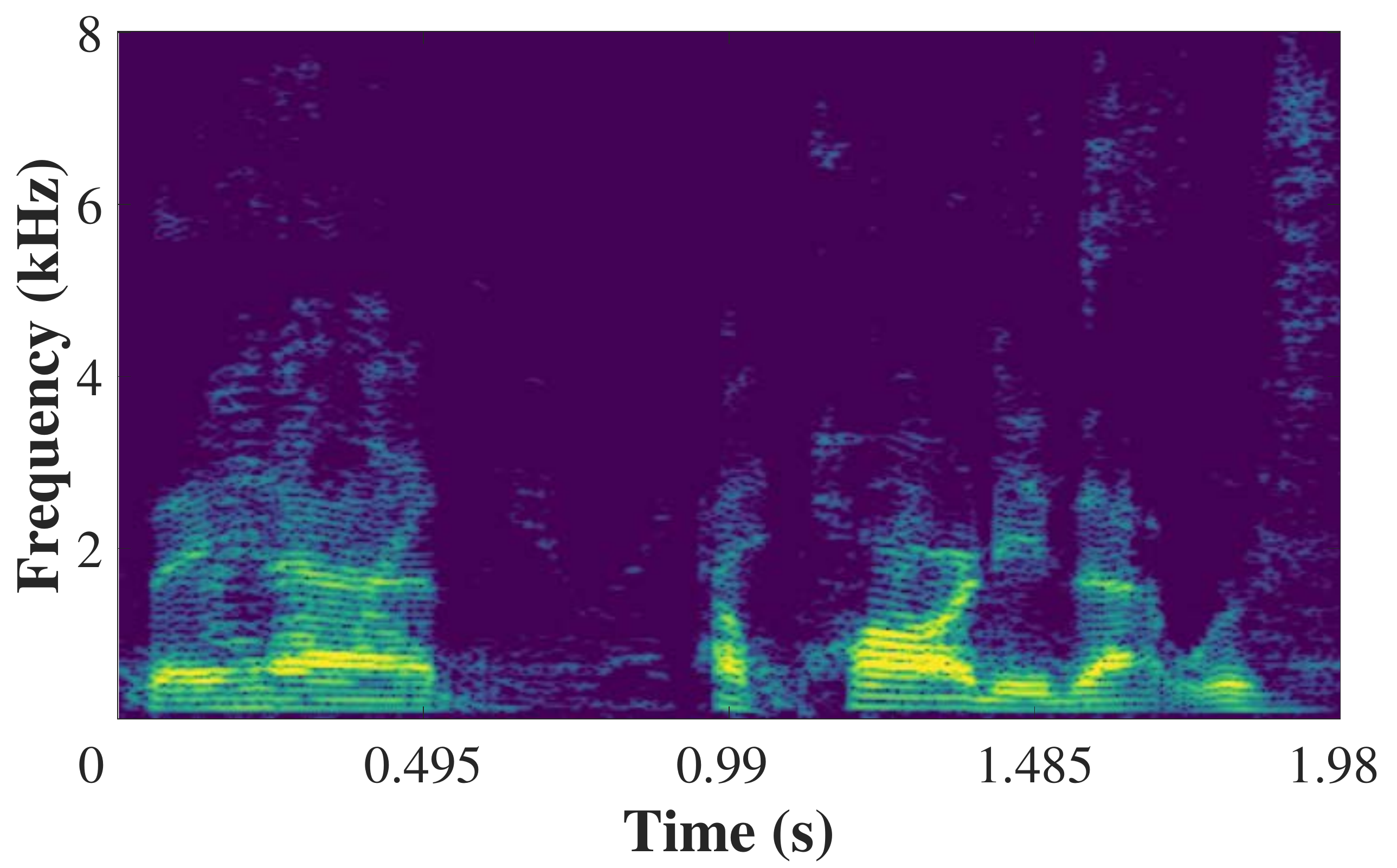}
        \end{minipage}\hfill
    }\hfill \\ \vspace{-11pt}
    
    \subfigure[]{
        \begin{minipage}{0.075\linewidth}\centering
            \includegraphics[width=\columnwidth]{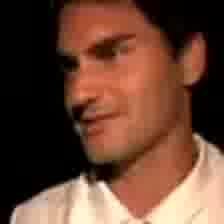} \\ \vspace{-3pt}
            \includegraphics[width=\columnwidth]{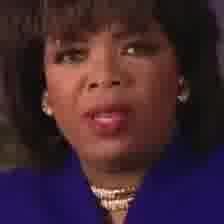}
        \end{minipage}\hfill
        \begin{minipage}{0.245\linewidth}\centering
            \includegraphics[width=\columnwidth]{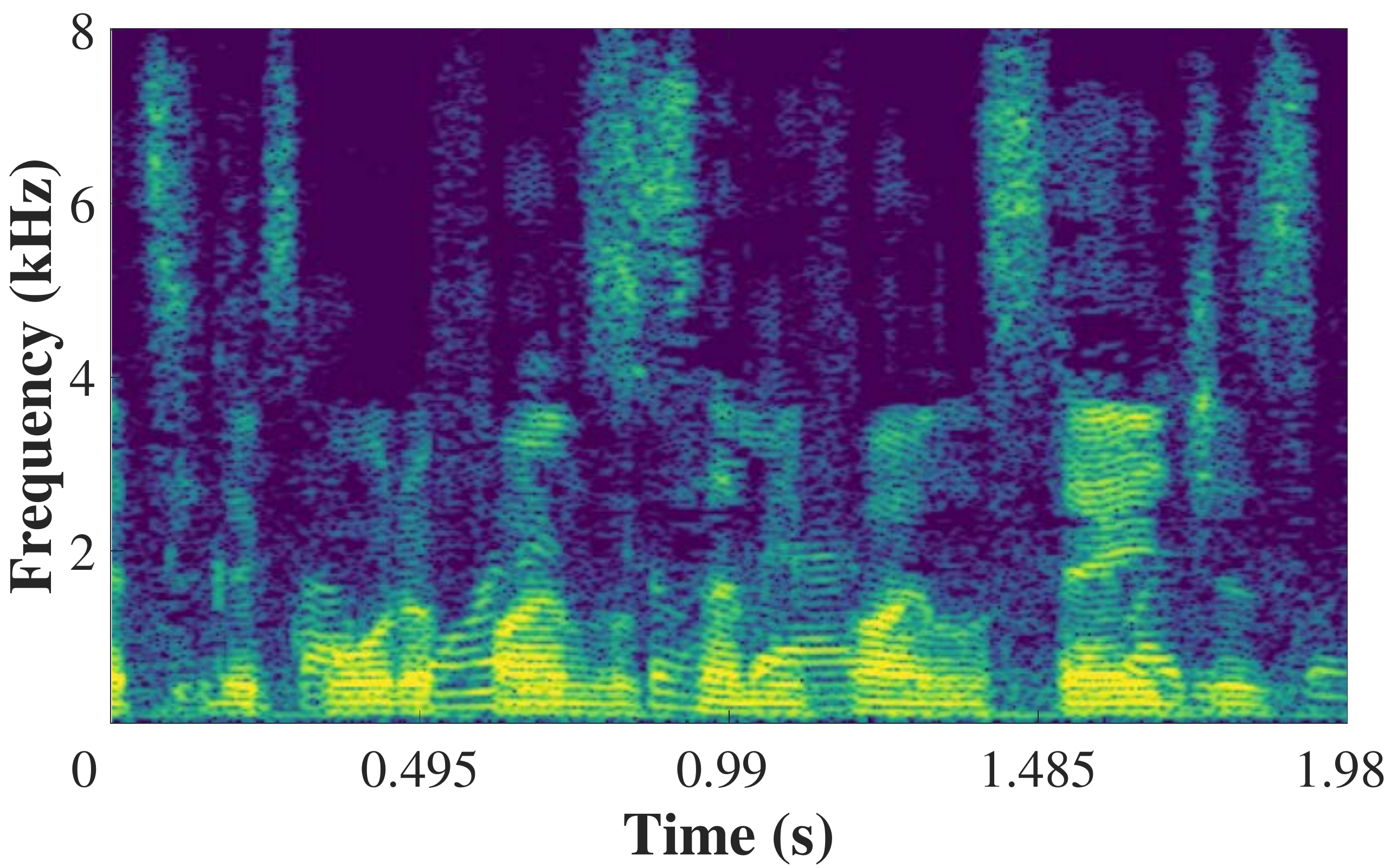}
        \end{minipage}\hfill
        }\hfill
        \subfigure[]{
        \begin{minipage}{0.075\linewidth}\centering
            \includegraphics[width=\columnwidth]{figsup_12_MF2_src.jpg}
        \end{minipage}\hfill
        \begin{minipage}{0.245\linewidth}\centering
            \includegraphics[width=\columnwidth]{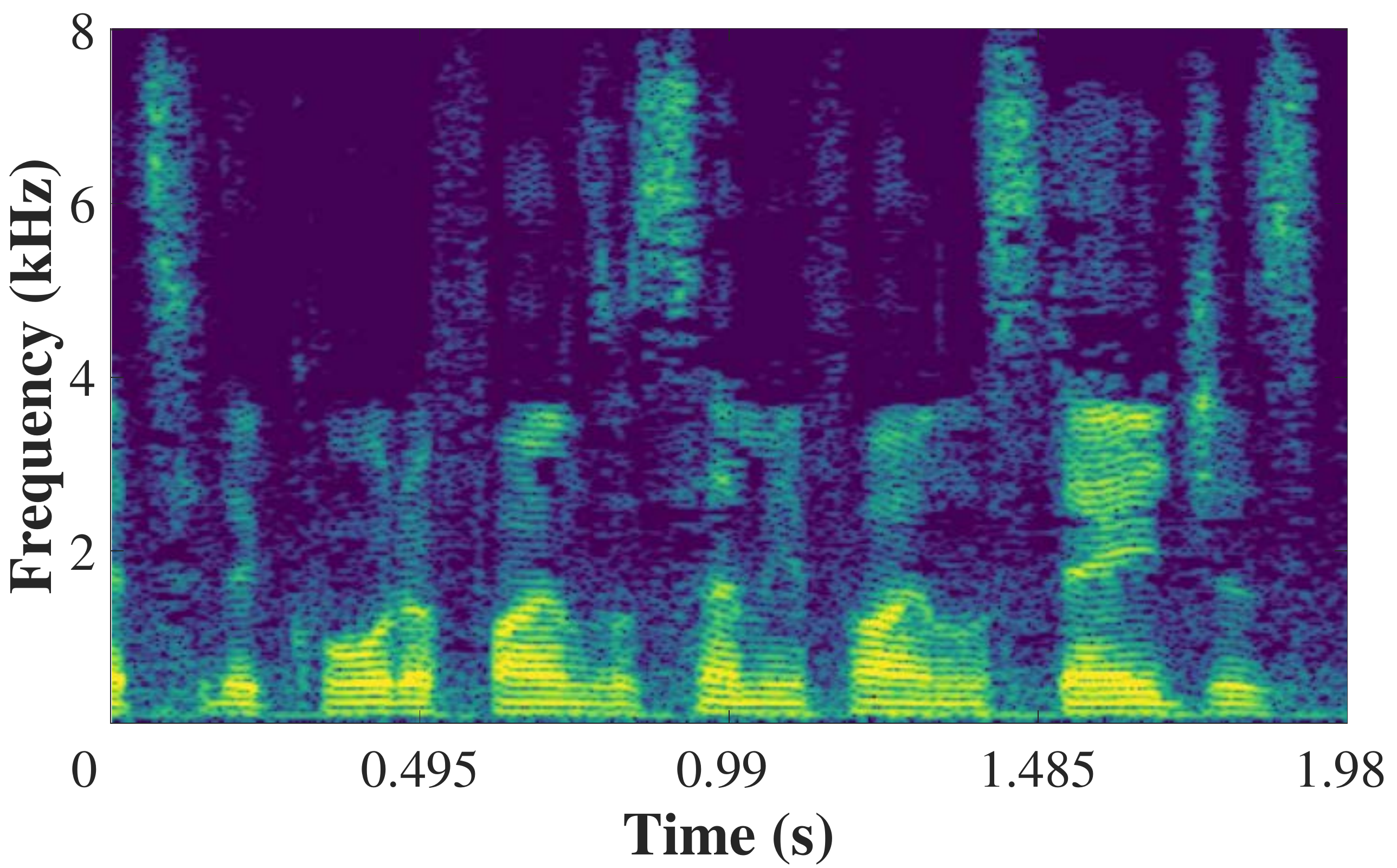}
        \end{minipage}\hfill
        }\hfill
        \subfigure[]{
        \begin{minipage}{0.075\linewidth}\centering
            \includegraphics[width=\columnwidth]{figsup_12_MF2_src.jpg} \\ \vspace{-3pt}
            \includegraphics[width=\columnwidth]{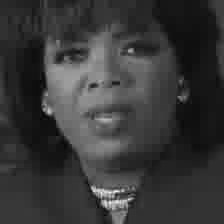}
        \end{minipage}\hfill
        \begin{minipage}{0.245\linewidth}\centering
            \includegraphics[width=\columnwidth]{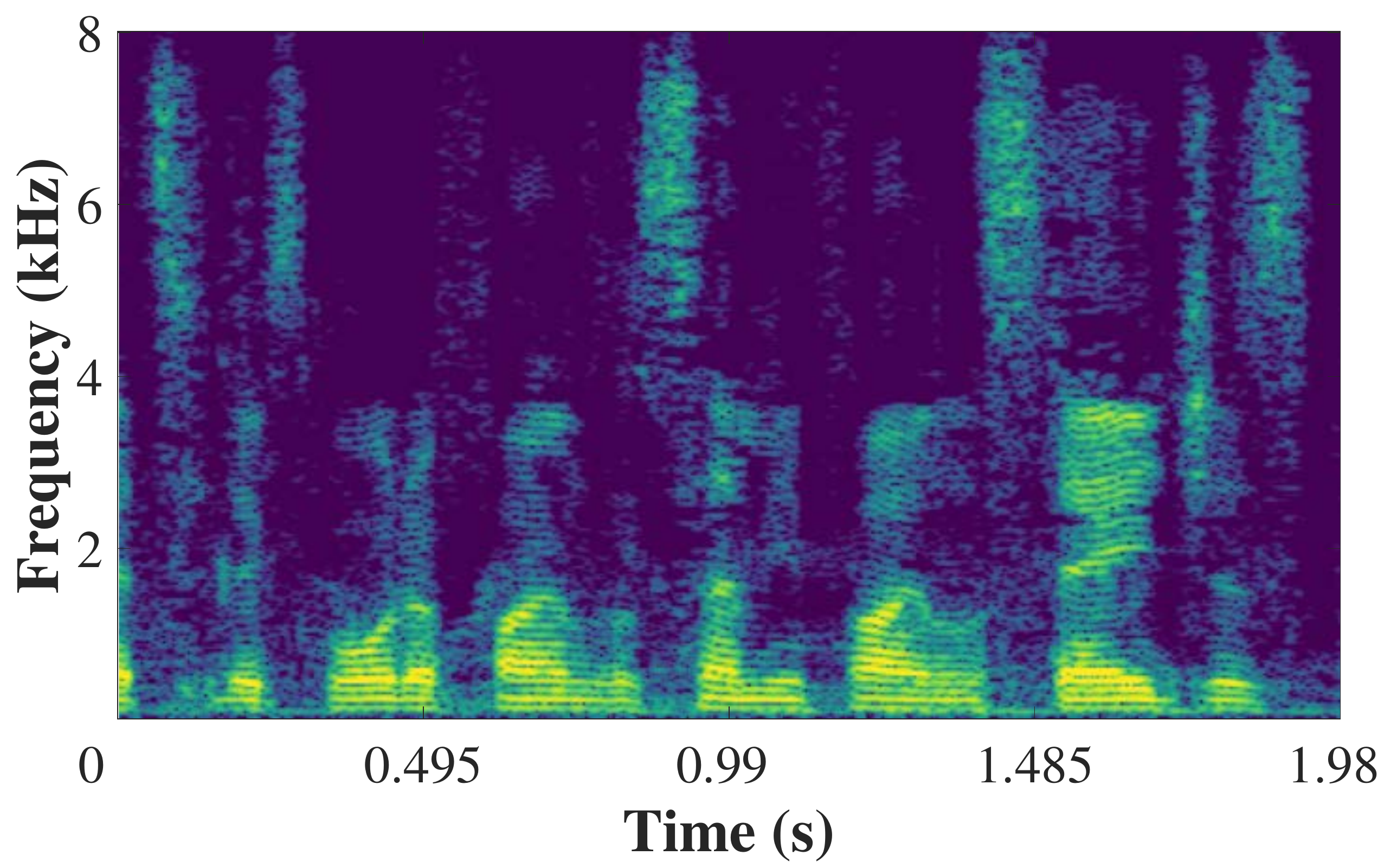}
        \end{minipage}\hfill
    }\hfill\\ \vspace{-11pt}
    
    \subfigure[(a) Mixture speech]{
    \begin{minipage}{0.075\linewidth}\centering
        \includegraphics[width=\columnwidth]{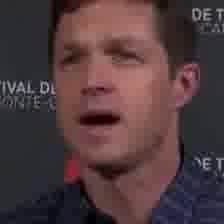} \\ \vspace{-3pt}
        \includegraphics[width=\columnwidth]{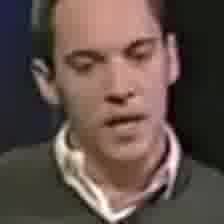}
    \end{minipage}\hfill
    \begin{minipage}{0.245\linewidth}\centering
        \includegraphics[width=\columnwidth]{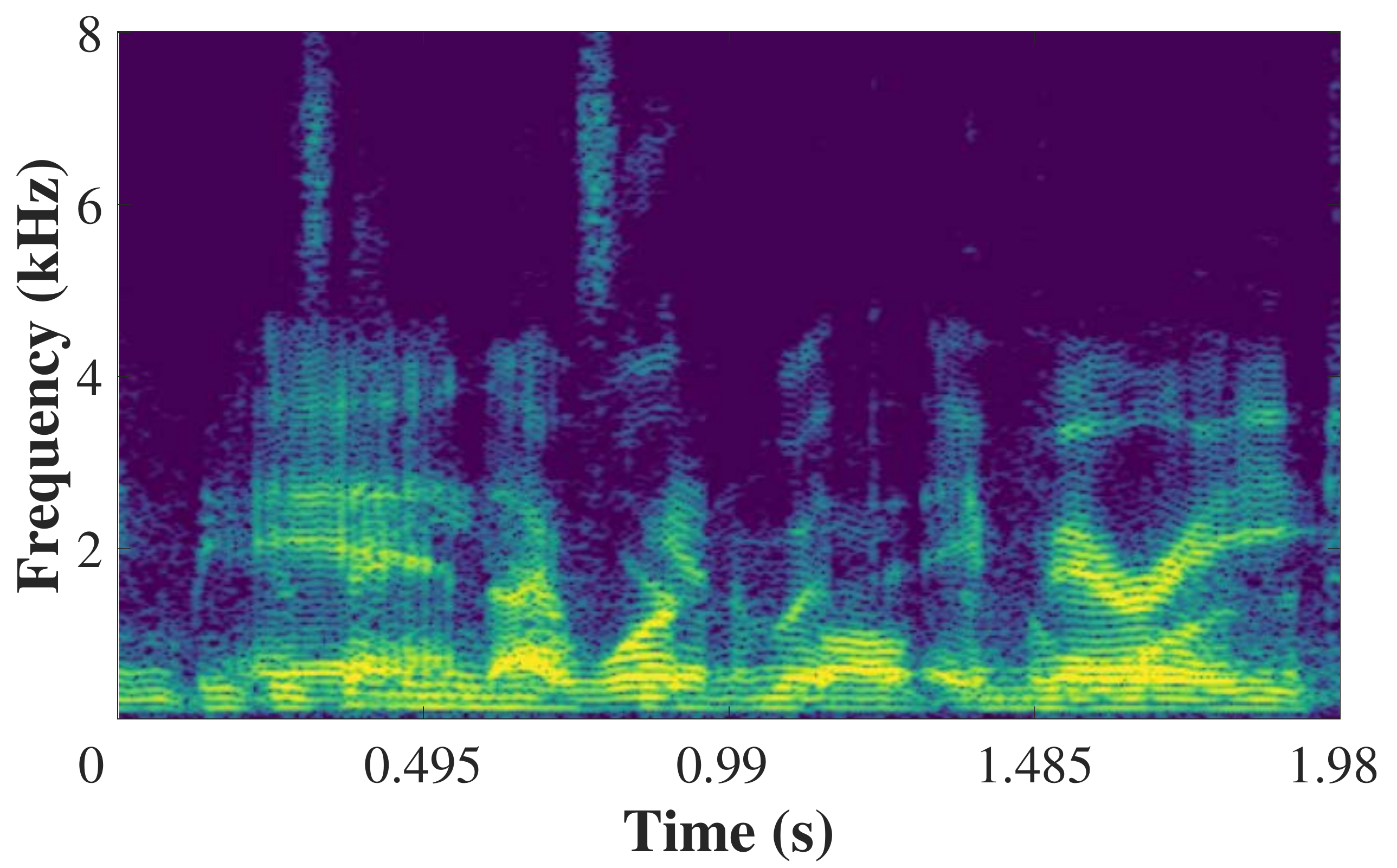}
    \end{minipage}\hfill
    }\hfill
    \subfigure[(b) Target speech]{
    \begin{minipage}{0.075\linewidth}\centering
        \includegraphics[width=\columnwidth]{figsup_12_MM1_src.jpg}
    \end{minipage}\hfill
    \begin{minipage}{0.245\linewidth}\centering
        \includegraphics[width=\columnwidth]{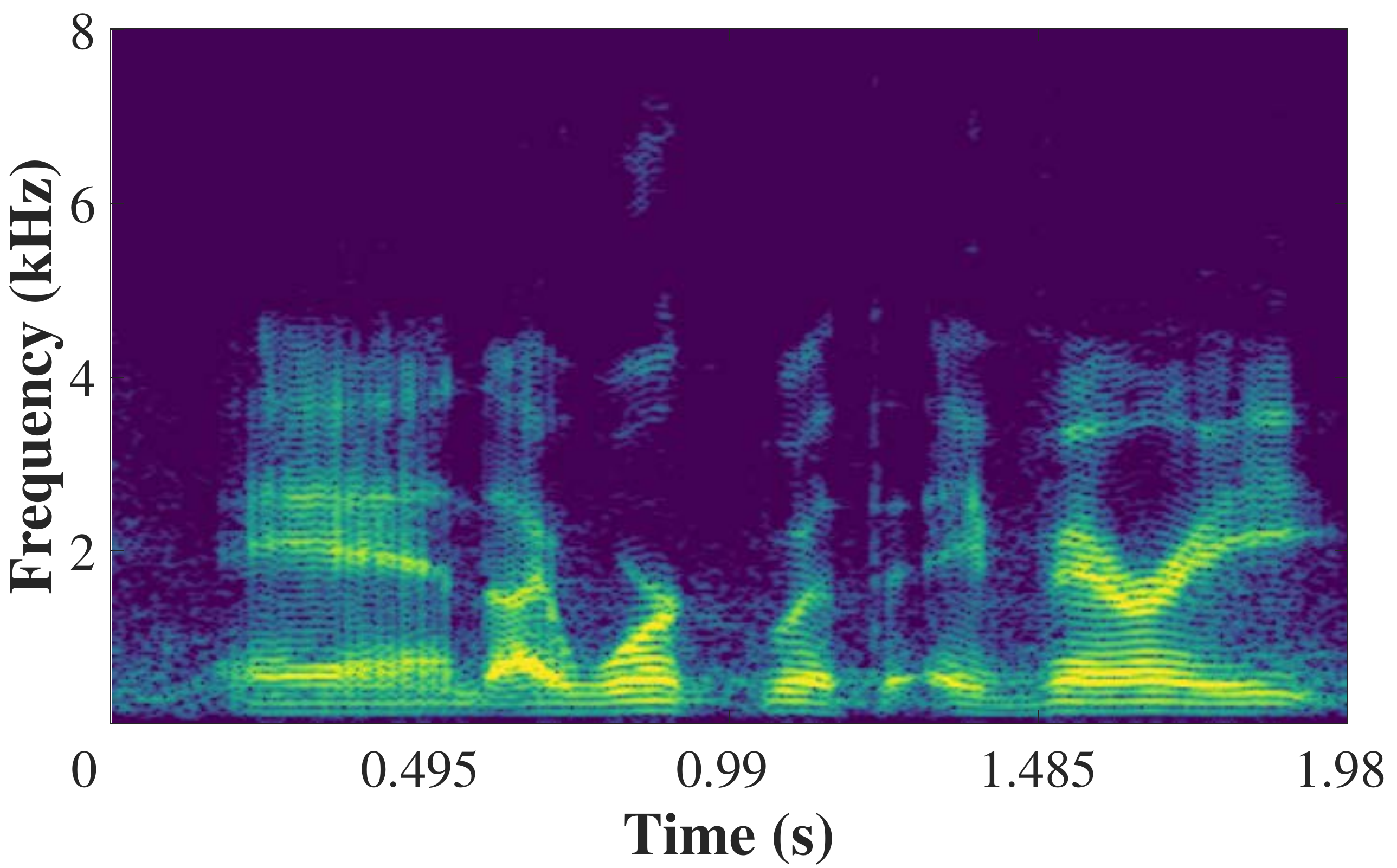}
    \end{minipage}\hfill
    }\hfill
    \subfigure[(c) Enhanced speech]{
    \begin{minipage}{0.075\linewidth}\centering
        \includegraphics[width=\columnwidth]{figsup_12_MM1_src.jpg} \\ \vspace{-3pt}
        \includegraphics[width=\columnwidth]{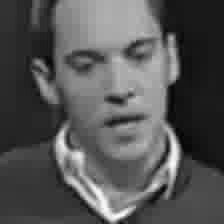}
    \end{minipage}\hfill
    \begin{minipage}{0.245\linewidth}\centering
        \includegraphics[width=\columnwidth]{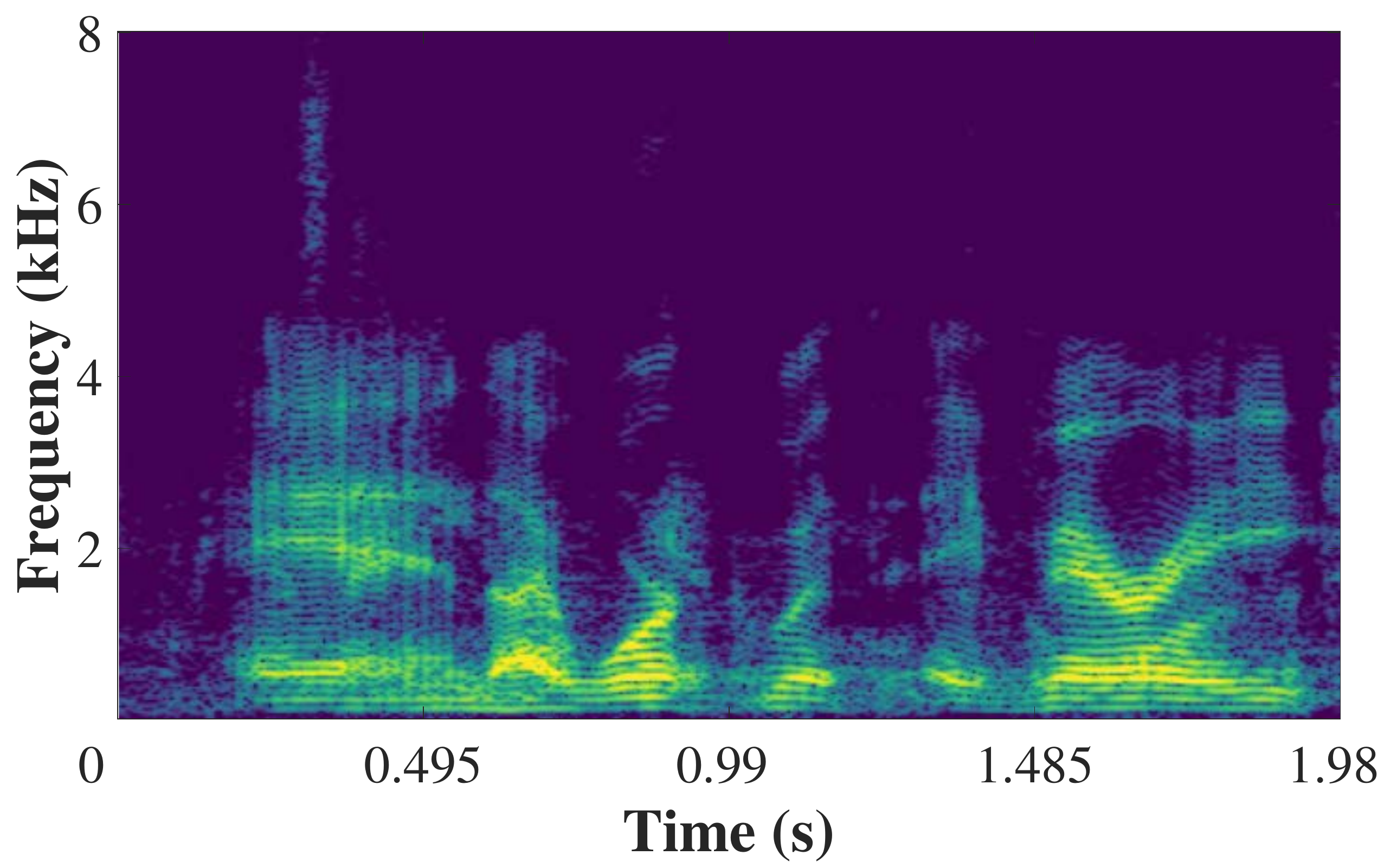}
    \end{minipage}\hfill
    }\hfill\\ \vspace{-11pt}
    \caption{Examples of speech separation results on VoxCeleb2 dataset. From left to right, each column notes mixed speech, target speech and separated output. These are randomly selected samples of \complexmodel with full prediction of magnitude and phase components.}
    \label{fig:vox2}\vspace{-11pt}
\end{figure*}

\paragraph{Qualitative Results of Spectrograms}
We provide more qualitative results of spectrograms.
\figref{fig:lrs2jitter} shows the examples with and without jitter.
\figref{fig:lrs2} includes the spectrogram results of three different testing samples.
In this experiment, we sample by immobilizing the audio stream at a certain time and moving the video segment according to the degree of delay.
Although the video stream does not synchronize exactly with the audio stream, the results are similarly predicted.
In~\figref{fig:lrs3}, we compare the results with V-Conv~\cite{afouras2018conversation}.
These results validate that the \basemodel are flexible and interpretable for the delay, thus widely applicable on real-world videos.
\figref{fig:vox2} demonstrates the results of spectrograms on VoxCeleb2 dataset.
At the last column in \figref{fig:vox2}, colored image means the target speaker and gray-scale image indicates the interfering speaker.
The target speaker's speech is isolated while the other's interfering speech is suppressed.

\paragraph{Video}
Our demo video\footnote{\url{https://youtu.be/9R2qQ7dGTp8}} includes the examples of videos on LRS2 dataset, and the examples from real-world news videos.

\end{document}